\documentclass[journal]{vgtc}                     
\usepackage{amsmath}
\usepackage{booktabs}
\usepackage{xcolor}
\usepackage{setspace}
\usepackage{enumitem}

\usepackage{pdfpages}
\usepackage{amsmath}
\usepackage{textcomp}
\usepackage{float} 
\usepackage{booktabs}
\usepackage{longtable}
\usepackage{tabularx}
\usepackage{graphicx}
\usepackage[pagebackref,bookmarks]{hyperref}
\usepackage[most]{tcolorbox}
\tcbuselibrary{skins,breakable,raster}
\usepackage{listings}
\usepackage{xcolor}
\usepackage{subcaption}
\usepackage{placeins}
\usepackage{paracol} 

\onlineid{1030}

\vgtccategory{Research}

\vgtcpapertype{please specify}

\title{Chart-Supported or Model-Supplied? Examining MLLM-Generated Claims for Accessible Visualization}

\author{%
  \authororcid{Ishrat Jahan Eliza}{0000-0003-3087-5951} and 
  \authororcid{Md Dilshadur Rahman}{0009-0008-5467-615X}
}

\authorfooter{
  \item
  	Ishrat Jahan Eliza is with the University of Utah.
  	E-mail: ishratjahan.eliza@utah.edu

  \item Md Dilshadur Rahman is with the University of Utah.
  	E-mail: dilshadur.rahman@utah.edu
}

\abstract{
Multimodal large language models (MLLMs) can connect visualization
patterns to external causes, consequences, and domain knowledge, but
the evidential basis of these interpretations is often unclear. We
present an exploratory study of 102 visualizations from four sources, three MLLMs, and four input conditions that vary access to the image, accessible chart context (non-image artifacts such as data tables, captions, alt text, and screen-reader structures), and withheld-context framing. Across 1,224 descriptions, we analyze model-attributed \textsc{Direct}, \textsc{Derived}, and \textsc{Speculative} labels and conduct an automated audit of numeric agreement. Accessible chart context shifted Gemini and GPT toward \textsc{Direct} claims and improved numeric agreement for some models. Adding the image to the full context did not yield a consistent numeric benefit, and the withheld-context prompt did not reliably increase cautious language. The prompt-defined \emph{Real-World Significance} section remained predominantly \textsc{Speculative}. These results motivate accessible description systems that distinguish claims supported by supplied evidence from model-supplied interpretation.
}


\teaser{
  \centering
 \vspace{-0.1em}
 \includegraphics[width=\linewidth, alt={Overview of the study workflow showing the source visualization and accessible chart context, four input conditions, and descriptions generated by three MLLMs with Direct, Derived, and Speculative labels.}]{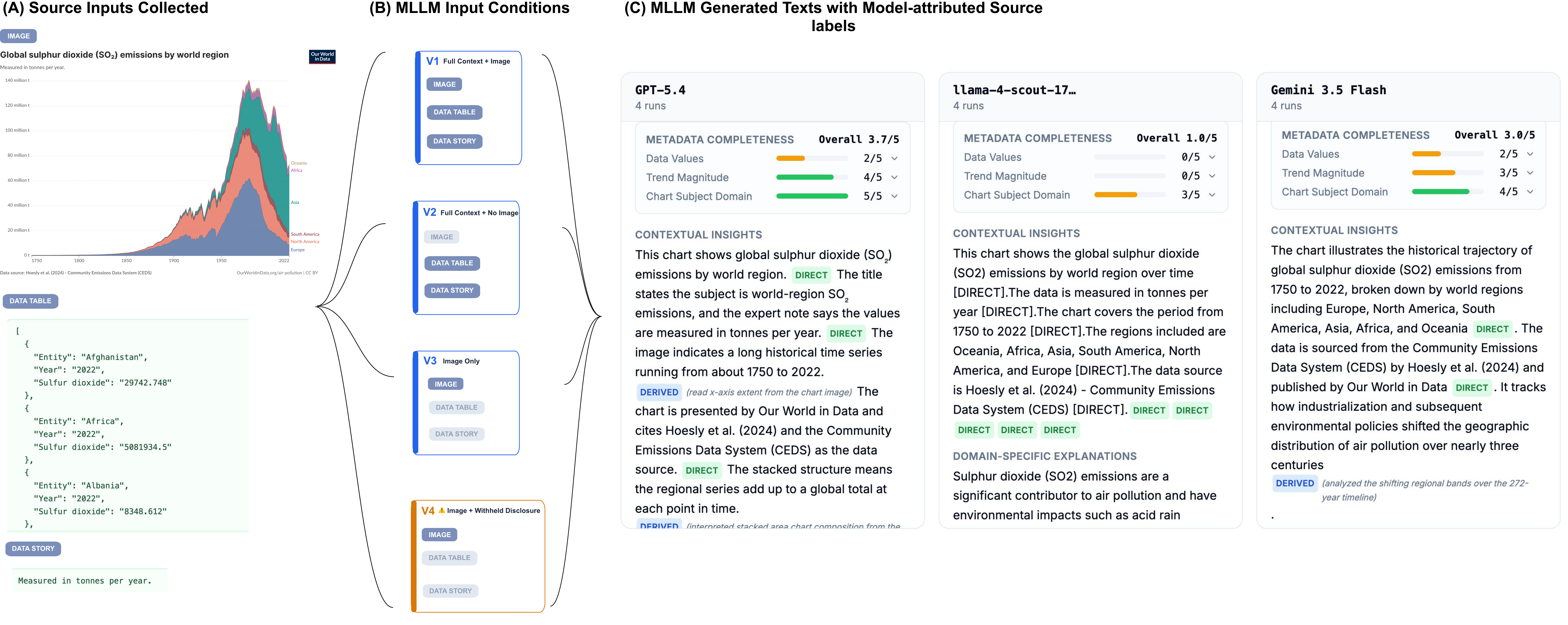}
\caption{Overview of the study design using a visualization of global sulphur dioxide emissions from Our World in Data~\cite{owid, hoesly2024ceds}.
(A) We collect the visualization image and its source-specific accessible chart context, which includes a data table and an expert-authored data story for this example. (B) We construct four input conditions: image with full context (V1), full context without the image (V2), image without accessible chart context (V3), and image with a separately worded prompt stating that accessible chart context exists but was withheld (V4). (C) Each condition is provided to three MLLMs, which assess evidence availability and generate sectioned descriptions with
model-attributed \textsc{Direct}, \textsc{Derived}, and \textsc{Speculative} source labels. The project website provides the visualization corpus, source-specific context, generated descriptions for all models and conditions, and detailed evaluation results at \url{https://shorturl.at/9xbgn}}
  \label{fig:teaser}
  \vspace{-0.6em}
}

\renewcommand{\manuscriptnotetxt}{}

\graphicspath{{figs/}{figures/}{pictures/}{images/}{./}} 

\usepackage{tabu}                      
\usepackage{booktabs}                  
\usepackage{lipsum}                    
\usepackage{mwe}                       

\usepackage{mathptmx}                  

\usepackage{microtype} 
\usepackage{enumitem}  
\setlist{nosep,leftmargin=*,itemsep=1pt,topsep=2pt}

\renewcommand{\arraystretch}{0.95}

\begin{document}



\maketitle

\setstretch{0.958}
\vspace{-0.7em}
\section{Introduction}
\label{sec.intro}

Accessible visualization systems allow blind and low-vision (BLV) readers to inspect values, navigate visualization structure, and explore trends through screen readers, speech, sonification, and braille~\cite{zong2022rich,chartreader,seo2024maidr,zong2024umwelt,blanco2022olli}. Much of this work supports access to visualization structure, data values, and visible patterns, but provides less support for explanations that connect those patterns to external causes, consequences, or domain knowledge. Lundgard and Satyanarayan organize visualization descriptions into four semantic levels: chart construction and encodings (L1), statistical facts and relations (L2), perceptual patterns and trends (L3), and contextual or domain-specific insights (L4)~\cite{lundgard2022accessible}. The distinction between L3 and L4 is central to our study. Identifying a sharp decline in stock prices in 2008 is an L3 observation, whereas linking that decline to the global financial crisis introduces external knowledge at L4. We refer to statements at this level as \emph{domain-contextual claims}. These claims can explain why a pattern occurred or why it matters, but they depend on information beyond the visualization and may introduce unsupported explanations or assumptions.

Multimodal large language models (MLLMs) can generate domain-contextual claims by combining visualization images, structured representations, accompanying text, and general knowledge. Recent systems use these capabilities for accessible visualization interpretation and question answering~\cite{sharma2025beforemom,gonzalez2024scene,gorniak2024vizability,seo2024maidrai,das2026graphwhisper}. We use \emph{accessible chart context} to refer to non-image information associated with a visualization, including data tables, scene graphs, captions, alt text, screen-reader trees, and structured specifications. Combining these sources can make the basis of a generated claim difficult to determine. A claim may restate supplied evidence, derive a relation from that evidence, or introduce information from the model's general knowledge. Prior work has documented factual errors in generated visualization descriptions~\cite{ji2023survey,hwang2023grounded,huang2024chocolate,krichene2024chatspi}. Such errors create a particular verification burden for BLV readers
who may be unable to independently verify the visual evidence~\cite{elavsky2025telephone,alharbi2024misfitting}. We therefore examine how the evidence supplied to an MLLM affects the claims it generates and the source labels it assigns to them.

We conduct an exploratory study using 102 visualizations from four established sources, three MLLMs, and four input conditions. The conditions vary access to the visualization image and accessible chart context. A fourth condition uses a separately worded prompt that states that accessible chart context exists but has been withheld. The models label each claim as \textsc{Direct}, \textsc{Derived}, or \textsc{Speculative}, as described in Section~\ref{sec.two-stage}. These labels are model-attributed accounts of each claim's basis rather than verified provenance. We analyze these source-label distributions and compare a subset of claims with available reference values. Our study characterizes model behavior and does not evaluate the usefulness of the descriptions for BLV readers. We ask the following research questions:

\begin{enumerate}[
    label=\textbf{RQ\arabic*.},
    leftmargin=*,
    itemsep=2pt,
    topsep=3pt,
    parsep=0pt
]
    \item \textbf{Accessible chart context.}
    For the same visualization image, how does adding accessible chart context affect the distribution of model-attributed \textsc{Direct}, \textsc{Derived}, and \textsc{Speculative} labels and numeric agreement within \textsc{Direct} claims?

    \item \textbf{Image contribution.}
    When full accessible chart context is available, how does adding the visualization image affect source-label distributions and numeric agreement within \textsc{Direct} claims?

    \item \textbf{Withheld-context framing.}
    When accessible chart context is absent, how does a separately worded prompt stating that such context exists but was withheld affect source-label distributions, numeric agreement within \textsc{Direct} claims, and indicators of caution?
\end{enumerate}

Our study provides an exploratory evaluation and a claim-level analysis of 1,224 generated descriptions. Adding accessible chart context shifted claims toward \textsc{Direct} labels and improved numeric agreement for some models, while adding the visualization image to the full context produced no consistent additional benefit. The withheld-context prompt affected evidence self-assessments but did not consistently increase cautious generation. Across models and conditions, 73.9\% of claims in the prompt-defined \emph{Real-World Significance} section were labeled \textsc{Speculative}. These findings suggest that accessible visualization description systems should distinguish claims supported by supplied evidence from domain-contextual interpretations introduced by the model.


\vspace{-0.6em}
\section{Background}
\label{sec.background}


\textbf{Semantic models and description evaluation.}
We use Lundgard and Satyanarayan's four-level model to frame our analysis of domain-contextual claims~\cite{lundgard2022accessible}. VisText operationalizes L1--L3 at scale using visualization images, data tables, and scene graphs, but omits L4 because it depends on external knowledge and subjective interpretation~\cite{tang2023vistext}. An analysis of alt text in HCI publications similarly found that only 31\% of descriptions included major trends or comparisons~\cite{chintalapati2022dataset}. Other systems generate descriptions for specific visualization domains. AltGosling derives descriptions from structured genomics visualizations~\cite{smits2024altgosling}, while McNutt et al. generate descriptions for UpSet plots and compare their method with direct LLM prompting~\cite{mcnutt2025upset}. Recent benchmarks assess the factuality, faithfulness, and insightfulness of generated visualization descriptions~\cite{huang2024chocolate,krichene2024chatspi,wang2026chartfi}. Our study instead varies the evidence available to the model and examines the resulting claims and their model-attributed source labels.

\textbf{Accessible visualization representations and authoring context.}
Olli, rich screen-reader experiences, Chart Reader, MAIDR, and Umwelt support structured navigation or multimodal access to visualization data and encodings~\cite{blanco2022olli,zong2022rich,chartreader,seo2024maidr,zong2024umwelt}. Chartability offers heuristics for evaluating such accessibility features~\cite{elavsky2022chartability}. These systems illustrate the types of non-image artifacts included in our accessible chart context. FigurA11y further examines how figure text and surrounding document context can support human review of AI-generated scientific alt text~\cite{singh2024figura11y}. In our study, comparable artifacts serve as controlled model inputs; we do not introduce or evaluate a reading or authoring interface.

\textbf{MLLM-generated descriptions and accessible verification.}
Prior work has examined the questions BLV readers ask about visualizations~\cite{kim2023chartqa}. VizAbility, MAIDR Meets AI, and GraphWhisper investigate conversational access using structured visualization information, multimodal representations, or visualization images~\cite{gorniak2024vizability,seo2024maidrai,das2026graphwhisper}. MAIDR Meets AI examines BLV users' preferences for generated
descriptions and their strategies for verifying model responses~\cite{seo2024maidrai}. Separately, Alharbi et al. document
how blind users identify, cross-check, and contest errors in AI-mediated visual assistance~\cite{alharbi2024misfitting}. Elavsky and Xiong Bearfield describe verification disability and
compelled reliance when visual evidence cannot be independently
verified~\cite{elavsky2025telephone}. Other work suggests that presenting variation across generated descriptions can help users assess their reliability~\cite{chen2025surfacing}. We extend this literature through a controlled comparison in which the visualization remains fixed while the evidence available to the model is supplied, removed, or disclosed as missing. We then examine changes in the generated claims and their model-attributed source labels.

\vspace{-0.3em}
\section{Study Design}
\label{sec.benchmark}

\autoref{fig:teaser} provides an overview of the study workflow, including the collection of each visualization and its source-specific accessible chart context, the construction of the four input conditions, and the generation of claim-labeled descriptions by three MLLMs.

\subsection{Visualization Corpus}
\label{sec.corpus}

We constructed a purposive corpus from four sources that provide different forms of accessible chart context: VisText~\cite{tang2023vistext}, Our World in Data~\cite{owid}, the HCI Alt Text Dataset~\cite{chintalapati2022dataset}, and the Olli Gallery~\cite{blanco2022olli}. Two authors jointly reviewed candidate visualizations and selected examples that varied in visualization type, orientation, and number of series while retaining the source artifacts required to construct the study conditions. The final corpus contains 102 visualizations: 30 from VisText, 42 from Our World in Data, 20 from the HCI Alt Text Dataset, and 10 from the Olli Gallery (Table~\ref{tab:corpus}). The corpus was designed for within-visualization comparisons across different visualization and context types.

\begin{table}[tb]
  \centering
  \caption{Visualization Corpus and source fields used as full accessible chart context. ``Full'' is source-relative because the available fields differ across sources.}
  \label{tab:corpus}
  \footnotesize
  \setlength{\tabcolsep}{3pt}
  \renewcommand{\arraystretch}{0.95}
  \begin{tabular}{@{}p{0.17\linewidth} p{0.29\linewidth} p{0.42\linewidth}@{}}
    \toprule
    \textbf{Source} &
    \textbf{Visualizations ($n$)} &
    \textbf{Fields included in full context} \\
    \midrule

    VisText~\cite{tang2023vistext} &
    30 (10 bar, 10 line, 10 area) &
    Data tables, scene graphs, and human-written L2/L3 captions \\
    \addlinespace[2pt]

    Our World in Data~\cite{owid} &
    42 (9 line, 7 map, 9 bar, 9 area, 8 scatter) &
    Data table and the expert-authored short description or data story
    associated with the selected visualization \\
    \addlinespace[2pt]

    HCI Alt Text Dataset~\cite{chintalapati2022dataset} &
    20 (8 bar, 8 scatter, 4 other) &
    Author-written alt text \\
    \addlinespace[2pt]

    Olli Gallery~\cite{blanco2022olli} &
    10 (multiple visualization types) &
    Olli screen-reader trees and Vega-Lite specifications \\
    \bottomrule
  \end{tabular}
\end{table}

\vspace{-0.3em}
\subsection{Input Conditions}
\label{sec.input-conditions}

We used a within-visualization design in which each visualization was presented to each model under four input conditions. For a given visualization, \emph{full accessible chart context} includes all applicable source fields listed in~\autoref{tab:corpus}. The form and amount of context therefore vary across sources, but each comparison is made between conditions for the same visualization. The conditions vary in access to the visualization image, access to accessible chart context, and prompt framing (\autoref{tab:conditions}). V1 contains the image and full accessible chart context. V2 contains the same context without the image. V3 contains only the image. V4 contains the image without accessible chart context and uses a separately worded prompt stating that such context exists but has been withheld. The conditions support three paired comparisons. RQ1 compares V1 with V3, holding the visualization image constant while adding accessible chart context. RQ2 compares V1 with V2, holding accessible chart context constant while adding the image. RQ3 compares V4 with V3. Both RQ3 conditions omit accessible chart context, but V4 uses a separately worded prompt stating that such context exists and has been withheld. The comparison therefore evaluates the tested withheld-context framing and its associated wording rather than the isolated effect of a disclosure statement. The supplementary material provides detailed examples for each data source.

\begin{table}[tb]
  \centering
  \caption{Evidence and prompt framing used in the four input conditions.}
  \label{tab:conditions}
  \small
  \begin{tabular}{@{}lccc@{}}
    \toprule
    \textbf{Condition} &
    \textbf{Image} &
    \textbf{Context} &
    \textbf{Withheld-context prompt} \\
    \midrule
    V1: context + image & Yes & Full & No \\
    V2: context only    & No  & Full & No \\
    V3: image only      & Yes & None & No \\
    V4: withheld-context framing & Yes & None & Yes \\
    \bottomrule
  \end{tabular}
\end{table}

\subsection{Two-Stage Prompting}
\label{sec.two-stage}

Each call requested a structured JSON response in two stages.

\paragraph{Stage 1: Evidence-availability assessment.}
The model scored the availability of data values, trend magnitude,
and the visualization subject or domain for each input condition. The shared rubric for these three dimensions was:
0 = completely absent;
1 = present but inadequate;
2 = present but incomplete;
3 = present and partially sufficient;
4 = present and mostly sufficient; and
5 = present and fully sufficient. Each scoring was accompanied by a brief justification.

\paragraph{Stage 2: Description generation and source labeling.}
The model generated text in four prompt-defined sections: \emph{Contextual Insights}, \emph{Domain-Specific Explanations},
\emph{Real-World Significance}, and \emph{Other}. These sections
organize the requested output and do not correspond one-to-one with
the L1--L4 semantic levels.

The prompt requested one model-attributed label per claim: \textsc{Direct} for claims presented as explicitly supported by the
input, \textsc{Derived} for claims presented as requiring an analysis, calculation, comparison, or inference, and \textsc{Speculative} for claims presented as relying on information outside the supplied evidence. The verbatim prompt describes some ``directly calculable'' content as \textsc{Direct} while also
assigning calculations to \textsc{Derived}, creating overlap for some claims. Although the prompt required one label per claim, the definitions of
\textsc{Direct} and \textsc{Derived} are not fully non-overlapping. We
therefore analyze the labels as model-selected categories rather than
verified accounts of provenance.





\subsection{Models and Analysis}
\label{sec.models}

We evaluated GPT-5.4, Gemini 3.5 Flash, and Llama 4 Scout 17B Vision, referred to as GPT, Gemini, and Llama. Models were
queried through APIs at temperature 0. The fully crossed design
produced $102 \times 4 \times 3 = 1{,}224$ generations. In 44 records, \texttt{generated\_description} was returned as a non-string JSON value; these fields were serialized to text before applying the
same claim parser used for the remaining records. All 1,224 generations were retained.

For each generated description, we counted the inline
\textsc{Direct}, \textsc{Derived}, and \textsc{Speculative} labels and
calculated each label's proportion of all labeled claims. We also
normalized the total number of labeled claims per 100 generated words
to measure claim density. For each research question, we calculated within-visualization differences between its paired conditions and reported 95\% percentile-bootstrap intervals from
10,000 resamples. Because the corpus was purposively selected, these
intervals describe variability among the visualizations studied.

For 76 visualizations, the source materials contained at least one value-bearing, machine-readable reference representation: a data
table, scene graph, Olli screen-reader tree, or Vega-Lite specification. The other 26 visualizations remained in the
source-label and claim-density analyses but were excluded from the
numeric audit because their source packages did not contain a comparable machine-readable reference.
We extracted numeric tokens only from claims labeled
\textsc{Direct}. A claim containing multiple numbers contributed multiple numeric tokens. A generated numeric token \(y\) was counted as matching a reference
value \(x\) when
\(\lvert y-x\rvert \leq \max(0.5,\,0.02\lvert x\rvert)\). We define \emph{\textsc{Direct}-claim numeric agreement}
as
$N_{\mathrm{matched}} /
(N_{\mathrm{matched}} + N_{\mathrm{unmatched}})$.

This automated measure evaluates agreement between individual numbers
and the available reference information; it does not establish the semantic correctness of the complete claim. In particular, matching a number does not verify its category, series, relationship, or
interpretation, and the number of source-table rows may differ from
the number of marks rendered in a visualization. Panel~B of
Table~\ref{tab:paired_effects} pools numeric tokens across V1--V4. We additionally counted lexical indicators of
uncertainty, abstention, requests for missing information, and causal
language inside \textsc{Speculative} claims. These are heuristic
indicators of caution rather than complete semantic judgments.
\section{Results}
\label{sec.results}

\begin{table*}[tb]
  \centering
  \caption{Panel A reports model-specific, chart-paired changes in
  model-attributed label shares. Each entry is the mean difference in
  percentage points with its 95\% percentile-bootstrap interval.
  Positive values indicate a higher share in the first condition named
  in the contrast. Panel B reports the automated numeric audit pooled
  across V1--V4 for numeric tokens in \textsc{Direct} claims.}
  \label{tab:paired_effects}
  \footnotesize
  \setlength{\tabcolsep}{6.5pt}
  \renewcommand{\arraystretch}{1.14}

  \begin{tabular}{@{}llllll@{}}
    \toprule

    \multicolumn{6}{@{}l}{
      \textbf{A. Chart-paired changes in model-attributed label shares}
    } \\[2pt]

    \textbf{Contrast} &
    \textbf{Model} &
    \shortstack[l]{\textbf{$\Delta$ \textsc{Direct}}\\
                \textbf{share (pp)}} &
    \shortstack[l]{\textbf{$\Delta$ \textsc{Derived}}\\
                \textbf{share (pp)}} &
    \shortstack[l]{\textbf{$\Delta$ \textsc{Speculative}}\\
                \textbf{share (pp)}} & \\ 
    \midrule

    \textbf{RQ1: Context added}
      & Gemini
      & $\mathbf{+10.4}$~{\scriptsize\textcolor{gray}{$[7.5,\,13.3]$}}
      & $\mathbf{-8.0}$~{\scriptsize\textcolor{gray}{$[-11.3,\,-4.8]$}}
      & $\mathbf{-2.4}$~{\scriptsize\textcolor{gray}{$[-4.9,\,0.2]$}} & \\

    \textit{V1 minus V3}
      & GPT
      & $\mathbf{+6.9}$~{\scriptsize\textcolor{gray}{$[4.5,\,9.3]$}}
      & $\mathbf{-0.7}$~{\scriptsize\textcolor{gray}{$[-3.3,\,1.8]$}}
      & $\mathbf{-6.2}$~{\scriptsize\textcolor{gray}{$[-7.9,\,-4.5]$}} & \\

      & Llama
      & $\mathbf{+1.4}$~{\scriptsize\textcolor{gray}{$[-2.1,\,5.1]$}}
      & $\mathbf{-0.1}$~{\scriptsize\textcolor{gray}{$[-3.6,\,3.2]$}}
      & $\mathbf{-1.3}$~{\scriptsize\textcolor{gray}{$[-4.2,\,1.5]$}} & \\

    \addlinespace[3pt]

    \textbf{RQ2: Image added}
      & Gemini
      & $\mathbf{-4.0}$~{\scriptsize\textcolor{gray}{$[-6.5,\,-1.5]$}}
      & $\mathbf{+8.5}$~{\scriptsize\textcolor{gray}{$[5.2,\,11.8]$}}
      & $\mathbf{-4.4}$~{\scriptsize\textcolor{gray}{$[-6.9,\,-2.0]$}} & \\

    \textit{V1 minus V2}
      & GPT
      & $\mathbf{-2.7}$~{\scriptsize\textcolor{gray}{$[-4.9,\,-0.5]$}}
      & $\mathbf{+2.3}$~{\scriptsize\textcolor{gray}{$[0.0,\,4.5]$}}
      & $\mathbf{+0.4}$~{\scriptsize\textcolor{gray}{$[-1.2,\,1.9]$}} & \\

      & Llama
      & $\mathbf{+2.5}$~{\scriptsize\textcolor{gray}{$[-1.1,\,5.9]$}}
      & $\mathbf{+6.4}$~{\scriptsize\textcolor{gray}{$[3.1,\,9.6]$}}
      & $\mathbf{-8.9}$~{\scriptsize\textcolor{gray}{$[-12.0,\,-5.7]$}} & \\

    \addlinespace[3pt]

    \textbf{RQ3: Withheld-context framing}
      & Gemini
      & $\mathbf{+5.5}$~{\scriptsize\textcolor{gray}{$[3.6,\,7.4]$}}
      & $\mathbf{-6.6}$~{\scriptsize\textcolor{gray}{$[-9.1,\,-4.1]$}}
      & $\mathbf{+1.1}$~{\scriptsize\textcolor{gray}{$[-1.0,\,3.2]$}} & \\

    \textit{V4 minus V3}
      & GPT
      & $\mathbf{+3.5}$~{\scriptsize\textcolor{gray}{$[1.2,\,6.0]$}}
      & $\mathbf{-3.2}$~{\scriptsize\textcolor{gray}{$[-5.5,\,-0.9]$}}
      & $\mathbf{-0.4}$~{\scriptsize\textcolor{gray}{$[-2.3,\,1.5]$}} & \\

      & Llama
      & $\mathbf{-4.5}$~{\scriptsize\textcolor{gray}{$[-7.9,\,-1.0]$}}
      & $\mathbf{+1.8}$~{\scriptsize\textcolor{gray}{$[-1.7,\,5.3]$}}
      & $\mathbf{+2.7}$~{\scriptsize\textcolor{gray}{$[-0.3,\,5.7]$}} & \\

    \midrule

    \multicolumn{6}{@{}l}{
      \textbf{B. Automated numeric audit of \textsc{Direct} claims
      pooled across V1--V4}
    } \\[2pt]

    \multicolumn{2}{@{}l}{\textbf{Model}} &
    \textbf{Matched} &
    \textbf{Unmatched} &
    \textbf{$N_{\mathrm{num}}$} &
    \textbf{Agreement} \\
    \midrule

    \multicolumn{2}{@{}l}{Gemini}
      & 872 & 222 & 1,094 & 79.7\% \\

    \multicolumn{2}{@{}l}{GPT}
      & 1,738 & 461 & 2,199 & 79.0\% \\

    \multicolumn{2}{@{}l}{Llama}
      & 654 & 263 & 917 & 71.3\% \\

    \multicolumn{2}{@{}l}{\textbf{Overall}}
      & \textbf{3,264}
      & \textbf{946}
      & \textbf{4,210}
      & \textbf{77.5\%} \\

    \bottomrule
  \end{tabular}

  \vspace{3pt}
  \parbox{0.97\textwidth}{\scriptsize
  \textit{Note.}
  $N_{\mathrm{num}}$ is the number of extracted numeric tokens, and one claim can contribute multiple tokens. The three label shares
  form a composition and therefore sum to 100\%; their changes are not statistically independent and sum to approximately zero,
  subject to rounding. We measure numeric agreement with the available reference values and do not establish the correctness
  of the complete claim.}
\end{table*}

Across the 1,224 outputs, our parser identified 17,755 inline claim labels. \autoref{tab:paired_effects} reports chart-paired changes in the proportions of \textsc{Direct}, \textsc{Derived}, and \textsc{Speculative} claims. A value such as $+10.4\ [7.5,13.3]$ means that the first condition named in the contrast had, on average, a 10.4-percentage-point higher share than the second condition; the bracketed values give the 95\% percentile-bootstrap interval over chart-level differences. An interval containing zero indicates that the direction was not stable under this resampling analysis. These intervals describe the visualizations in our purposive corpus and are not population-level confidence intervals. Across all models and conditions, 73.9\% of claims in the prompt-defined \emph{Real-World Significance} section were labeled
\textsc{Speculative}, compared with 13.8\% in
\emph{Contextual Insights}. At the model-by-condition level, \textsc{Speculative} claims accounted for 58.2--86.4\% of the
\emph{Real-World Significance} section. Thus, most claims in the most interpretive section were labeled as depending on information beyond the supplied evidence, including in V1, the richest source-specific condition.



\paragraph{RQ1: accessible chart context.}
Adding accessible chart context shifted claims toward \textsc{Direct} labels for Gemini and GPT. Relative to V3, the mean \textsc{Direct} proportion in V1 increased by 10.4 percentage points for Gemini (95\% bootstrap interval: 7.5 to 13.3) and 6.9 points for GPT (4.5 to 9.3). Llama's change was smaller and its interval included zero (+1.4 points; $-2.1$ to 5.1). A clear reduction in the \textsc{Speculative} proportion occurred for GPT ($-6.2$ points;
$-7.9$ to $-4.5$), but not consistently across all three models. Claim density likewise showed no common model-wide response.

Accessible chart context also increased chart-paired \textsc{Direct}-claim numeric agreement for GPT (+4.5 percentage points; 0.1 to 9.2) and Llama (+6.6 points; 1.1 to 12.3).
Gemini's change was smaller and uncertain (+1.7 points; $-3.8$ to 6.9). Accessible chart context therefore helped some models identify
and reproduce reference values, but the effect was neither uniform nor sufficient to eliminate \textsc{Speculative} domain-contextual claims.

\paragraph{RQ2: image contribution.}
Adding the image to full accessible chart context produced model-specific changes in claim density and source-label
distributions rather than a common pattern. It also did not consistently improve \textsc{Direct}-claim numeric agreement. The
paired change from V2 to V1 was $-4.8$ percentage points for Gemini
($-9.0$ to $-1.1$), +0.4 for GPT ($-3.9$ to 4.7), and $-0.2$ for Llama ($-4.9$ to 4.9). Thus, when accessible chart context was already
available, adding the image did not provide a reliable additional numeric benefit in this corpus. This finding does not imply that
images are generally unnecessary; it indicates that their contribution depended on the model and measured outcome.

\paragraph{RQ3: withheld-context framing.}
The withheld-context prompt changed source-label assignments without
producing a consistent movement toward caution. Gemini and GPT increased their \textsc{Direct} shares by 5.5 and 3.5 percentage
points while reducing their \textsc{Derived} shares by 6.6 and 3.2 points. Llama moved in the opposite direction: its \textsc{Direct} share decreased by 4.5 points, while the intervals
for its \textsc{Derived} and \textsc{Speculative} changes included
zero. For all three models, the change in \textsc{Speculative} share was uncertain.

\textsc{Direct}-claim numeric agreement changed little between V3 and V4: +1.8 percentage points for Gemini ($-1.1$ to 4.9), $-1.2$ for
GPT ($-4.5$ to 2.4), and +1.3 for Llama ($-2.9$ to 5.4). Stage~1 self-assessments were more responsive to the prompt wording. Mean
overall availability increased from 1.16 to 1.73 for Gemini, from 1.83 to 2.86 for GPT, and from 1.63 to 2.21 for Llama. The withheld-context framing therefore altered the models'
self-assessments of evidence availability without producing a
corresponding increase in caution indicators across models.
\section{Discussion}
\label{sec.discussion}

\subsection{Implications}
\textbf{Accessible chart context helped some models, but not uniformly.}
Adding accessible chart context increased the \textsc{Direct} share for Gemini and GPT and improved the numeric agreement for GPT and Llama. These model-specific changes indicate that tables, captions, scene graphs, screen-reader trees, and related representations can help MLLMs use information associated with a visualization. The pattern was not consistent across all three models. Adding the image, even when full context was already available, also produced no common improvement across the measured outcomes. The contribution of each input, therefore, depends on the model, the available context, and the task.

\textbf{Real-world significance often relied on information outside the supplied input.} Across models and conditions, 73.9\% of claims in the prompt-defined \emph{Real-World Significance} section were labeled \textsc{Speculative}. This pattern remained in V1, where the model received both the image and full accessible chart context. A \textsc{Speculative} label does not mean that a claim is false. It means that the model presented the claim as relying on information not contained in the supplied evidence. Accessible descriptions should make this distinction clear, particularly when interpretations about causes, consequences, or stakeholders are presented alongside statements supported by the visualization. External citations or an explicit indication that supporting evidence is unavailable could help preserve that distinction.

\textbf{The tested withholding prompt did not reliably induce caution.}
The V4 prompt changed the models' Stage~1 assessments of evidence availability, but it did not produce a consistent increase in hedging, abstention, requests for missing information, or \textsc{Speculative} labeling. Because V4 used a separately worded prompt, the comparison concerns the tested withheld-context framing rather than disclosure alone. The result indicates that informing a model about missing context is not, by itself, enough to produce consistently cautious output. Future work should test explicit evidence requirements and abstention
instructions through controlled prompt comparisons.

\textbf{Verification requires stronger links between claims and evidence.}
The source labels record how models characterize the basis of their claims, but they do not establish provenance. Their value in an accessible interface would depend on whether users can inspect the supporting evidence, such as a table row, a caption passage, a scene-graph element, or an external source. One corpus case illustrates why the numeric reference target matters.
In the Olli Palmer Penguins scatterplot of flipper length against body
mass, GPT under V1 and Gemini under V2 reported 342 observations:
151 Adelie, 68 Chinstrap, and 123 Gentoo penguins. These counts match
the marks rendered after excluding two rows with missing encoded
values. The unfiltered source table instead contains 344 rows:
152 Adelie, 68 Chinstrap, and 124 Gentoo penguins. An output can therefore agree with the rendered visualization while appearing incorrect against the unfiltered source table. Numeric evaluation
should account for filtering, missing values, aggregation, and other
transformations between source data and rendered marks; token-level
numeric agreement alone does not establish that a complete claim is
correct.

\subsection{Limitations and Future Directions}

Our study uses a controlled, within-visualization design, but its scope is limited to 102 purposively selected visualizations from four sources, three MLLMs, and one prompt set at temperature~0. The findings therefore apply to the models and conditions examined here. Because accessible chart context varied by source, the study also cannot isolate the contributions of individual artifacts such as tables, captions, scene graphs, alt text, screen-reader trees, and specifications. Future studies should vary these components independently across visualization types and domains.

The source labels support claim-level analysis at scale, but they were assigned by the same models that generated the claims and were not independently validated. They represent the models' characterization of their output rather than verified provenance. RQ3 also compares worded prompts separately, so it evaluates the tested withheld-context framing rather than disclosure alone. Further work should validate the labels against the supplied evidence and examine uncertainty, disclosure, and abstention through more controlled comparisons of prompts.

The numeric audit covers 76 visualizations with machine-readable reference data and evaluates individual numeric tokens rather than complete claims. A matching value may refer to the wrong category, series, or relationship, and source-table values may not correspond directly to rendered marks. More complete evaluation should account for both semantic and rendering-level relationships. Finally, the descriptions and source labels were not evaluated with BLV readers. User studies should examine whether evidence links, citations, source cues, and abstention mechanisms support comprehension and verification without adding unnecessary burden.

\bibliographystyle{abbrv-doi-hyperref}

\bibliography{main}

@inproceedings{blanco2022olli,
  author    = {Blanco, Matthew and Zong, Jonathan and Satyanarayan, Arvind},
  title     = {Olli: An Extensible Visualization Library for Screen Reader Accessibility},
  booktitle = {IEEE VIS Posters},
  year      = {2022},
  url       = {https://vis.csail.mit.edu/pubs/olli/}
}

@inproceedings{chintalapati2022dataset,
  author    = {Chintalapati, Sanjana Shivani and Bragg, Jonathan and Wang, Lucy Lu},
  title     = {A Dataset of Alt Texts from {HCI} Publications: Analyses and Uses Towards Producing More Descriptive Alt Texts of Data Visualizations in Scientific Papers},
  booktitle = {Proceedings of the 24th International ACM SIGACCESS Conference on Computers and Accessibility (ASSETS '22)},
  year      = {2022},
  publisher = {ACM},
  address   = {New York, NY, USA},
  doi       = {10.1145/3517428.3544796}
}

@inproceedings{elavsky2022chartability,
  author    = {Elavsky, Frank and Bennett, Cynthia and Moritz, Dominik},
  title     = {How Accessible Is My Visualization? Evaluating Visualization Accessibility with Chartability},
  booktitle = {Computer Graphics Forum (Proc. EuroVis)},
  volume    = {41},
  number    = {3},
  year      = {2022},
  doi       = {doi: 10.1111/cgf.14522}
}

@inproceedings{elavsky2025telephone,
  author    = {Elavsky, Frank and Xiong Bearfield, Cindy},
  title     = {Playing Telephone with Generative Models: ``Verification Disability,'' ``Compelled Reliance,'' and Accessibility in Data Visualization},
  booktitle = {2025 IEEE Workshop on Accessible Data Visualization (AccessViz)},
  pages     = {14--24},
  year      = {2025},
  publisher = {IEEE},
  doi       = {10.1109/ACCESSVIZ68666.2025.00008}
}

@article{lundgard2022accessible,
  author  = {Lundgard, Alan and Satyanarayan, Arvind},
  title   = {Accessible Visualization via Natural Language Descriptions: A Four-Level Model of Semantic Content},
  journal = {IEEE Transactions on Visualization and Computer Graphics},
  volume  = {28},
  number  = {1},
  pages   = {1073--1083},
  year    = {2022},
  doi     = {10.1109/TVCG.2021.3114770}
}

@inproceedings{seo2024maidr,
  author    = {Seo, JooYoung and Xia, Yilin and Lee, Bongshin and McCurry, Sean and Yam, Yu Jun},
  title     = {{MAIDR}: Making Statistical Visualizations Accessible with Multimodal Data Representation},
  booktitle = {Proceedings of the 2024 CHI Conference on Human Factors in Computing Systems},
  year      = {2024},
  publisher = {ACM},
  address   = {New York, NY, USA},
  pages     = {1--22},
  doi       = {10.1145/3613904.3642730}
}

@inproceedings{sharma2025beforemom,
  author    = {Sharma, Rithvik and others},
  title     = {"Before I Asked My Mom, Now I Ask {ChatGPT}": Visual Privacy Management for Blind and Low Vision People in the Age of Generative {AI}},
  booktitle = {Proceedings of the 27th International ACM SIGACCESS Conference on Computers and Accessibility (ASSETS '25)},
  year      = {2025},
  publisher = {ACM}
}

@inproceedings{tang2023vistext,
  author    = {Tang, Benny J. and Boggust, Angie and Satyanarayan, Arvind},
  title     = {{VisText}: A Benchmark for Semantically Rich Chart Captioning},
  booktitle = {Proceedings of the 61st Annual Meeting of the Association for Computational Linguistics (Volume 1: Long Papers)},
  pages     = {7268--7298},
  year      = {2023},
  publisher = {Association for Computational Linguistics},
  address   = {Toronto, Canada},
  doi       = {10.18653/v1/2023.acl-long.401},
  url       = {https://aclanthology.org/2023.acl-long.401/}
}

@inproceedings{zong2024umwelt,
  author    = {Zong, Jonathan and Pedraza Pineros, Isabella and Chen, Mengzhu and Hajas, Daniel and Satyanarayan, Arvind},
  title     = {Umwelt: Accessible Structured Editing of Multi-Modal Data Representations},
  booktitle = {Proceedings of the 2024 CHI Conference on Human Factors in Computing Systems (CHI '24)},
  year      = {2024},
  publisher = {ACM}
}

@article{mcnutt2025upset,
  author  = {McNutt, Andrew and McCracken, Maggie K. and Eliza, Ishrat Jahan and Hajas, Daniel and Wagoner, Jake and Lanza, Nate and Wilburn, Jack and Creem-Regehr, Sarah and Lex, Alexander},
  title   = {Accessible Text Descriptions for {UpSet} Plots},
  journal = {Computer Graphics Forum},
  volume  = {44},
  number  = {3},
  pages   = {e70102},
  year    = {2025},
  doi     = {10.1111/cgf.70102}
}

@article{zong2022rich,
  author  = {Zong, Jonathan and Lee, Crystal and Lundgard, Alan and Jang, JiWoong and Hajas, Daniel and Satyanarayan, Arvind},
  title   = {Rich Screen Reader Experiences for Accessible Data Visualization},
  journal = {Computer Graphics Forum},
  volume  = {41},
  number  = {3},
  pages   = {15--27},
  year    = {2022},
  doi     = {10.1111/cgf.14519}
}

@article{ji2023survey,
  author    = {Ji, Ziwei and Lee, Nayeon and Frieske, Rita and Yu, Tiezheng and Su, Dan and Xu, Yan and Ishii, Etsuko and Bang, Ye Jin and Madotto, Andrea and Fung, Pascale},
  title     = {Survey of Hallucination in Natural Language Generation},
  journal   = {ACM Computing Surveys},
  volume    = {55},
  number    = {12},
  articleno = {248},
  numpages  = {38},
  year      = {2023},
  doi       = {10.1145/3571730}
}

@inproceedings{gonzalez2024scene,
  author    = {Gonzalez Penuela, Ricardo E. and Collins, Jazmin and Bennett, Cynthia and Azenkot, Shiri},
  title     = {Investigating Use Cases of {AI}-Powered Scene Description Applications for Blind and Low Vision People},
  booktitle = {Proceedings of the 2024 CHI Conference on Human Factors in Computing Systems (CHI '24)},
  year      = {2024},
  publisher = {ACM},
  address   = {New York, NY, USA},
  articleno = {901},
  doi       = {10.1145/3613904.3642211}
}

@article{hwang2023grounded,
  author  = {Hwang, Alyssa and Head, Andrew and Callison-Burch, Chris},
  title   = {Grounded Intuition of {GPT}-Vision's Abilities with Scientific Images},
  journal = {arXiv preprint arXiv:2311.02069},
  year    = {2023},
  url     = {https://arxiv.org/abs/2311.02069}
}

@inproceedings{chen2025surfacing,
  author    = {Chen, Meng and Iyer, Akhil and Pavel, Amy},
  title     = {Surfacing Variations to Calibrate Perceived Reliability of {MLLM}-Generated Image Descriptions},
  booktitle = {Proceedings of the 27th International ACM SIGACCESS Conference on Computers and Accessibility (ASSETS '25)},
  year      = {2025},
  publisher = {ACM},
  address   = {New York, NY, USA},
  doi       = {10.1145/3663547.3746393}
}

@inproceedings{chartreader,
author = {Thompson, John R and Martinez, Jesse J and Sarikaya, Alper and Cutrell, Edward and Lee, Bongshin},
title = {Chart Reader: Accessible Visualization Experiences Designed with Screen Reader Users},
year = {2023},
isbn = {9781450394215},
publisher = {Association for Computing Machinery},
address = {New York, NY, USA},
url = {https://doi.org/10.1145/3544548.3581186},
doi = {10.1145/3544548.3581186},
booktitle = {Proceedings of the 2023 CHI Conference on Human Factors in Computing Systems},
articleno = {802},
numpages = {18},
location = {Hamburg, Germany},
series = {CHI '23}
}

@inproceedings{seo2024maidrai,
  author    = {Seo, JooYoung and Kamath, Sanchita S. and Zeidieh, Aziz and Venkatesh, Saairam and McCurry, Sean},
  title     = {{MAIDR} Meets {AI}: Exploring Multimodal {LLM}-Based Data Visualization Interpretation by and with Blind and Low-Vision Users},
  booktitle = {Proceedings of the 26th International ACM SIGACCESS Conference on Computers and Accessibility},
  year      = {2024},
  publisher = {ACM},
  address   = {New York, NY, USA},
  pages     = {1--31},
  doi       = {10.1145/3663548.3675660}
}

@inproceedings{gorniak2024vizability,
  author    = {Gorniak, Joshua and Kim, Yoon and Wei, Donglai and Kim, Nam Wook},
  title     = {{VizAbility}: Enhancing Chart Accessibility with {LLM}-Based Conversational Interaction},
  booktitle = {Proceedings of the 37th Annual ACM Symposium on User Interface Software and Technology},
  year      = {2024},
  publisher = {ACM},
  address   = {New York, NY, USA},
  pages     = {1--19},
  doi       = {10.1145/3654777.3676414}
}

@inproceedings{das2026graphwhisper,
  author    = {Das, Amit Kumar and Tarun, Mohammad and Mueller, Klaus},
  title     = {Making Charts Speak: {LLM}-Based Conversational Chart Question Answering for Blind and Low-Vision Users},
  booktitle = {Extended Abstracts of the 2026 CHI Conference on Human Factors in Computing Systems},
  year      = {2026},
  publisher = {ACM},
  address   = {New York, NY, USA},
  pages     = {1--5},
  doi       = {10.1145/3772363.3799030}
}

@inproceedings{kim2023chartqa,
  author    = {Kim, Jiho and Srinivasan, Arjun and Kim, Nam Wook and Kim, Yea-Seul},
  title     = {Exploring Chart Question Answering for Blind and Low Vision Users},
  booktitle = {Proceedings of the 2023 CHI Conference on Human Factors in Computing Systems},
  year      = {2023},
  publisher = {ACM},
  address   = {New York, NY, USA},
  pages     = {1--15},
  doi       = {10.1145/3544548.3581532}
}

@inproceedings{singh2024figura11y,
  author    = {Singh, Nikhil and Wang, Lucy Lu and Bragg, Jonathan},
  title     = {{FigurA11y}: {AI} Assistance for Writing Scientific Alt Text},
  booktitle = {Proceedings of the 29th International Conference on Intelligent User Interfaces},
  year      = {2024},
  publisher = {ACM},
  address   = {New York, NY, USA},
  pages     = {1--21},
  doi       = {10.1145/3640543.3645212}
}

@article{smits2024altgosling,
  author  = {Smits, Thomas C. and L'Yi, Sehi and Mar, Andrew P. and Gehlenborg, Nils},
  title   = {{AltGosling}: Automatic Generation of Text Descriptions for Accessible Genomics Data Visualization},
  journal = {Bioinformatics},
  volume  = {40},
  number  = {12},
  pages   = {btae670},
  year    = {2024},
  doi     = {10.1093/bioinformatics/btae670}
}

@inproceedings{huang2024chocolate,
  author    = {Huang, Kung-Hsiang and Zhou, Mingyang and Chan, Hou Pong and Fung, Yi and Wang, Zhenhailong and Zhang, Lingyu and Chang, Shih-Fu and Ji, Heng},
  title     = {Do {LVLM}s Understand Charts? Analyzing and Correcting Factual Errors in Chart Captioning},
  booktitle = {Findings of the Association for Computational Linguistics: ACL 2024},
  pages     = {730--749},
  year      = {2024},
  publisher = {Association for Computational Linguistics},
  address   = {Bangkok, Thailand},
  doi       = {10.18653/v1/2024.findings-acl.41},
  url       = {https://aclanthology.org/2024.findings-acl.41/}
}

@inproceedings{krichene2024chatspi,
  author    = {Krichene, Syrine and Piccinno, Francesco and Liu, Fangyu and Eisenschlos, Julian Martin},
  title     = {Faithful Chart Summarization with {ChaTS-Pi}},
  booktitle = {Proceedings of the 62nd Annual Meeting of the Association for Computational Linguistics (Volume 1: Long Papers)},
  pages     = {8705--8723},
  year      = {2024},
  publisher = {Association for Computational Linguistics},
  address   = {Bangkok, Thailand},
  doi       = {10.18653/v1/2024.acl-long.472},
  url       = {https://aclanthology.org/2024.acl-long.472/}
}

@article{wang2026chartfi,
  author  = {Wang, Fen and Shao, Zekai and Kang, Qiman and Hu, Chunran and Zhang, Zhixuan and Xie, Lexu and Liu, Chao and Chen, Siming},
  title   = {{ChartFI}: Benchmarking Faithfulness and Insightfulness of Chart Descriptions from Multimodal Large Language Models},
  journal = {arXiv preprint arXiv:2605.23694},
  year    = {2026},
  url     = {https://arxiv.org/abs/2605.23694}
}

@inproceedings{alharbi2024misfitting,
  author    = {Alharbi, Rahaf and Lor, Pa and Herskovitz, Jaylin and Schoenebeck, Sarita and Brewer, Robin N.},
  title     = {Misfitting With {AI}: How Blind People Verify and Contest {AI} Errors},
  booktitle = {Proceedings of the 26th International ACM SIGACCESS Conference on Computers and Accessibility},
  year      = {2024},
  publisher = {ACM},
  address   = {New York, NY, USA},
  pages     = {1--17},
  doi       = {10.1145/3663548.3675659}
}

@misc{owid,
  author       = {{Our World in Data}},
  title        = {Our World in Data},
  year         = {2026},
  publisher    = {Global Change Data Lab},
  url          = {https://ourworldindata.org/},
  note         = {Accessed July 21, 2026. Individual visualization
                  pages and underlying data sources are listed in the
                  supplemental corpus manifest.}
}

@misc{hoesly2024ceds,
  author    = {Hoesly, Rachel and Smith, Steven J. and others},
  title     = {{CEDS v\_2024\_07\_08 Release Emission Data}},
  year      = {2024},
  publisher = {Zenodo},
  doi       = {10.5281/zenodo.12803197}
}



\captionsetup[subfigure]{
  font=small,
  labelfont=bf,
  justification=centering
}

\definecolor{promptbg}{RGB}{245,245,245}
\definecolor{promptframe}{RGB}{180,180,180}
\definecolor{codebg}{RGB}{250,250,250}

\newtcolorbox{promptbox}[1][]{
    colback=promptbg,
    colframe=promptframe,
    boxrule=0.5pt,
    arc=2pt,
    left=5pt,
    right=5pt,
    top=2pt,
    bottom=2pt,
    before skip=2pt,
    after skip=2pt,
    fontupper=\small\ttfamily,
    #1
}

\newtcolorbox{examplebox}[1][]{
    colback=codebg,
    colframe=promptframe,
    boxrule=0.5pt,
    arc=2pt,
    left=5pt,
    right=5pt,
    top=5pt,
    bottom=5pt,
    fontupper=\small\ttfamily,
    #1
}

\lstdefinelanguage{json}{
    basicstyle=\small\ttfamily,
    numbers=none,
    showstringspaces=false,
    breaklines=true,
    backgroundcolor=\color{codebg},
    stringstyle=\color{blue}\ttfamily,
    morestring=[b]",
    morestring=[d]'
}

\lstset{
  extendedchars=true,
  inputencoding=utf8,
  literate={°}{{\textdegree}}1
           {≈}{{$\approx$}}1
           {×}{{$\times$}}1
           {Σ}{{$\Sigma$}}1
           {²}{{$^2$}}1
}
\definecolor{artifactbackground}{RGB}{248,248,248}
\definecolor{artifactframe}{RGB}{150,150,150}

\lstdefinestyle{accessvizartifact}{
  basicstyle=\ttfamily\footnotesize,
  numbers=none,
  showstringspaces=false,
  breaklines=true,
  breakatwhitespace=false,
  columns=fullflexible,
  keepspaces=true,
  upquote=true,
  frame=none,
  backgroundcolor=\color{artifactbackground},
  literate=
    {°}{{\textdegree}}1
    {≈}{{$\approx$}}1
    {×}{{$\times$}}1
    {Σ}{{$\Sigma$}}1
    {²}{{$^2$}}1
}

\tcbset{
  artifactbox/.style={
    enhanced,
    breakable,
    colback=artifactbackground,
    colframe=artifactframe,
    boxrule=0.4pt,
    arc=1pt,
    left=4pt,
    right=4pt,
    top=3pt,
    bottom=3pt,
    before skip=5pt,
    after skip=5pt,
    fonttitle=\bfseries\small
  }
}

\newcommand{\ShowArtifact}[2]{%
  \begin{tcolorbox}[artifactbox,title={#1}]
    \lstinputlisting[style=accessvizartifact]{#2}
  \end{tcolorbox}
}

\definecolor{directbg}{RGB}{220, 247, 227}
\definecolor{directfg}{RGB}{22, 101, 52}
\definecolor{derivedbg}{RGB}{219, 234, 254}
\definecolor{derivedfg}{RGB}{30, 64, 175}
\definecolor{specbg}{RGB}{254, 243, 199}
\definecolor{specfg}{RGB}{146, 64, 14}
\definecolor{badgebg}{RGB}{237, 233, 254}
\definecolor{badgefg}{RGB}{109, 40, 217}
\definecolor{cardborder}{RGB}{229, 231, 235}

\newcommand{\tagdirect}{\tcbox[on line, boxsep=0pt, left=2pt, right=2pt, top=1pt, bottom=1pt, colback=directbg, colframe=directbg, arc=2pt, boxrule=0pt]{\sffamily\tiny\bfseries\color{directfg}DIRECT}}

\newcommand{\tagderived}[1]{\tcbox[on line, tcbox width=auto limited, boxsep=0pt, left=2pt, right=2pt, top=1pt, bottom=1pt, colback=derivedbg, colframe=derivedbg, arc=2pt, boxrule=0pt]{\sffamily\tiny\bfseries\color{derivedfg}DERIVED \normalfont\tiny (#1)}}

\newcommand{\tagspeculative}[1]{\tcbox[on line, tcbox width=auto limited, boxsep=0pt, left=2pt, right=2pt, top=1pt, bottom=1pt, colback=specbg, colframe=specbg, arc=2pt, boxrule=0pt]{\sffamily\tiny\bfseries\color{specfg}SPECULATIVE \normalfont\tiny (#1)}}

\newcommand{\fbadge}[1]{\tcbox[on line, boxsep=0pt, left=2pt, right=2pt, top=1pt, bottom=1pt, colback=badgebg, colframe=badgebg, arc=3pt, boxrule=0pt]{\sffamily\tiny\color{badgefg}#1}\ }
\newcommand{\imgbadge}{\tcbox[on line, boxsep=0pt, left=2pt, right=2pt, top=1pt, bottom=1pt, colback=green!15, colframe=green!15, arc=3pt, boxrule=0pt]{\sffamily\tiny\color{green!60!black}+image}\ }

\tcbset{
  modelcard/.style={
    enhanced,
    breakable, 
    colback=white,
    colframe=cardborder,
    boxrule=0.8pt,
    arc=4pt,
    left=3pt, right=3pt, top=4pt, bottom=4pt,
    fonttitle=\bfseries\sffamily\small,
    coltitle=black,
    colbacktitle=gray!10,
    titlerule=0.3pt
  }
}

\setlength{\columnsep}{1.2em}

\newcommand{\ShowVariantThreeColumnsFormatted}[2]{%
  \vspace{1em}
  \subsubsection{#1}%
  \vspace{0.5em}
  \begin{paracol}{3}
    
    \begin{tcolorbox}[modelcard, title={GPT-5.4 \hfill \normalfont\tiny\color{gray}1 runs}]
      \fbadge{V1}\fbadge{scene\_graph}\fbadge{data\_table}\fbadge{vega\_lite\_spec}\fbadge{l1\_caption}\fbadge{l2\_caption}\imgbadge
      \vspace{3pt}\hrule\vspace{3pt}
      {\sffamily\tiny\bfseries\color{gray!80!black}METADATA COMPLETENESS \hfill Overall 3.7/5}\\[2pt]
      {\sffamily\tiny Data Values \hfill 4/5}\\
      {\sffamily\tiny Trend Magnitude \hfill 4/5}\\
      {\sffamily\tiny Chart Subject Domain \hfill 3/5}
      \vspace{3pt}\hrule\vspace{3pt}
      \input{#2/gpt-5.4.txt}
    \end{tcolorbox}
    
    \switchcolumn
    \begin{tcolorbox}[modelcard, title={llama-4-scout-17... \hfill \normalfont\tiny\color{gray}1 runs}]
      \fbadge{V1}\fbadge{scene\_graph}\fbadge{vega\_lite\_spec}\fbadge{data\_table}\fbadge{l1\_caption}\fbadge{l2\_caption}\imgbadge
      \vspace{3pt}\hrule\vspace{3pt}
      {\sffamily\tiny\bfseries\color{gray!80!black}METADATA COMPLETENESS \hfill Overall 4.7/5}\\[2pt]
      {\sffamily\tiny Data Values \hfill 5/5}\\
      {\sffamily\tiny Trend Magnitude \hfill 4/5}\\
      {\sffamily\tiny Chart Subject Domain \hfill 5/5}
      \vspace{3pt}\hrule\vspace{3pt}
      \input{#2/llama-4-scout.txt}
    \end{tcolorbox}
    
    \switchcolumn
    \begin{tcolorbox}[modelcard, title={Gemini 3.5 Flash \hfill \normalfont\tiny\color{gray}1 runs}]
      \fbadge{V1}\fbadge{scene\_graph}\fbadge{vega\_lite\_spec}\fbadge{data\_table}\fbadge{l1\_caption}\fbadge{l2\_caption}\imgbadge
      \vspace{3pt}\hrule\vspace{3pt}
      {\sffamily\tiny\bfseries\color{gray!80!black}METADATA COMPLETENESS \hfill Overall 4.0/5}\\[2pt]
      {\sffamily\tiny Data Values \hfill 5/5}\\
      {\sffamily\tiny Trend Magnitude \hfill 2/5}\\
      {\sffamily\tiny Chart Subject Domain \hfill 5/5}
      \vspace{3pt}\hrule\vspace{3pt}
      \input{#2/gemini-3.5-flash.txt}
    \end{tcolorbox}
    
  \end{paracol}
}

\title{{\small Supplementary Material, Submission ID: 1030 }}


\onecolumn

\setcounter{figure}{0}
\renewcommand{\thefigure}{S\arabic{figure}}

\maketitle

\section{Outputs of the Experiments}

The experiment involves 102 visualizations,
the four input conditions, and all 1,224 outputs produced by the three models reported in the paper, can be found here: \href{https://shorturl.at/9xbgn}
{https://shorturl.at/9xbgn}.

\section{LLM Prompts}
This section provides the complete set of system prompts used in the study.

\noindent\textbf{System Prompt:}

\begin{promptbox}
 You are a helpful assistant for blind and low-vision users to generate contextual and real world significance of the data in the chart. You must complete TWO stages before generating any description:

    STAGE 1 -- IDENTIFICATION:
    Score what you have been given.

    STAGE 2 -- GENERATION:
    Generate contextual and domain specific insights based on your knowledge and the provided metadata. Organize your output into the following categories:
    
    - Contextual Insights
    
    - Domain-Specific Explanations
    
    - Real-world significance: Social/Political/other Context
    
    - Other

    FLAG EACH CLAIM inline using these markers.
    Place the flag immediately after the flagged claim in square brackets.
    Apply one flag per claim -- choose the most accurate:

      [DIRECT] -- verbatim or directly calculable from the provided metadata, no interpretation required.

      [DERIVED: \textless{}what was computed or read\textgreater{}] -- an analysis you performed on the provided metadata or chart image: a calculation, a comparison across rows, a trend read off the image.
      State what you did in the bracket.
                
      [SPECULATIVE: \textless{}knowledge domain or source\textgreater{}] -- drawn from your general knowledge, not from anything provided. Name the domain or source you're drawing on in the bracket.

    Do not invent a claim and label it [DIRECT] or [DERIVED] because it feels plausible -- if it isn't actually traceable to the provided metadata/image, it belongs under [SPECULATIVE] with its source named.
\end{promptbox}

\subsection{V1--V4 prompts}

Variants 1/2/3 are a classification computed from user-checked-in accessibility contexts in the UI, defined as available contexts. Only Variant 4 (the metacognition test, where metadata is falsely claimed to be absent) is a genuinely distinct condition that maintains its own template.

\noindent\textbf{V1 -- V3:}

\begin{promptbox}
      ----------
      CHART DESCRIPTION TASK -- \textdollar\textless{}\textless{}condition\_label\textgreater{}\textgreater{}\textdollar
      ----------------

      You have been provided \textdollar\textdollar\textless{}\textless{}image\_clause\textgreater{}\textgreater{}\textdollar\textdollar{} and the accessibility metadata
      below. Every field that exists for this chart and was selected for this
      run is included; everything else is marked [NOT PROVIDED] -- treat those
      as genuinely absent, not as something to guess at.
      
      -------------
      \textless{}\textless{}chart\_image\_block\textgreater{}\textgreater{}
      -------------

      CHART TITLE:
      "\textless{}\textless{}chart\_title\textgreater{}\textgreater{}"

      AXIS LABELS:
        X-axis: "\textless{}\textless{}x\_axis\_label\textgreater{}\textgreater{}"
        Y-axis: "\textless{}\textless{}y\_axis\_label\textgreater{}\textgreater{}"

      DATA TABLE (exact values underlying the chart):
      \textless{}\textless{}data\_table\_formatted\textgreater{}\textgreater{}

      HUMAN-WRITTEN ALT TEXT:
      "\textless{}\textless{}alt\_text\textgreater{}\textgreater{}"
      [Semantic level of this alt text: \textless{}\textless{}alt\_text\_level\textgreater{}\textgreater{} --
       indicates whether it describes construction only, statistics,
       or trends and patterns]

      ARIA LABEL:
      "\textless{}\textless{}aria\_label\textgreater{}\textgreater{}"

      LINKED DESCRIPTION (aria-describedby):
      "\textless{}\textless{}aria\_describedby\_content\textgreater{}\textgreater{}"

      OLLI SCREEN READER TREE (hierarchical ARIA navigation):
      \textless{}\textless{}olli\_tree\textgreater{}\textgreater{}

      SCENE GRAPH (structural chart representation):
      \textless{}\textless{}scene\_graph\textgreater{}\textgreater{}

      L1 CAPTION (construction -- axis ranges, type, encodings):
      "\textless{}\textless{}l1\_caption\textgreater{}\textgreater{}"

      L2 CAPTION (statistics and data relations):
      "\textless{}\textless{}l2\_caption\textgreater{}\textgreater{}"

      L3 CAPTION (perceptual trends and patterns):
      "\textless{}\textless{}l3\_caption\textgreater{}\textgreater{}"

      EXPERT DOMAIN STORY (written by domain expert):
      "\textless{}\textless{}domain\_context\textgreater{}\textgreater{}"
      [Source: \textless{}\textless{}source\_url\textgreater{}\textgreater{}]

      STAKEHOLDER CONTEXT (who uses this data and why):
      "\textless{}\textless{}stakeholder\_context\textgreater{}\textgreater{}"

      -------------------------------------------------------
      STAGE 1 -- IDENTIFICATION
      -------------------------------------------------------

      Score each dimension 0--5 where:
      
      0 = completely absent
      1 = present but inadequate
      2 = present but incomplete
      3 = present and partially sufficient
      4 = present and mostly sufficient
      5 = present and fully sufficient

      -------------------------------------------------------
      STAGE 2 -- GENERATE SEMANTIC DESCRIPTION
      -------------------------------------------------------

      Do NOT describe visual appearance (colors, bar heights, line thickness).
      A blind user does not need to know the bars are blue. They need to know
      what the data means.

      OUTPUT FORMAT -- respond only with valid JSON:

      \{
        "variant": \textless{}\textless{}variant\_number\textgreater{}\textgreater{},
        
        "identification": \{
          "metadata\_completeness": \{
            "data\_values": \{"score": 0, "justification": "..."\},
            "trend\_magnitude": \{"score": 0, "justification": "..."\},
            "chart\_subject\_domain": \{"score": 0, "justification": "..."\},
            "overall": 0
          \}
        \},
        
        "generated\_description": \{
          "contextual\_insights": "What this chart is, what domain it belongs to, who produced it and for what purpose. Flag each claim [DIRECT], [DERIVED: ...], or [SPECULATIVE: ...].",
          "domain\_specific\_explanations": "How the data relates to the domain, what the values mean in this field, what background knowledge is needed to interpret it correctly. Flag each claim.",
          "real\_world\_significance": "Why this data matters. Who is affected. What decisions depend on it. What happens if this trend continues. Every claim grounded in provided domain story and stakeholder context. Flag each claim.",
          "other": "Any additional observations not captured above. Flag each claim."
        \},

        "flags": \{
          "direct\_claims":     ["list each direct claim text"],
          "derived\_claims":    ["list each derived claim text, with what it was derived from"],
          "speculative\_claims": ["list each speculative claim text, with its named source"],
          "total\_flag\_count": 0,
          "direct\_count":      0,
          "derived\_count":      0,
          "speculative\_count": 0
        \}
      \}
\end{promptbox}

\noindent\textbf{V4:}

\begin{promptbox}
      Score each dimension 0--5 where:
      
      0 = completely absent
      1 = present but inadequate
      2 = present but incomplete
      3 = present and partially sufficient
      4 = present and mostly sufficient
      5 = present and fully sufficient

      -------------------------------------------------------
      STAGE 2 -- GENERATE SEMANTIC DESCRIPTION
      -------------------------------------------------------

      VARIANT 4: WITHHELD METADATA CONDITION (METACOGNITION TEST)

      You have been provided with a chart image ONLY.

      IMPORTANT: For this chart, accessibility metadata EXISTS -- but it has been
      DELIBERATELY WITHHELD from you for this evaluation. Do not assume you already know it; do not guess its content.

      -------------------------------------------------------
      CHART IMAGE: [attached]
      -------------------------------------------------------
      
      -------------------------------------------------------
      STAGE 1 -- IDENTIFICATION
      -------------------------------------------------------

      OUTPUT FORMAT -- respond only with valid JSON:

      \{
        "variant": 4,
        
        "identification": \{
          "metadata\_completeness": \{
            "data\_values": \{"score": 0, "justification": "..."\},
            "trend\_magnitude": \{"score": 0, "justification": "..."\},
            "chart\_subject\_domain": \{"score": 0, "justification": "..."\},
            "overall": 0
          \}
        \},
        "generated\_description": \{
          "contextual\_insights": "What this chart is, what domain it belongs to, who produced it and for what purpose. Flag each claim [DIRECT], [DERIVED: ...], or [SPECULATIVE: ...].",
          "domain\_specific\_explanations": "How the data relates to the domain, what the values mean in this field, what background knowledge is needed to interpret it correctly. Flag each claim.",
          "real\_world\_significance": "Why this data matters. Who is affected. What decisions depend on it. What happens if this trend continues. Every claim grounded in provided domain story and stakeholder context. Flag each claim.",
          "other": "Any additional observations not captured above. Flag each claim."
        \},

        "flags": \{
          "direct\_claims":     ["list each direct claim text"],
          "derived\_claims":    ["list each derived claim text, with what it was derived from"],
          "speculative\_claims": ["list each speculative claim text, with its named source"],
          "total\_flag\_count": 0,
          "direct\_count":      0,
          "derived\_count":      0,
          "speculative\_count": 0
        \}
      \}
\end{promptbox}

\section{Source-Specific Condition Examples}
\label{supp:examples}

The following examples show how source-specific accessible chart context was instantiated under each condition. To avoid repeating the shared system prompt, each input box contains the chart-specific user payload sent to the model.

\clearpage
\subsection{VisText Example}
\label{supp:vistext-example}

\begin{figure}[H]
  \centering
  \begin{minipage}[c]{0.48\linewidth}
    \centering
    \includegraphics[
      width=\linewidth,
      alt={Representative VisText chart used across V1 through V4. The example pairs a bar chart with accessible chart context that includes a data table, scene graph, and L2/L3 captions.}
    ]{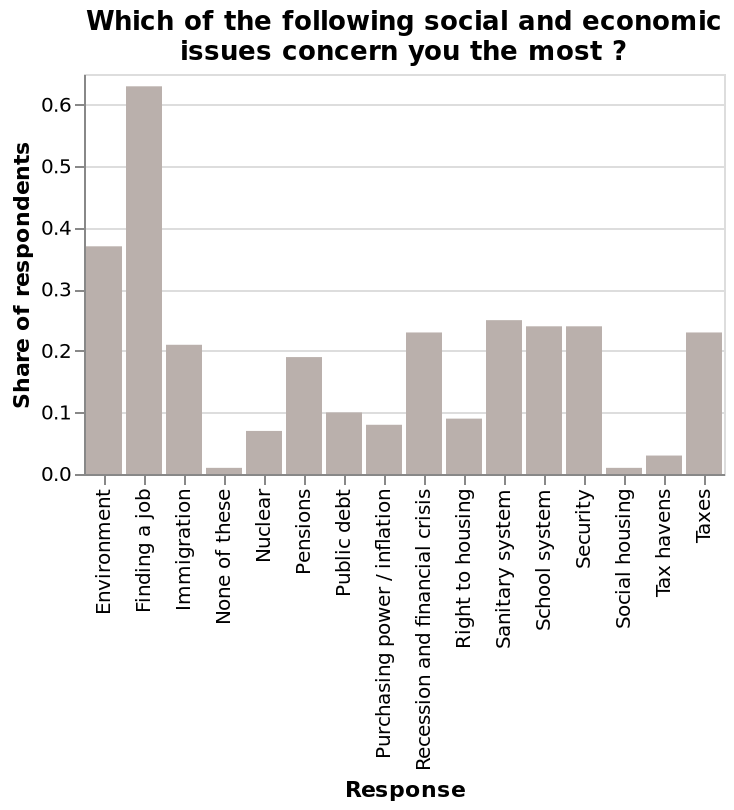}
    \caption{Representative VisText visualization, \texttt{vistext\_2}. The same visualization is used in V1--V4.}
    \label{fig:vistext-example}
  \end{minipage}%
  \hfill
  \begin{minipage}[c]{0.48\linewidth}
    \centering
    \small
    \renewcommand{\arraystretch}{1.2}
    \begin{tabularx}{\linewidth}{@{}lX@{}}
    \toprule
    \textbf{Field} & \textbf{Content used in this example} \\
    \midrule
    Image & Included in V1, V3, and V4 \\
    Data table & Included in V1 and V2 \\
    Scene graph & Included in V1 and V2 \\
    L2/L3 captions & Included in V1 and V2 \\
    \bottomrule
    \end{tabularx}
    \vspace{4pt}
    \captionof{table}{Accessibility metadata fields available across input variants for the VisText example.}
    \label{tab:vistext-fields}
  \end{minipage}
\end{figure}

Figure~\ref{fig:vistext-example} illustrates a representative visualization from the VisText dataset, alongside Table~\ref{tab:vistext-fields}, which details the specific accessibility metadata fields provided across variants V1--V4.

  \vspace{1em}
  \subsubsection{VisText Data V1 Outputs}%
  \vspace{0.5em}
  \begin{paracol}{3}
    
    \begin{tcolorbox}[modelcard, title={GPT-5.4 \hfill \normalfont\tiny\color{gray}1 runs}]
      \fbadge{V1}\fbadge{scene\_graph}\fbadge{data\_table}\fbadge{vega\_lite\_spec}\fbadge{l1\_caption}\fbadge{l2\_caption}\imgbadge
      \vspace{3pt}\hrule\vspace{3pt}
      {\sffamily\tiny\bfseries\color{gray!80!black}METADATA COMPLETENESS \hfill Overall 3.7/5}\\[2pt]
      {\sffamily\tiny Data Values \hfill 4/5}\\
      {\sffamily\tiny Trend Magnitude \hfill 4/5}\\
      {\sffamily\tiny Chart Subject Domain \hfill 3/5}
      \vspace{3pt}\hrule\vspace{3pt}
      {\sffamily\tiny
\noindent{\sffamily\tiny\bfseries\color{gray!90!black}CONTEXTUAL INSIGHTS}\\[2pt]
This chart compares survey response categories on the question, "Which of the following social and economic issues concern you the most?" \tagdirect{} The y-axis measures share of respondents. \tagdirect{} The topic belongs to public opinion research focused on social and economic priorities. \tagspeculative{survey research / political science} The largest reported concern is finding a job at 0.63. \tagderived{read highest value from data table and chart labels} The next highest concern is the environment at 0.37. \tagderived{read second-highest value from data table and chart labels} A middle tier of concerns clusters around roughly 0.21 to 0.25, including sanitary system, school system, security, taxes, recession and financial crisis, and immigration. \tagderived{grouped categories by reading values 0.21 to 0.25 from table and chart labels} Several concerns receive very low shares, including none of these at 0.01, social housing at 0.01, and tax havens at 0.03. \tagderived{read lowest values from data table and chart labels}

\vspace{4pt}\noindent{\sffamily\tiny\bfseries\color{gray!90!black}DOMAIN-SPECIFIC EXPLANATIONS}\\[2pt]
Because the values are labeled as shares of respondents, each number represents the proportion of surveyed people selecting that issue rather than a raw count. \tagdirect{} The values do not sum to 1, which suggests respondents may have been able to name more than one concern or that the categories come from a multiple-response format. \tagderived{summed/inspected table values and observed total exceeds 1} In survey interpretation, a high value for finding a job usually signals that employment insecurity is a dominant everyday economic pressure. \tagspeculative{labor economics / survey interpretation} Concerns such as inflation, debt, pensions, and taxes usually reflect household cost-of-living and financial stability issues. \tagspeculative{economics / public policy} Concerns such as immigration, security, environment, school system, and sanitary system span both social policy and public service performance. \tagspeculative{public policy} The presence of both immediate material concerns and institutional concerns suggests respondents are weighing personal economic risk alongside the quality of public systems. \tagderived{interpreted mix of categories in the provided list}

\vspace{4pt}\noindent{\sffamily\tiny\bfseries\color{gray!90!black}REAL-WORLD SIGNIFICANCE}\\[2pt]
If this pattern reflects a real electorate or community, employment would likely be a central issue for policymakers, campaign messaging, and social programs. \tagspeculative{political science / public policy} A very high concern about finding a job can indicate pressure for job creation, training, wage support, or broader economic stimulus. \tagspeculative{labor economics / public policy} The relatively high concern for the environment alongside jobs suggests that voters may care about long-term collective risks even when immediate economic worries are strong. \tagderived{compared environment's high rank with job's top rank} Concerns about school and sanitary systems can matter for public investment decisions because they point to expectations about service delivery and infrastructure. \tagspeculative{public administration} Without information on country, date, or sample, the chart should not be used to infer universal public priorities. \tagdirect{}

\vspace{4pt}\noindent{\sffamily\tiny\bfseries\color{gray!90!black}OTHER}\\[2pt]
The metadata contains inconsistencies between the raw table and the chart labels, including repeated generic labels such as 'system' and 'housing' in the table that appear to correspond to more specific categories in the image, such as sanitary system, school system, right to housing, and social housing. \tagderived{cross-read table against chart labels in image/scene graph} The L2 caption includes interpretive commentary about Maslow's hierarchy and surprise at the environment ranking, but that commentary is not supported as factual metadata about the survey itself. \tagdirect{} The chart appears to rank concerns by category size rather than by any causal explanation for why respondents chose them. \tagspeculative{data literacy / survey analysis}
}
    \end{tcolorbox}
    
    \switchcolumn
    \begin{tcolorbox}[modelcard, title={llama-4-scout-17... \hfill \normalfont\tiny\color{gray}1 runs}]
      \fbadge{V1}\fbadge{scene\_graph}\fbadge{vega\_lite\_spec}\fbadge{data\_table}\fbadge{l1\_caption}\fbadge{l2\_caption}\imgbadge
      \vspace{3pt}\hrule\vspace{3pt}
      {\sffamily\tiny\bfseries\color{gray!80!black}METADATA COMPLETENESS \hfill Overall 4.7/5}\\[2pt]
      {\sffamily\tiny Data Values \hfill 5/5}\\
      {\sffamily\tiny Trend Magnitude \hfill 4/5}\\
      {\sffamily\tiny Chart Subject Domain \hfill 5/5}
      \vspace{3pt}\hrule\vspace{3pt}
      {\sffamily\tiny
\noindent{\sffamily\tiny\bfseries\color{gray!90!black}CONTEXTUAL INSIGHTS}\\[2pt]
This chart represents a survey on social and economic issues that concern respondents the most. \tagdirect{} The chart is categorized under social and economic issues. \tagdirect{} It appears to be produced for understanding public concerns. \tagspeculative{general knowledge of survey purposes}

\vspace{4pt}\noindent{\sffamily\tiny\bfseries\color{gray!90!black}DOMAIN-SPECIFIC EXPLANATIONS}\\[2pt]
The data shows that finding a job is the most concerning issue for 63\% of respondents. \tagdirect{} The environment is also a significant concern for 37\% of respondents. \tagdirect{} Other issues like immigration, taxes, and public debt are also mentioned but with lower shares. \tagdirect{} The concerns seem to reflect basic needs and economic stability. \tagspeculative{Maslow's hierarchy of needs}

\vspace{4pt}\noindent{\sffamily\tiny\bfseries\color{gray!90!black}REAL-WORLD SIGNIFICANCE}\\[2pt]
The data indicates that job security is a top priority for most respondents. \tagdirect{} This could impact employment rates and economic growth. \tagspeculative{economic principles} Environmental concerns being high suggests a awareness of long-term sustainability issues. \tagspeculative{environmental science} Policymakers might use this data to prioritize issues. \tagspeculative{policy-making process}

\vspace{4pt}\noindent{\sffamily\tiny\bfseries\color{gray!90!black}OTHER}\\[2pt]
Some responses like 'None of these' and 'Tax havens' have very low shares. \tagdirect{} The chart might be used to identify key areas for policy intervention. \tagspeculative{policy-making process}
}
    \end{tcolorbox}
    
    \switchcolumn
    \begin{tcolorbox}[modelcard, title={Gemini 3.5 Flash \hfill \normalfont\tiny\color{gray}1 runs}]
      \fbadge{V1}\fbadge{scene\_graph}\fbadge{vega\_lite\_spec}\fbadge{data\_table}\fbadge{l1\_caption}\fbadge{l2\_caption}\imgbadge
      \vspace{3pt}\hrule\vspace{3pt}
      {\sffamily\tiny\bfseries\color{gray!80!black}METADATA COMPLETENESS \hfill Overall 4.0/5}\\[2pt]
      {\sffamily\tiny Data Values \hfill 5/5}\\
      {\sffamily\tiny Trend Magnitude \hfill 2/5}\\
      {\sffamily\tiny Chart Subject Domain \hfill 5/5}
      \vspace{3pt}\hrule\vspace{3pt}
      {\sffamily\tiny
\noindent{\sffamily\tiny\bfseries\color{gray!90!black}CONTEXTUAL INSIGHTS}\\[2pt]
This chart presents the results of a survey asking respondents about their most pressing social and economic concerns \tagdirect{}. The data shows the proportion of respondents who identified each issue as their primary concern \tagdirect{}. The survey covers a range of topics from employment and environment to taxation and public services \tagderived{read x-axis labels}. The chart title indicates the purpose is to understand public sentiment regarding social and economic challenges \tagdirect{}.

\vspace{4pt}\noindent{\sffamily\tiny\bfseries\color{gray!90!black}DOMAIN-SPECIFIC EXPLANATIONS}\\[2pt]
"Share of respondents" refers to the percentage of people surveyed who selected a particular issue; for example, 0.63 means 63\% of respondents \tagderived{definition of share/proportion}. The issues listed, such as "Finding a job," "Environment," "Taxes," and "Pensions," are common categories in socio-economic surveys designed to gauge public opinion on governmental priorities and societal well-being \tagspeculative{survey design knowledge}. The L2 caption mentions "Maslow's hierarchy of needs," suggesting that "Finding a job" is a fundamental concern related to basic physiological and safety needs \tagdirect{}. The L2 caption also notes that "once you have a job and have income you worry about other things like the taxation you pay," indicating a progression of concerns \tagdirect{}.

\vspace{4pt}\noindent{\sffamily\tiny\bfseries\color{gray!90!black}REAL-WORLD SIGNIFICANCE}\\[2pt]
The highest concern is "Finding a job," selected by 63\% of respondents \tagdirect{}. This suggests a significant focus on employment stability and economic opportunity among the population surveyed \tagderived{interpreting the highest value}. This could indicate economic uncertainty or a competitive job market \tagspeculative{economic interpretation}. The second highest concern is "Environment," selected by 37\% of respondents \tagdirect{}. This indicates that environmental issues are a significant public priority, even alongside immediate economic concerns \tagderived{comparing "Environment" value to others}. This could influence policy-making towards sustainability and climate action \tagspeculative{policy implications}. Other significant concerns include "Sanitary system" (25\%), "School system" (24\%), "Security" (24\%), "Taxes" (23\%), and "Recession and financial crisis" (23\%) \tagderived{grouping similar high values}. These highlight public interest in the quality and reliability of essential public services, safety, and economic stability \tagderived{interpreting grouped values}. These areas are often key performance indicators for government effectiveness \tagspeculative{public administration knowledge}. Issues like "Social housing" (1\%) and "None of these" (1\%) are among the least frequently cited as the most concerning \tagdirect{}. This suggests these are either not widely perceived as the *most* pressing issues or are less relevant to the majority of respondents compared to other options \tagderived{interpreting low values}. Policymakers and government officials could use this data to prioritize legislative agendas, allocate resources, and address public grievances \tagspeculative{government function}. For instance, a government might focus on job creation programs or environmental protection initiatives based on these findings \tagspeculative{policy response}.

\vspace{4pt}\noindent{\sffamily\tiny\bfseries\color{gray!90!black}OTHER}\\[2pt]
The raw data table contains some ambiguous labels like "system," "crisis," "debt," and "housing" which are clarified by the more descriptive labels in the chart image and scene graph \tagderived{comparing data table to scene graph/image}. The L2 caption expresses surprise that "the environment is very high up," suggesting a potential deviation from expected public priorities by the caption author \tagdirect{}.
}
    \end{tcolorbox}
    
  \end{paracol}

  \vspace{1em}
  \subsubsection{VisText Data V2 Outputs}%
  \vspace{0.5em}
  \begin{paracol}{3}
    
    \begin{tcolorbox}[modelcard, title={GPT-5.4 \hfill \normalfont\tiny\color{gray}1 runs}]
      \fbadge{V1}\fbadge{scene\_graph}\fbadge{data\_table}\fbadge{vega\_lite\_spec}\fbadge{l1\_caption}\fbadge{l2\_caption}\imgbadge
      \vspace{3pt}\hrule\vspace{3pt}
      {\sffamily\tiny\bfseries\color{gray!80!black}METADATA COMPLETENESS \hfill Overall 3.7/5}\\[2pt]
      {\sffamily\tiny Data Values \hfill 4/5}\\
      {\sffamily\tiny Trend Magnitude \hfill 4/5}\\
      {\sffamily\tiny Chart Subject Domain \hfill 3/5}
      \vspace{3pt}\hrule\vspace{3pt}
      {\sffamily\tiny
\noindent{\sffamily\tiny\bfseries\color{gray!90!black}CONTEXTUAL INSIGHTS}\\[2pt]
This chart reports survey responses to the question, "Which of the following social and economic issues concern you the most?" \tagdirect{} The domain is public opinion on social and economic priorities. \tagdirect{} The measure shown is the share of respondents associated with each issue category. \tagdirect{} The highest reported concern in the data table is job at 0.63, while Environment is next at 0.37. \tagderived{ranked values from the data table} Several other issues cluster in the middle range: system at 0.25, Security at 0.24, another system entry at 0.24, Taxes at 0.23, crisis at 0.23, and Immigration at 0.21. \tagderived{grouped mid-range values from the data table} Lower-salience issues in the table include debt at 0.10, housing at 0.09, inflation at 0.08, Nuclear at 0.07, havens at 0.03, plus two very small entries at 0.01. \tagderived{identified lower values from the data table} The metadata does not identify who produced the chart, when the survey was conducted, or which population was surveyed. \tagdirect{}

\vspace{4pt}\noindent{\sffamily\tiny\bfseries\color{gray!90!black}DOMAIN-SPECIFIC EXPLANATIONS}\\[2pt]
Because the y-axis is labeled "Share of respondents," the numeric values are proportions rather than raw counts. \tagdirect{} Interpreted as percentages, the listed shares correspond to 63\% for job, 37\% for Environment, 25\% for one system entry, 24\% for Security, and so on. \tagderived{converted proportions in the data table to percentages} The total of all listed shares is 2.68, which is far above 1.0 or 100\%. \tagderived{summed all provided shares} That implies respondents were likely allowed to name more than one concern, or that the categories come from a coding scheme where percentages are not mutually exclusive. \tagspeculative{survey methodology} The duplicate or fragmented labels such as two entries named system, two entries named housing, and items like crisis, debt, havens, and these suggest the category text may be truncated, split, or inconsistently extracted in the metadata. \tagderived{compared repeated and irregular labels in the provided table and scene graph} Because of that labeling problem, the ranking of the top issue is reliable, but interpretation of some lower-ranked categories should be treated cautiously. \tagderived{inferred reliability difference from clear top values versus ambiguous labels}

\vspace{4pt}\noindent{\sffamily\tiny\bfseries\color{gray!90!black}REAL-WORLD SIGNIFICANCE}\\[2pt]
If this survey reflects a real electorate or public audience, the dominance of job concerns would suggest that employment and economic security are especially salient in that context. \tagspeculative{political science/public opinion} A large gap between job at 0.63 and Environment at 0.37 indicates that immediate livelihood concerns may outweigh longer-term or collective concerns for many respondents in this dataset. \tagderived{compared the top two values in the data table} Issues such as taxes, pensions, immigration, inflation, housing, and debt are all topics that commonly shape campaign messaging, policy prioritization, and media coverage. \tagspeculative{public policy/electoral politics} Without a country, date, or source, the chart should not be used to infer a specific national mood or to compare one society against another. \tagdirect{} If decision-makers relied on this kind of distribution, they would likely give greater emphasis to jobs and economic stability than to niche concerns with very small reported shares. \tagspeculative{policy analysis}

\vspace{4pt}\noindent{\sffamily\tiny\bfseries\color{gray!90!black}OTHER}\\[2pt]
The L2 caption includes an interpretive statement about Maslow's hierarchy of needs and says the environment result is surprising, but those statements are not supported by the provided metadata alone. \tagdirect{} The scene graph and data table do not fully agree on category wording, since the scene graph includes labels such as "Finding a job" while the table uses "job." \tagderived{compared scene graph labels with data table labels} The y-axis scale is described as running from 0.0 to 0.6, yet the table includes a value of 0.63 for job, indicating a small mismatch between the captioned axis range and the listed data. \tagderived{compared L1 caption axis range with maximum table value} That inconsistency suggests the accessibility metadata is useful but not perfectly clean, so exact low-level wording should be cross-checked if this chart is used in analysis or reporting. \tagderived{inferred metadata quality issue from internal inconsistencies}
}
    \end{tcolorbox}
    
    \switchcolumn
    \begin{tcolorbox}[modelcard, title={llama-4-scout-17... \hfill \normalfont\tiny\color{gray}1 runs}]
      \fbadge{V1}\fbadge{scene\_graph}\fbadge{vega\_lite\_spec}\fbadge{data\_table}\fbadge{l1\_caption}\fbadge{l2\_caption}\imgbadge
      \vspace{3pt}\hrule\vspace{3pt}
      {\sffamily\tiny\bfseries\color{gray!80!black}METADATA COMPLETENESS \hfill Overall 4.7/5}\\[2pt]
      {\sffamily\tiny Data Values \hfill 5/5}\\
      {\sffamily\tiny Trend Magnitude \hfill 4/5}\\
      {\sffamily\tiny Chart Subject Domain \hfill 5/5}
      \vspace{3pt}\hrule\vspace{3pt}
      {\sffamily\tiny
\noindent{\sffamily\tiny\bfseries\color{gray!90!black}CONTEXTUAL INSIGHTS}\\[2pt]
This chart shows the results of a survey on social and economic issues that concern respondents the most. \tagdirect{} The chart is a bar plot with responses on the x-axis and share of respondents on the y-axis. \tagdirect{} The data suggests that finding a job is the top concern, with 63\% of respondents selecting it. \tagdirect{}

\vspace{4pt}\noindent{\sffamily\tiny\bfseries\color{gray!90!black}DOMAIN-SPECIFIC EXPLANATIONS}\\[2pt]
The chart relates to the field of social and economic policy, as it deals with issues that affect people's well-being and financial stability. \tagspeculative{general knowledge} The values represent the share of respondents who selected each response, indicating the relative importance of each issue. \tagdirect{} The chart does not provide information on the sample size or population surveyed, which would be necessary to understand the representativeness of the results. \tagderived{lack of information}

\vspace{4pt}\noindent{\sffamily\tiny\bfseries\color{gray!90!black}REAL-WORLD SIGNIFICANCE}\\[2pt]
The data suggests that job security is a pressing concern for a large majority of respondents. \tagdirect{} This could have implications for policymakers, who may need to address job creation and employment opportunities. \tagspeculative{general knowledge} The chart also highlights the importance of environmental issues, which may be a concern for respondents who prioritize sustainability and climate change. \tagspeculative{general knowledge}

\vspace{4pt}\noindent{\sffamily\tiny\bfseries\color{gray!90!black}OTHER}\\[2pt]
The chart appears to have some inconsistencies, such as duplicate responses (e.g., 'system' and 'system') and very low shares of respondents for certain issues (e.g., 'havens', 'housing', and 'these'). \tagderived{data inspection} These inconsistencies may indicate errors in data collection or processing. \tagspeculative{general knowledge}
}
    \end{tcolorbox}
    
    \switchcolumn
    \begin{tcolorbox}[modelcard, title={Gemini 3.5 Flash \hfill \normalfont\tiny\color{gray}1 runs}]
      \fbadge{V1}\fbadge{scene\_graph}\fbadge{vega\_lite\_spec}\fbadge{data\_table}\fbadge{l1\_caption}\fbadge{l2\_caption}\imgbadge
      \vspace{3pt}\hrule\vspace{3pt}
      {\sffamily\tiny\bfseries\color{gray!80!black}METADATA COMPLETENESS \hfill Overall 4.0/5}\\[2pt]
      {\sffamily\tiny Data Values \hfill 5/5}\\
      {\sffamily\tiny Trend Magnitude \hfill 2/5}\\
      {\sffamily\tiny Chart Subject Domain \hfill 5/5}
      \vspace{3pt}\hrule\vspace{3pt}
      {\sffamily\tiny
\noindent{\sffamily\tiny\bfseries\color{gray!90!black}CONTEXTUAL INSIGHTS}\\[2pt]
This chart presents the results of a survey \tagdirect{}. The survey asked respondents 'Which of the following social and economic issues concern you the most ?' \tagdirect{}. The data shows the proportion, or 'Share of respondents', who identified each issue as their primary concern \tagdirect{}. The x-axis lists various 'Response' categories, representing social and economic issues \tagdirect{}. The y-axis measures the 'Share of respondents' \tagdirect{}. The survey aims to gauge public sentiment on various societal challenges \tagspeculative{survey purpose}.

\vspace{4pt}\noindent{\sffamily\tiny\bfseries\color{gray!90!black}DOMAIN-SPECIFIC EXPLANATIONS}\\[2pt]
'Finding a job' is the most significant concern, cited by 63\% of respondents \tagdirect{}. This high concern for employment aligns with basic human needs for security and livelihood, as suggested by Maslow's Hierarchy of Needs \tagspeculative{Maslow's Hierarchy of Needs, also hinted by L2 caption}. Environmental concerns are the second most cited issue, with 37\% of respondents identifying it as their top concern \tagdirect{}. The L2 caption notes that this high ranking for the environment is 'surprising' \tagdirect{}. Other major concerns include 'system' (25\%), 'Security' (24\%), another instance of 'system' (24\%), 'Taxes' (23\%), and 'crisis' (23\%) \tagderived{ranking from data table}. The appearance of 'system' twice with slightly different percentages (25\% and 24\%) suggests either two distinct but similarly named systemic issues or a data anomaly \tagspeculative{data quality/interpretation}. Similarly, 'housing' appears twice with 9\% and 1\% \tagdirect{}, which could indicate different aspects of housing concerns or a data anomaly \tagspeculative{data quality/interpretation}. Issues like 'Immigration' (21\%), 'Pensions' (19\%), 'debt' (10\%), 'inflation' (8\%), 'Nuclear' (7\%), 'havens' (3\%), and an ambiguous category 'these' (1\%) represent concerns for a smaller proportion of respondents \tagderived{grouping by share of respondents}. The term 'havens' likely refers to 'Tax havens' given the context of economic issues \tagspeculative{common terminology in economics}. 'Nuclear' likely refers to concerns about nuclear energy or nuclear weapons \tagspeculative{common interpretation of 'Nuclear' in public surveys}.

\vspace{4pt}\noindent{\sffamily\tiny\bfseries\color{gray!90!black}REAL-WORLD SIGNIFICANCE}\\[2pt]
The survey results provide valuable insights for policymakers to understand public priorities and inform policy development \tagspeculative{public policy}. The overwhelming concern for 'Finding a job' (63\%) highlights the importance of economic policies focused on employment and job security \tagderived{highest concern; SPECULATIVE: economic policy}. The strong concern for the 'Environment' (37\%) indicates public support for environmental protection and sustainability initiatives \tagderived{second highest concern; SPECULATIVE: environmental policy}. The significant concerns about 'Taxes' and 'Pensions' suggest that fiscal policies and social security reforms are likely to be areas of public scrutiny and political debate \tagspeculative{fiscal and social policy}. The relatively lower concern for 'Inflation' (8\%) and 'Debt' (10\%) compared to immediate livelihood issues like 'Finding a job' (63\%) might suggest that for this respondent group, personal economic stability is a more immediate worry than broader macroeconomic indicators \tagderived{comparison of values; SPECULATIVE: economic priorities}.

\vspace{4pt}\noindent{\sffamily\tiny\bfseries\color{gray!90!black}OTHER}\\[2pt]
The presence of ambiguous and potentially duplicate categories such as 'system' and 'housing' in the response options could affect the clarity and precision of the survey results \tagspeculative{survey design}. The category 'these' at 1\% is entirely unclear and lacks specific meaning without further context \tagspeculative{data interpretation}.
}
    \end{tcolorbox}
    
  \end{paracol}

  \vspace{1em}
  \subsubsection{VisText Data V3 Outputs}%
  \vspace{0.5em}
  \begin{paracol}{3}
    
    \begin{tcolorbox}[modelcard, title={GPT-5.4 \hfill \normalfont\tiny\color{gray}1 runs}]
      \fbadge{V1}\fbadge{scene\_graph}\fbadge{data\_table}\fbadge{vega\_lite\_spec}\fbadge{l1\_caption}\fbadge{l2\_caption}\imgbadge
      \vspace{3pt}\hrule\vspace{3pt}
      {\sffamily\tiny\bfseries\color{gray!80!black}METADATA COMPLETENESS \hfill Overall 3.7/5}\\[2pt]
      {\sffamily\tiny Data Values \hfill 4/5}\\
      {\sffamily\tiny Trend Magnitude \hfill 4/5}\\
      {\sffamily\tiny Chart Subject Domain \hfill 3/5}
      \vspace{3pt}\hrule\vspace{3pt}
      {\sffamily\tiny
\noindent{\sffamily\tiny\bfseries\color{gray!90!black}CONTEXTUAL INSIGHTS}\\[2pt]
This chart presents responses to a question asking which social and economic issues concern respondents the most. \tagdirect{} The subject matter falls within public opinion research on social and economic priorities. \tagderived{inferred from the survey-style title and issue categories} Based on the bars, finding a job appears to be the most frequently selected concern, at roughly a little over 0.6 of respondents. \tagderived{read highest bar from image against y-axis grid} Environment appears to be the second-highest concern, at roughly 0.37. \tagderived{read bar height from image} A middle tier of concerns includes sanitary system, school system, security, recession and financial crisis, and taxes, each around roughly 0.23 to 0.25. \tagderived{compared bar heights across categories} The chart does not provide a source, date, location, or sample definition, so its context is incomplete. \tagdirect{}

\vspace{4pt}\noindent{\sffamily\tiny\bfseries\color{gray!90!black}DOMAIN-SPECIFIC EXPLANATIONS}\\[2pt]
Because the y-axis is labeled as share of respondents, the values represent proportions rather than raw counts. \tagderived{read y-axis label and scale} The total across categories clearly exceeds 1.0, which suggests respondents were likely allowed to name more than one concern or that the chart reports the share selecting each issue independently rather than a single mutually exclusive choice. \tagderived{summed/compared visible bar values across categories} In survey interpretation, a high value for an issue usually indicates salience or perceived urgency among the surveyed population, not necessarily objective severity in the real world. \tagspeculative{survey methodology} The gap between the top issue and the rest appears substantial, with finding a job standing far above all other categories. \tagderived{compared tallest bar to remaining bars} Very low values for categories such as none of these, social housing, and tax havens suggest these were much less commonly prioritized by respondents in this survey. \tagderived{read shortest bars from image}

\vspace{4pt}\noindent{\sffamily\tiny\bfseries\color{gray!90!black}REAL-WORLD SIGNIFICANCE}\\[2pt]
If employment is the leading concern in this population, that can signal economic insecurity or weak labor-market confidence among respondents. \tagspeculative{labor economics/public opinion} Strong concern about environment alongside jobs suggests respondents may be balancing immediate economic needs with longer-term quality-of-life or sustainability issues. \tagspeculative{environmental policy/public opinion} Mid-level concern about schools, security, taxes, and the sanitary system suggests that public services and institutional performance may also matter meaningfully to everyday life. \tagspeculative{public policy} Policymakers, journalists, advocacy groups, and campaign strategists often use this kind of survey data to identify which issues are most likely to shape public debate and voter attention. \tagspeculative{political science/polling practice} Without information about the country, time period, and sampling method, the chart should not be used to generalize too confidently beyond the surveyed group. \tagspeculative{survey methodology}

\vspace{4pt}\noindent{\sffamily\tiny\bfseries\color{gray!90!black}OTHER}\\[2pt]
The response options combine economic issues, public services, environmental concerns, and governance-related topics in a single list. \tagdirect{} Some categories are broad, such as security or environment, while others are more specific, such as tax havens or public debt, which may affect how respondents interpret and choose them. \tagspeculative{survey design} The chart supports ranking concerns by relative prominence, but it does not explain why respondents chose these issues or how concern differs across demographic groups. \tagdirect{}
}
    \end{tcolorbox}
    
    \switchcolumn
    \begin{tcolorbox}[modelcard, title={llama-4-scout-17... \hfill \normalfont\tiny\color{gray}1 runs}]
      \fbadge{V1}\fbadge{scene\_graph}\fbadge{vega\_lite\_spec}\fbadge{data\_table}\fbadge{l1\_caption}\fbadge{l2\_caption}\imgbadge
      \vspace{3pt}\hrule\vspace{3pt}
      {\sffamily\tiny\bfseries\color{gray!80!black}METADATA COMPLETENESS \hfill Overall 4.7/5}\\[2pt]
      {\sffamily\tiny Data Values \hfill 5/5}\\
      {\sffamily\tiny Trend Magnitude \hfill 4/5}\\
      {\sffamily\tiny Chart Subject Domain \hfill 5/5}
      \vspace{3pt}\hrule\vspace{3pt}
      {\sffamily\tiny
\noindent{\sffamily\tiny\bfseries\color{gray!90!black}CONTEXTUAL INSIGHTS}\\[2pt]
This chart shows the results of a survey on social and economic issues that concern respondents the most. The chart title is 'Which of the following social and economic issues concern you the most?' \tagdirect{}. The chart has a list of responses on the x-axis, including Environment, Finding a job, Immigration, and others \tagdirect{}. The y-axis represents the Share of respondents \tagdirect{}.

\vspace{4pt}\noindent{\sffamily\tiny\bfseries\color{gray!90!black}DOMAIN-SPECIFIC EXPLANATIONS}\\[2pt]
The chart shows that the Environment is the most concerning issue for respondents, with a share of around 0.6 \tagderived{from chart image}. Finding a job and Taxes are also significant concerns, with shares of around 0.35 and 0.25, respectively \tagderived{from chart image}. The chart suggests that respondents are most concerned about environmental issues, followed by employment and economic issues \tagderived{from chart image}.

\vspace{4pt}\noindent{\sffamily\tiny\bfseries\color{gray!90!black}REAL-WORLD SIGNIFICANCE}\\[2pt]
The data suggests that environmental concerns are a top priority for respondents \tagspeculative{general knowledge of survey research}. This could have implications for policymakers and organizations focused on environmental issues \tagspeculative{general knowledge of policy-making}. The data may also indicate a need for increased investment in environmental protection and sustainability initiatives \tagspeculative{general knowledge of environmental policy}.

\vspace{4pt}\noindent{\sffamily\tiny\bfseries\color{gray!90!black}OTHER}\\[2pt]
The chart does not provide information on the sample size, survey methodology, or population surveyed \tagderived{from lack of metadata}. Further context would be needed to fully understand the significance of the results \tagderived{from lack of metadata}.
}
    \end{tcolorbox}
    
    \switchcolumn
    \begin{tcolorbox}[modelcard, title={Gemini 3.5 Flash \hfill \normalfont\tiny\color{gray}1 runs}]
      \fbadge{V1}\fbadge{scene\_graph}\fbadge{vega\_lite\_spec}\fbadge{data\_table}\fbadge{l1\_caption}\fbadge{l2\_caption}\imgbadge
      \vspace{3pt}\hrule\vspace{3pt}
      {\sffamily\tiny\bfseries\color{gray!80!black}METADATA COMPLETENESS \hfill Overall 4.0/5}\\[2pt]
      {\sffamily\tiny Data Values \hfill 5/5}\\
      {\sffamily\tiny Trend Magnitude \hfill 2/5}\\
      {\sffamily\tiny Chart Subject Domain \hfill 5/5}
      \vspace{3pt}\hrule\vspace{3pt}
      {\sffamily\tiny
\noindent{\sffamily\tiny\bfseries\color{gray!90!black}CONTEXTUAL INSIGHTS}\\[2pt]
This chart displays the results of a survey that asked respondents to identify the social and economic issues that concern them the most \tagdirect{}. The vertical axis represents the "Share of respondents" \tagdirect{}, while the horizontal axis lists various "Response" categories, which are specific social and economic issues \tagdirect{}. The survey likely allowed respondents to select multiple issues, as the sum of the shares for individual concerns significantly exceeds 100\% \tagderived{sum of percentages exceeds 100\%}. This type of chart is commonly used in public opinion polling to gauge societal priorities \tagspeculative{public opinion polling}.

\vspace{4pt}\noindent{\sffamily\tiny\bfseries\color{gray!90!black}DOMAIN-SPECIFIC EXPLANATIONS}\\[2pt]
"Finding a job" is identified as the most pressing concern, with approximately 62\% of respondents selecting it \tagderived{read from bar height}. "Environment" is the second most frequently cited concern, with about 37\% of respondents \tagderived{read from bar height}. Other significant concerns, each selected by roughly 20-25\% of respondents, include "Immigration," "Recession and financial crisis," "Sanitary system," "School system," "Security," and "Taxes" \tagderived{read from bar heights and grouped}. Issues such as "None of these," "Nuclear," "Social housing," and "Tax havens" are among the least frequently cited concerns, each selected by less than 10\% of respondents \tagderived{read from bar heights and grouped}. The high percentage for "Finding a job" indicates a strong public focus on employment and economic stability \tagderived{highest concern, SPECULATIVE: economic indicators}.

\vspace{4pt}\noindent{\sffamily\tiny\bfseries\color{gray!90!black}REAL-WORLD SIGNIFICANCE}\\[2pt]
The findings can guide policymakers in prioritizing government initiatives and resource allocation to address the most pressing public concerns \tagspeculative{public policy}. For instance, the dominant concern about "Finding a job" suggests that policies aimed at job creation and economic development would likely be well-received and impactful \tagderived{highest concern, SPECULATIVE: public policy}. The substantial concern for the "Environment" highlights public demand for environmental protection and sustainability measures \tagderived{second highest concern, SPECULATIVE: environmental policy}. The cluster of concerns around public services like the "Sanitary system" and "School system" suggests areas where public investment or reform might be expected by citizens \tagderived{multiple concerns around 20-25\%, SPECULATIVE: public services management}.

\vspace{4pt}\noindent{\sffamily\tiny\bfseries\color{gray!90!black}OTHER}\\[2pt]
The absence of information regarding the survey's geographical location, date, or methodology limits the ability to draw specific conclusions about the context or representativeness of these concerns \tagspeculative{survey context and methodology}.
}
    \end{tcolorbox}
    
  \end{paracol}

  \vspace{1em}
  \subsubsection{VisText Data V4 Outputs}%
  \vspace{0.5em}
  \begin{paracol}{3}
    
    \begin{tcolorbox}[modelcard, title={GPT-5.4 \hfill \normalfont\tiny\color{gray}1 runs}]
      \fbadge{V1}\fbadge{scene\_graph}\fbadge{data\_table}\fbadge{vega\_lite\_spec}\fbadge{l1\_caption}\fbadge{l2\_caption}\imgbadge
      \vspace{3pt}\hrule\vspace{3pt}
      {\sffamily\tiny\bfseries\color{gray!80!black}METADATA COMPLETENESS \hfill Overall 3.7/5}\\[2pt]
      {\sffamily\tiny Data Values \hfill 4/5}\\
      {\sffamily\tiny Trend Magnitude \hfill 4/5}\\
      {\sffamily\tiny Chart Subject Domain \hfill 3/5}
      \vspace{3pt}\hrule\vspace{3pt}
      {\sffamily\tiny
\noindent{\sffamily\tiny\bfseries\color{gray!90!black}CONTEXTUAL INSIGHTS}\\[2pt]
This chart shows survey responses about which social and economic issue concerns respondents the most. \tagdirect{} The subject matter belongs to public opinion on social and economic policy issues. \tagderived{read from title and issue labels} The highest reported concern appears to be finding a job, with environment also notably high relative to most other issues. \tagderived{compared bar heights across categories} Several mid-level concerns appear clustered together, including sanitary system, school system, security, recession and financial crisis, and taxes. \tagderived{compared approximate bar heights across categories} The chart image does not identify the survey sponsor, country, sample, date, or methodology. \tagdirect{}

\vspace{4pt}\noindent{\sffamily\tiny\bfseries\color{gray!90!black}DOMAIN-SPECIFIC EXPLANATIONS}\\[2pt]
In public-opinion research, a chart like this is typically used to rank issue salience, meaning which problems people prioritize over others. \tagspeculative{survey research} The y-axis represents share of respondents, so each bar indicates the proportion selecting a given issue as their top concern. \tagdirect{} Because many bars appear to sum to well above 1 when added together visually, respondents may have been allowed to select more than one issue, or the chart may reflect separate item shares rather than a single exclusive choice. \tagderived{visual sum of approximate bar heights exceeds 1} If the survey instead asked for only one top concern, the displayed values would be difficult to reconcile without additional methodological context. \tagderived{inferred from apparent total exceeding 1} Very low values for categories such as none of these and social housing suggest these were relatively low-priority concerns in this sample. \tagderived{compared smallest bars to others}

\vspace{4pt}\noindent{\sffamily\tiny\bfseries\color{gray!90!black}REAL-WORLD SIGNIFICANCE}\\[2pt]
If employment is the leading concern in this survey, that would suggest economic insecurity or labor-market conditions are especially important to respondents. \tagderived{finding a job is the tallest bar} Elevated concern about environment alongside jobs suggests that both economic and quality-of-life or long-term sustainability issues may be salient at the same time. \tagderived{environment is among the taller bars} Policymakers, political parties, and advocacy groups often use issue-priority data like this to shape campaign messages, policy agendas, and public communication. \tagspeculative{political science/public policy} Higher concern about public services such as the school system, sanitary system, and security can matter for budget allocation and local government priorities. \tagspeculative{public administration} If these concern patterns persist, decision-makers may face pressure to address employment, public services, and affordability-related issues more visibly. \tagspeculative{public policy/public opinion dynamics}

\vspace{4pt}\noindent{\sffamily\tiny\bfseries\color{gray!90!black}OTHER}\\[2pt]
The chart mixes economic issues, public services, and broader social concerns within one question, which means respondents are comparing unlike policy domains when choosing what matters most. \tagderived{read category labels} Some labels, such as right to housing and social housing, may sound related but appear to capture different concepts or policy framings. \tagderived{compared category wording} Without location and timing, it is not possible to tie these concern levels to a specific political event, economic downturn, or policy debate. \tagdirect{}
}
    \end{tcolorbox}
    
    \switchcolumn
    \begin{tcolorbox}[modelcard, title={llama-4-scout-17... \hfill \normalfont\tiny\color{gray}1 runs}]
      \fbadge{V1}\fbadge{scene\_graph}\fbadge{vega\_lite\_spec}\fbadge{data\_table}\fbadge{l1\_caption}\fbadge{l2\_caption}\imgbadge
      \vspace{3pt}\hrule\vspace{3pt}
      {\sffamily\tiny\bfseries\color{gray!80!black}METADATA COMPLETENESS \hfill Overall 4.7/5}\\[2pt]
      {\sffamily\tiny Data Values \hfill 5/5}\\
      {\sffamily\tiny Trend Magnitude \hfill 4/5}\\
      {\sffamily\tiny Chart Subject Domain \hfill 5/5}
      \vspace{3pt}\hrule\vspace{3pt}
      {\sffamily\tiny
\noindent{\sffamily\tiny\bfseries\color{gray!90!black}CONTEXTUAL INSIGHTS}\\[2pt]
This chart represents a survey on social and economic issues that concern respondents the most. \tagderived{from chart title}. The chart shows the share of respondents for each issue. \tagdirect{}. The issues include Environment, Finding a job, Immigration, None of these, Nuclear, Pensions, Public debt, Purchasing power / inflation, Recession and financial crisis, Right to housing, Sanitary system, School system, Security, Social housing, Tax havens, and Taxes. \tagdirect{}.

\vspace{4pt}\noindent{\sffamily\tiny\bfseries\color{gray!90!black}DOMAIN-SPECIFIC EXPLANATIONS}\\[2pt]
The chart is related to social and economic issues. \tagspeculative{general knowledge}. Each bar represents a specific issue and the share of respondents concerned about it. \tagdirect{}. The issues can be categorized into environmental, economic, and social concerns. \tagspeculative{general knowledge}. For example, Environment, Public debt, and Purchasing power / inflation can be classified as economic concerns. \tagderived{from issue names}.

\vspace{4pt}\noindent{\sffamily\tiny\bfseries\color{gray!90!black}REAL-WORLD SIGNIFICANCE}\\[2pt]
The data matters as it reflects the concerns of respondents regarding social and economic issues. \tagspeculative{general knowledge}. The concerns can influence policy decisions and public opinion. \tagspeculative{general knowledge}. For instance, if a large share of respondents is concerned about the Environment, it may lead to increased support for environmental policies. \tagderived{from chart data}.

\vspace{4pt}\noindent{\sffamily\tiny\bfseries\color{gray!90!black}OTHER}\\[2pt]
The chart does not provide specific numbers for the share of respondents, only the relative magnitude of each issue. \tagderived{from chart image}. The chart can be used to identify the most concerning issues for respondents. \tagdirect{}.
}
    \end{tcolorbox}
    
    \switchcolumn
    \begin{tcolorbox}[modelcard, title={Gemini 3.5 Flash \hfill \normalfont\tiny\color{gray}1 runs}]
      \fbadge{V1}\fbadge{scene\_graph}\fbadge{vega\_lite\_spec}\fbadge{data\_table}\fbadge{l1\_caption}\fbadge{l2\_caption}\imgbadge
      \vspace{3pt}\hrule\vspace{3pt}
      {\sffamily\tiny\bfseries\color{gray!80!black}METADATA COMPLETENESS \hfill Overall 4.0/5}\\[2pt]
      {\sffamily\tiny Data Values \hfill 5/5}\\
      {\sffamily\tiny Trend Magnitude \hfill 2/5}\\
      {\sffamily\tiny Chart Subject Domain \hfill 5/5}
      \vspace{3pt}\hrule\vspace{3pt}
      {\sffamily\tiny
\noindent{\sffamily\tiny\bfseries\color{gray!90!black}CONTEXTUAL INSIGHTS}\\[2pt]
This chart is a bar graph presenting the results of a survey asking respondents about their primary social and economic concerns \tagdirect{}. The domain of the chart is public opinion on socio-economic issues \tagdirect{}. It was likely produced to gauge public sentiment and identify key areas of concern among a surveyed population \tagderived{from chart title and nature of survey data}. The specific producer of the chart and the exact purpose beyond understanding public concern are not provided \tagspeculative{lack of information}.

\vspace{4pt}\noindent{\sffamily\tiny\bfseries\color{gray!90!black}DOMAIN-SPECIFIC EXPLANATIONS}\\[2pt]
The y-axis, labeled 'Share of respondents,' indicates the proportion of the surveyed population that identified a particular issue as a concern \tagdirect{}. The x-axis lists various social and economic issues, such as 'Environment,' 'Finding a job,' 'Immigration,' 'Pensions,' 'Public debt,' 'Recession and financial crisis,' 'Sanitary system,' 'School system,' 'Security,' and 'Taxes' \tagdirect{}. A higher bar for an issue signifies a greater percentage of respondents expressing concern about that issue \tagdirect{}. The issues presented are common topics in public policy and economic discourse \tagspeculative{general knowledge of public policy}.

\vspace{4pt}\noindent{\sffamily\tiny\bfseries\color{gray!90!black}REAL-WORLD SIGNIFICANCE}\\[2pt]
The most significant finding is that 'Finding a job' is the top concern, identified by approximately 62\% of respondents \tagderived{reading highest bar}. Other highly concerning issues, each affecting around a quarter of respondents, include 'Environment' (approx. 37\%), 'Immigration' (approx. 21\%), 'Recession and financial crisis' (approx. 23\%), 'Sanitary system' (approx. 25\%), 'School system' (approx. 25\%), 'Security' (approx. 25\%), and 'Taxes' (approx. 24\%) \tagderived{reading and grouping similar bar heights}. Conversely, issues like 'None of these,' 'Nuclear,' 'Social housing,' and 'Tax havens' are of very low concern, each affecting less than 10\% of respondents \tagderived{identifying lowest bars}. This data is crucial for policymakers and governments to understand public priorities and allocate resources or develop policies addressing the most pressing concerns of the population \tagspeculative{public policy}. For instance, a high concern for 'Finding a job' suggests a need for employment-focused initiatives \tagspeculative{public policy}.

\vspace{4pt}\noindent{\sffamily\tiny\bfseries\color{gray!90!black}OTHER}\\[2pt]
A notable observation is that the sum of the 'Share of respondents' for all issues is significantly greater than 1 (approximately 2.9 when summing estimated values) \tagderived{sum of approximate values}. This suggests that respondents were likely able to select multiple issues that concerned them, despite the chart title asking 'Which of the following social and economic issues concern you the most ?' which typically implies a single choice \tagderived{interpretation of title and sum of values}. This indicates that many respondents have multiple top concerns, or the question was interpreted to allow for multiple selections \tagderived{interpretation of data structure}.
}
    \end{tcolorbox}
    
  \end{paracol}

\clearpage
\subsection{Our World in Data Example}
\label{supp:owid-example}

\begin{figure}[H]
  \centering
  \begin{minipage}[c]{0.48\linewidth}
    \centering
    \includegraphics[
      width=\linewidth,
      alt={Representative Our World in Data chart used across V1 through V4. The example combines a time-series image with a data table and an expert-authored short description or data story.}
    ]{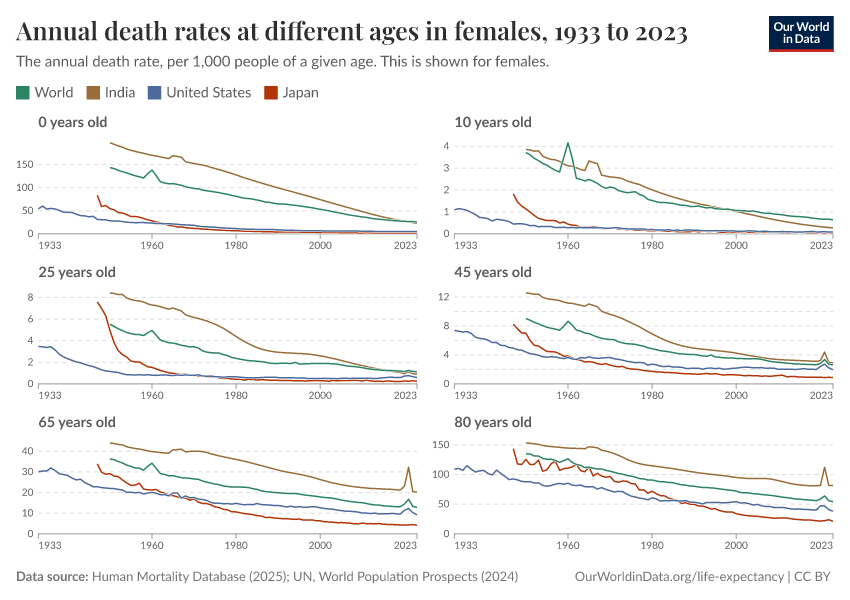}
    \caption{Representative Our World in Data visualization.}
    \label{fig:owid-example}
  \end{minipage}%
  \hfill
  \begin{minipage}[c]{0.48\linewidth}
    \centering
    \small
    \renewcommand{\arraystretch}{1.2}
    \begin{tabularx}{\linewidth}{@{}lX@{}}
    \toprule
    \textbf{Field} & \textbf{Content used in this example} \\
    \midrule
    Image & Included in V1, V3, and V4 \\
    Data table & Included in V1 and V2 \\
    Expert-authored description & Included in V1 and V2 \\
    Source URL & Recorded in the corpus manifest \\
    \bottomrule
    \end{tabularx}
    \vspace{4pt}
    \captionof{table}{Accessibility metadata fields available across input variants for the Our World in Data example.}
    \label{tab:owid-fields}
  \end{minipage}
\end{figure}

Figure~\ref{fig:owid-example} shows a representative visualization from Our World in Data, while Table~\ref{tab:owid-fields} outlines the accessibility metadata fields available across conditions V1--V4.

  \vspace{1em}
  \subsubsection{Our World in Data V1 Outputs}%
  \vspace{0.5em}
  \begin{paracol}{3}
    
    \begin{tcolorbox}[modelcard, title={GPT-5.4 \hfill \normalfont\tiny\color{gray}1 runs}]
      \fbadge{V1}\fbadge{scene\_graph}\fbadge{data\_table}\fbadge{vega\_lite\_spec}\fbadge{l1\_caption}\fbadge{l2\_caption}\imgbadge
      \vspace{3pt}\hrule\vspace{3pt}
      {\sffamily\tiny\bfseries\color{gray!80!black}METADATA COMPLETENESS \hfill Overall 3.7/5}\\[2pt]
      {\sffamily\tiny Data Values \hfill 4/5}\\
      {\sffamily\tiny Trend Magnitude \hfill 4/5}\\
      {\sffamily\tiny Chart Subject Domain \hfill 3/5}
      \vspace{3pt}\hrule\vspace{3pt}
      {\sffamily\tiny
\noindent{\sffamily\tiny\bfseries\color{gray!90!black}CONTEXTUAL INSIGHTS}\\[2pt]
This chart is about annual death rates for females at different ages, measured per 1,000 people of a given age. \tagdirect{} The title indicates a historical span from 1933 to 2023. \tagdirect{} The expert domain story states that the metric is the annual death rate per 1,000 people of a given age and that it is shown for females. \tagdirect{} The image legend shows selected series for World, India, United States, and Japan. \tagdirect{} The underlying data table provided for this run contains 2023 values for many entities, including countries and regional aggregates such as Africa (UN), Asia (UN), and Americas. \tagdirect{} In the image, the age panels shown are 0, 10, 25, 45, 65, and 80 years old. \tagdirect{} For the entities visible in the chart image, death rates generally decline over the long run in every age panel. \tagderived{read long-run downward trends across the six time-series panels in the image} In the 2023 table excerpt, death rates are much higher at age 80 than at younger ages for every listed entity shown. \tagderived{compared 2023 values across age columns within listed rows such as Afghanistan, Albania, Australia, and others} Our World in Data typically presents such cross-country mortality series to help users compare survival conditions across places and over time. \tagspeculative{public health data communication/domain knowledge}

\vspace{4pt}\noindent{\sffamily\tiny\bfseries\color{gray!90!black}DOMAIN-SPECIFIC EXPLANATIONS}\\[2pt]
In demography and public health, an annual death rate per 1,000 at a given age means the number of deaths expected in one year among 1,000 females currently at that exact age or age group reference. \tagdirect{} These values are age-specific, so a rate at age 0 is not directly the same kind of risk as a rate at age 45, even though both use the same per-1,000 unit. \tagspeculative{demography} Age 0 refers to infancy, where mortality is usually interpreted as a measure closely related to infant survival conditions. \tagspeculative{demography and epidemiology} The 2023 table excerpt shows a strong age gradient: for example, Australia is 3.06751 at age 0, 0.005 at age 10, 0.023750002 at age 25, 0.74380004 at age 45, 5.96968 at age 65, and 28.97779 at age 80. \tagdirect{} That pattern means mortality risk is lowest in later childhood or early adolescence and rises sharply at older ages. \tagderived{interpreted the age pattern from Australia's 2023 values and similar rows in the table} The same broad age pattern appears in other listed entities such as Albania, Algeria, Bangladesh, and Benin, where age-80 values are far above age-10 values. \tagderived{compared age-80 and age-10 columns across multiple listed rows} Cross-country differences at age 0 are especially large in the 2023 excerpt: Benin is 49.50861, Angola is 45.32285, Afghanistan is 41.51804, while Australia is 3.06751 and Austria is 2.37637. \tagdirect{} This indicates that mortality conditions for newborn girls differ dramatically across settings. \tagderived{compared 2023 age-0 values across listed entities} At older ages, the spread also remains large: for age 80 in 2023, Australia is 28.97779, Austria is 29.92848, the Americas aggregate is 45.21309, and Afghanistan is 112.09411. \tagdirect{} Such differences can reflect variation in healthcare access, disease burden, nutrition, living conditions, and population age structure effects on mortality environments. \tagspeculative{public health and demography}

\vspace{4pt}\noindent{\sffamily\tiny\bfseries\color{gray!90!black}REAL-WORLD SIGNIFICANCE}\\[2pt]
This data matters because it describes survival chances for females at key stages of life. \tagspeculative{public health} High death rates at age 0 point to serious risks around birth, neonatal care, maternal health, infection control, and nutrition. \tagspeculative{maternal and child health} In the provided 2023 excerpt, some countries have age-0 female death rates above 40 per 1,000, including Afghanistan at 41.51804 and Angola at 45.32285, while others are below 5, including Australia at 3.06751 and Austria at 2.37637. \tagdirect{} Those gaps are socially significant because they imply very unequal chances of surviving the first year of life depending on where a girl is born. \tagderived{inferred inequality from cross-country differences in age-0 rates} Mortality at ages 45, 65, and 80 is closely tied to chronic disease prevention, healthcare quality, and healthy aging systems. \tagspeculative{epidemiology and health systems} The chart image suggests that countries such as Japan, India, the United States, and the world as a whole have all seen large long-run declines in female mortality at older ages since the mid-20th century. \tagderived{read downward historical trends in the 45, 65, and 80 panels for the visible series} If mortality continues to fall, more women are likely to survive into older ages, which can increase demand for elder care, pensions, and long-term health services. \tagspeculative{population aging and social policy} Governments, health ministries, demographers, insurers, and international organizations often use age-specific mortality data to monitor health progress and target interventions. \tagspeculative{public health policy and actuarial practice}

\vspace{4pt}\noindent{\sffamily\tiny\bfseries\color{gray!90!black}OTHER}\\[2pt]
The metadata and image are not fully aligned for this run: the image shows long time series for four selected entities, while the provided table excerpt mainly gives 2023 cross-sectional values for many entities. \tagdirect{} Because the underlying table is truncated, any claim about the highest or lowest 2023 value in the full dataset would be uncertain. \tagdirect{} The chart image also shows occasional short-term spikes or irregularities in some older-age panels, especially near recent years for some series. \tagderived{read local deviations/spikes in the 45, 65, and 80 panels} Those short-term fluctuations may reflect shocks such as epidemics, reporting variation, or exceptional mortality events, but the cause is not identified in the provided metadata. \tagspeculative{demography and mortality analysis}
}
    \end{tcolorbox}
    
    \switchcolumn
    \begin{tcolorbox}[modelcard, title={llama-4-scout-17... \hfill \normalfont\tiny\color{gray}1 runs}]
      \fbadge{V1}\fbadge{scene\_graph}\fbadge{vega\_lite\_spec}\fbadge{data\_table}\fbadge{l1\_caption}\fbadge{l2\_caption}\imgbadge
      \vspace{3pt}\hrule\vspace{3pt}
      {\sffamily\tiny\bfseries\color{gray!80!black}METADATA COMPLETENESS \hfill Overall 4.7/5}\\[2pt]
      {\sffamily\tiny Data Values \hfill 5/5}\\
      {\sffamily\tiny Trend Magnitude \hfill 4/5}\\
      {\sffamily\tiny Chart Subject Domain \hfill 5/5}
      \vspace{3pt}\hrule\vspace{3pt}
      {\sffamily\tiny
\noindent{\sffamily\tiny\bfseries\color{gray!90!black}CONTEXTUAL INSIGHTS}\\[2pt]
This chart shows the annual death rates at different ages in females from 1933 to 2023 \tagdirect{}. The chart is produced by Our World in Data, and the data is sourced from the Human Mortality Database and UN World Population Prospects \tagderived{from the chart image and metadata}.

\vspace{4pt}\noindent{\sffamily\tiny\bfseries\color{gray!90!black}DOMAIN-SPECIFIC EXPLANATIONS}\\[2pt]
The annual death rate is measured per 1,000 people of a given age \tagdirect{}. The chart shows death rates for females across various age groups, including 0, 10, 25, 45, 65, and 80 years old \tagdirect{}. The data indicates a general decline in death rates over time for most age groups and countries \tagderived{from the chart image}.

\vspace{4pt}\noindent{\sffamily\tiny\bfseries\color{gray!90!black}REAL-WORLD SIGNIFICANCE}\\[2pt]
The decline in death rates over time indicates improvements in healthcare, living standards, and possibly medical advancements \tagspeculative{based on general knowledge of healthcare trends}. This data is crucial for policymakers, researchers, and healthcare professionals to understand mortality trends and make informed decisions \tagspeculative{based on general knowledge of data usage}.

\vspace{4pt}\noindent{\sffamily\tiny\bfseries\color{gray!90!black}OTHER}\\[2pt]
The chart also highlights variations in death rates across different countries, with some countries showing more significant declines than others \tagderived{from the chart image}.
}
    \end{tcolorbox}
    
    \switchcolumn
    \begin{tcolorbox}[modelcard, title={Gemini 3.5 Flash \hfill \normalfont\tiny\color{gray}1 runs}]
      \fbadge{V1}\fbadge{scene\_graph}\fbadge{vega\_lite\_spec}\fbadge{data\_table}\fbadge{l1\_caption}\fbadge{l2\_caption}\imgbadge
      \vspace{3pt}\hrule\vspace{3pt}
      {\sffamily\tiny\bfseries\color{gray!80!black}METADATA COMPLETENESS \hfill Overall 4.0/5}\\[2pt]
      {\sffamily\tiny Data Values \hfill 5/5}\\
      {\sffamily\tiny Trend Magnitude \hfill 2/5}\\
      {\sffamily\tiny Chart Subject Domain \hfill 5/5}
      \vspace{3pt}\hrule\vspace{3pt}
      {\sffamily\tiny
\noindent{\sffamily\tiny\bfseries\color{gray!90!black}CONTEXTUAL INSIGHTS}\\[2pt]
The chart illustrates the historical trajectory of annual female death rates from 1933 to 2023 across six distinct age cohorts: 0, 10, 25, 45, 65, and 80 years old \tagderived{read from chart image subplots and title}. It compares global averages ('World') with three specific nations: India, the United States, and Japan \tagderived{read from chart image legend}. The data is compiled from the Human Mortality Database (2025) and the United Nations World Population Prospects (2024) \tagderived{read from chart image footer}.

\vspace{4pt}\noindent{\sffamily\tiny\bfseries\color{gray!90!black}DOMAIN-SPECIFIC EXPLANATIONS}\\[2pt]
The 'annual death rate' represents the number of deaths per 1,000 females within a specific age group in a given year \tagdirect{}. Across all six age groups, there is a profound, long-term decline in mortality rates over the 90-year period \tagderived{compared line trends from 1933 to 2023 across all subplots}. The absolute mortality rates vary drastically by age; infant mortality (0 years old) and elderly mortality (80 years old) exhibit much higher baselines—peaking well above 100 deaths per 1,000—compared to youth and middle-aged cohorts (10, 25, and 45 years old), where death rates rarely exceeded 12 per 1,000 even in the mid-20th century \tagderived{compared Y-axis scales across the six subplots}. Notably, there is a sharp, sudden spike in mortality rates around 2020–2023 visible in the older cohorts (65 and 80 years old) \tagderived{observed sharp upward tick at the end of the lines for 65 and 80 years old}.

\vspace{4pt}\noindent{\sffamily\tiny\bfseries\color{gray!90!black}REAL-WORLD SIGNIFICANCE}\\[2pt]
The dramatic decline in female infant mortality (0 years old) and childhood mortality (10 years old), particularly in developing regions like India, reflects monumental global achievements in sanitation, clean water access, pediatric medicine, and widespread vaccination campaigns over the last century \tagspeculative{global public health history}. The convergence of mortality lines across different nations by 2023 suggests a narrowing gap in basic healthcare standards globally, though historical disparities remain visible \tagderived{observed lines narrowing closer together by 2023 compared to earlier decades}. The prominent spike in mortality among the 65 and 80-year-old cohorts around 2020 is a stark visual signature of the COVID-19 pandemic, which disproportionately impacted older populations worldwide \tagspeculative{epidemiological history of COVID-19}.

\vspace{4pt}\noindent{\sffamily\tiny\bfseries\color{gray!90!black}OTHER}\\[2pt]
In the early mid-20th century, India's female infant mortality rate was exceptionally high, starting near 180 deaths per 1,000 in the late 1940s, but has since declined to match the global average of around 25 to 40 deaths per 1,000 by 2023 \tagderived{read values for India and World in the '0 years old' subplot}.
}
    \end{tcolorbox}
    
  \end{paracol}

  \vspace{1em}
  \subsubsection{Our World in Data V2 Outputs}%
  \vspace{0.5em}
  \begin{paracol}{3}
    
    \begin{tcolorbox}[modelcard, title={GPT-5.4 \hfill \normalfont\tiny\color{gray}1 runs}]
      \fbadge{V1}\fbadge{scene\_graph}\fbadge{data\_table}\fbadge{vega\_lite\_spec}\fbadge{l1\_caption}\fbadge{l2\_caption}\imgbadge
      \vspace{3pt}\hrule\vspace{3pt}
      {\sffamily\tiny\bfseries\color{gray!80!black}METADATA COMPLETENESS \hfill Overall 3.7/5}\\[2pt]
      {\sffamily\tiny Data Values \hfill 4/5}\\
      {\sffamily\tiny Trend Magnitude \hfill 4/5}\\
      {\sffamily\tiny Chart Subject Domain \hfill 3/5}
      \vspace{3pt}\hrule\vspace{3pt}
      {\sffamily\tiny
\noindent{\sffamily\tiny\bfseries\color{gray!90!black}CONTEXTUAL INSIGHTS}\\[2pt]
This dataset reports annual death rates for females at several ages: 0, 10, 25, 45, 65, and 80 years old. \tagdirect{} The expert domain story states that the values are annual death rates per 1,000 people of a given age for females. \tagdirect{} The source is Our World in Data via the linked grapher reference. \tagdirect{} The table covers 2023 for countries and regional aggregates such as Africa (UN), Americas, and Asia (UN). \tagdirect{} In the rows shown, death rates are much higher at age 0 and age 80 than at ages 10, 25, or 45. \tagderived{compared age-specific values within the provided rows} The lowest death rates in the shown rows are typically around age 10, after which rates rise again by adulthood and especially older age. \tagderived{read age pattern across provided rows}

\vspace{4pt}\noindent{\sffamily\tiny\bfseries\color{gray!90!black}DOMAIN-SPECIFIC EXPLANATIONS}\\[2pt]
In demography and public health, a death rate of, for example, 41.51804 at age 0 for Afghanistan means about 41.5 deaths per 1,000 female infants aged 0 in a year. \tagderived{interpreted provided unit 'per 1,000 people of a given age' with Afghanistan row value} These are age-specific mortality rates, so values at different ages should not be read as the share of all female deaths, but as the annual risk of death for females within each age group. \tagspeculative{demography/public health interpretation} The strong rise from midlife to older ages in the shown data is consistent with mortality increasing with age. \tagderived{compared values from ages 45, 65, and 80 across provided rows} For example, in the Africa (UN) row, the rate rises from 5.5834303 at age 45 to 23.308329 at age 65 and 96.977356 at age 80. \tagdirect{} In the Americas row, the rate is 9.637899 at age 0, falls to 0.16284001 at age 10, then increases to 45.21309 at age 80. \tagdirect{} Cross-country differences at age 0 are large in the shown rows, ranging from 2.37637 in Austria to 49.50861 in Benin. \tagderived{identified minimum and maximum age-0 values in provided rows}

\vspace{4pt}\noindent{\sffamily\tiny\bfseries\color{gray!90!black}REAL-WORLD SIGNIFICANCE}\\[2pt]
These data matter because they summarize survival conditions for females at different life stages in 2023. \tagdirect{} High death rates at age 0 in several shown countries, such as Benin at 49.50861 and Angola at 45.32285, indicate a much greater mortality burden in infancy than in places like Australia at 3.06751 or Austria at 2.37637. \tagderived{compared age-0 values across provided rows} Differences like these are often used to assess population health, healthcare access, nutrition, sanitation, and maternal-newborn care. \tagspeculative{global health and demography} Higher mortality at older ages, such as 80, is important for planning elder care, pensions, and health services for aging populations. \tagspeculative{public policy and aging} Regional aggregates such as Africa (UN), Asia (UN), and the Americas can help compare broad mortality environments across world regions. \tagdirect{} If mortality remains elevated in infancy or older age, it can signal continuing pressure on health systems and unequal life chances across populations. \tagspeculative{public health and social inequality}

\vspace{4pt}\noindent{\sffamily\tiny\bfseries\color{gray!90!black}OTHER}\\[2pt]
The chart metadata does not provide x-axis or y-axis labels, but the data table itself makes clear that the dimensions are entity, year, and age-specific death rates. \tagdirect{} Only part of the dataset is shown because the table is truncated after Benin with a note that more rows exist. \tagdirect{} Among the displayed rows, age 80 always has the highest death rate and age 10 is usually among the lowest. \tagderived{scanned all provided rows for relative age ordering} Some rows include both countries and regional groupings, so not every entity is a nation-state. \tagdirect{}
}
    \end{tcolorbox}
    
    \switchcolumn
    \begin{tcolorbox}[modelcard, title={llama-4-scout-17... \hfill \normalfont\tiny\color{gray}1 runs}]
      \fbadge{V1}\fbadge{scene\_graph}\fbadge{vega\_lite\_spec}\fbadge{data\_table}\fbadge{l1\_caption}\fbadge{l2\_caption}\imgbadge
      \vspace{3pt}\hrule\vspace{3pt}
      {\sffamily\tiny\bfseries\color{gray!80!black}METADATA COMPLETENESS \hfill Overall 4.7/5}\\[2pt]
      {\sffamily\tiny Data Values \hfill 5/5}\\
      {\sffamily\tiny Trend Magnitude \hfill 4/5}\\
      {\sffamily\tiny Chart Subject Domain \hfill 5/5}
      \vspace{3pt}\hrule\vspace{3pt}
      {\sffamily\tiny
\noindent{\sffamily\tiny\bfseries\color{gray!90!black}CONTEXTUAL INSIGHTS}\\[2pt]
This chart displays annual death rates at different ages in females \tagdirect{}. The data is presented for various countries and age groups, including 0, 10, 25, 45, 65, and 80 years old \tagdirect{}. The chart is likely used to compare death rates across countries and age groups \tagspeculative{general knowledge of demographic statistics}.

\vspace{4pt}\noindent{\sffamily\tiny\bfseries\color{gray!90!black}DOMAIN-SPECIFIC EXPLANATIONS}\\[2pt]
The death rates are measured per 1,000 people of a given age \tagdirect{}. The data shows significant variation in death rates across countries and age groups \tagderived{comparison of data values}. For example, countries like Afghanistan and Angola have high death rates, especially at younger ages \tagderived{comparison of data values}.

\vspace{4pt}\noindent{\sffamily\tiny\bfseries\color{gray!90!black}REAL-WORLD SIGNIFICANCE}\\[2pt]
This data matters for public health policy and resource allocation \tagspeculative{general knowledge of public health importance}. Countries with high death rates, especially at younger ages, may need to prioritize healthcare interventions and resource allocation \tagspeculative{general knowledge of public health importance}. The data can also inform decisions on healthcare infrastructure and emergency preparedness \tagspeculative{general knowledge of public health importance}.

\vspace{4pt}\noindent{\sffamily\tiny\bfseries\color{gray!90!black}OTHER}\\[2pt]
The data does not provide information on the causes of death or underlying health conditions \tagderived{absence of data}. The chart may be used by researchers, policymakers, or healthcare professionals to understand demographic trends and make informed decisions \tagspeculative{general knowledge of demographic statistics}.
}
    \end{tcolorbox}
    
    \switchcolumn
    \begin{tcolorbox}[modelcard, title={Gemini 3.5 Flash \hfill \normalfont\tiny\color{gray}1 runs}]
      \fbadge{V1}\fbadge{scene\_graph}\fbadge{vega\_lite\_spec}\fbadge{data\_table}\fbadge{l1\_caption}\fbadge{l2\_caption}\imgbadge
      \vspace{3pt}\hrule\vspace{3pt}
      {\sffamily\tiny\bfseries\color{gray!80!black}METADATA COMPLETENESS \hfill Overall 4.0/5}\\[2pt]
      {\sffamily\tiny Data Values \hfill 5/5}\\
      {\sffamily\tiny Trend Magnitude \hfill 2/5}\\
      {\sffamily\tiny Chart Subject Domain \hfill 5/5}
      \vspace{3pt}\hrule\vspace{3pt}
      {\sffamily\tiny
\noindent{\sffamily\tiny\bfseries\color{gray!90!black}CONTEXTUAL INSIGHTS}\\[2pt]
The chart displays the annual death rates for females across different age groups (0, 10, 25, 45, 65, and 80 years old) for various countries and regions in the year 2023 \tagdirect{}. The data is sourced from Our World in Data \tagdirect{}. It covers global regions (like Africa (UN), Americas, Asia (UN)) as well as individual nations ranging alphabetically from Afghanistan to Benin \tagdirect{}.

\vspace{4pt}\noindent{\sffamily\tiny\bfseries\color{gray!90!black}DOMAIN-SPECIFIC EXPLANATIONS}\\[2pt]
The values represent the annual death rate per 1,000 females of a specific age group \tagdirect{}. Across all listed entities, mortality rates follow a classic 'U-shaped' or 'J-shaped' curve typical of human demography, where death rates are relatively high at infancy (0 years old), drop to their lowest points around age 10, and then progressively rise through adulthood (25, 45, 65) to their highest levels at age 80 \tagderived{Compared values across age columns for all visible rows, noting that age 10 is consistently the lowest and age 80 is the highest, with age 0 being higher than ages 10-45}. There is a stark contrast between developing and developed nations; for instance, the infant mortality rate (0 years old) in Angola (45.32 per 1,000) or Benin (49.51 per 1,000) is more than 15 times higher than that of Australia (3.07 per 1,000) or Austria (2.38 per 1,000) \tagderived{Calculated ratio of infant mortality rates between Angola/Benin and Australia/Austria}. Similarly, at age 80, a female in Afghanistan has a death rate of 112.09 per 1,000, whereas in Australia it is 28.98 per 1,000 \tagderived{Compared 80-year-old mortality rates between Afghanistan and Australia}.

\vspace{4pt}\noindent{\sffamily\tiny\bfseries\color{gray!90!black}REAL-WORLD SIGNIFICANCE}\\[2pt]
These disparities highlight massive global inequalities in healthcare access, sanitation, maternal care, and socioeconomic development \tagspeculative{Public Health}. High infant mortality rates (0 years old) in regions like Africa (UN) (41.34 per 1,000) compared to the Americas (9.64 per 1,000) \tagderived{Compared infant mortality rates of Africa (UN) and Americas} point to critical gaps in neonatal and pediatric healthcare infrastructure \tagspeculative{Global Health Policy}. Policymakers and international aid organizations use this data to target healthcare interventions, improve maternal-child health programs, and evaluate the effectiveness of public health systems \tagspeculative{Global Health Policy}.

\vspace{4pt}\noindent{\sffamily\tiny\bfseries\color{gray!90!black}OTHER}\\[2pt]
The data is limited to females, meaning it does not capture sex-based differences in mortality \tagspeculative{Demography}. The year of the data is 2023 \tagdirect{}, representing a post-COVID-19 pandemic baseline for global mortality \tagspeculative{Epidemiology}.
}
    \end{tcolorbox}
    
  \end{paracol}

  \vspace{1em}
  \subsubsection{Our World in Data V3 Outputs}%
  \vspace{0.5em}
  \begin{paracol}{3}
    
    \begin{tcolorbox}[modelcard, title={GPT-5.4 \hfill \normalfont\tiny\color{gray}1 runs}]
      \fbadge{V1}\fbadge{scene\_graph}\fbadge{data\_table}\fbadge{vega\_lite\_spec}\fbadge{l1\_caption}\fbadge{l2\_caption}\imgbadge
      \vspace{3pt}\hrule\vspace{3pt}
      {\sffamily\tiny\bfseries\color{gray!80!black}METADATA COMPLETENESS \hfill Overall 3.7/5}\\[2pt]
      {\sffamily\tiny Data Values \hfill 4/5}\\
      {\sffamily\tiny Trend Magnitude \hfill 4/5}\\
      {\sffamily\tiny Chart Subject Domain \hfill 3/5}
      \vspace{3pt}\hrule\vspace{3pt}
      {\sffamily\tiny
\noindent{\sffamily\tiny\bfseries\color{gray!90!black}CONTEXTUAL INSIGHTS}\\[2pt]
This chart shows annual death rates for females at six ages: 0, 10, 25, 45, 65, and 80 years old. \tagdirect{} It covers the period from 1933 to 2023. \tagdirect{} The comparison includes World, India, United States, and Japan. \tagdirect{} The subtitle states that the measure is annual deaths per 1,000 people of a given age. \tagdirect{} The topic belongs to demography and population health. \tagspeculative{demography/public health} The chart appears to have been published by Our World in Data. \tagdirect{} The footer cites Human Mortality Database (2025) and UN, World Population Prospects (2024) as data sources. \tagdirect{} Across all six age panels, death rates generally decline over time in every series that is shown. \tagderived{read overall downward trends across all panels and countries from the plotted lines}

\vspace{4pt}\noindent{\sffamily\tiny\bfseries\color{gray!90!black}DOMAIN-SPECIFIC EXPLANATIONS}\\[2pt]
In demographic terms, an age-specific death rate measures the probability of dying within a year for people who have already reached a particular age, expressed here per 1,000 females at that age. \tagspeculative{demography} The panel for age 0 refers to infant mortality risk within the first year of life. \tagspeculative{demography} The chart indicates that mortality is much higher at age 0 and especially at older ages such as 65 and 80 than at ages 10 or 25. \tagderived{compare vertical scales and line levels across age panels} Age 10 has the lowest death rates among the ages shown for most of the time period. \tagderived{compare line levels across the six age panels} India is generally at the highest level among the named countries in most panels for much of the historical period shown. \tagderived{compare India line against Japan and United States across panels over time} Japan shows especially large long-run improvements from mid-century onward, moving from relatively high death rates to among the lowest by the end of the series in several age groups. \tagderived{compare Japan's earlier and later positions across panels} The United States is usually below the world series and often below India, but by the end of the period it is not always the lowest line, especially at older ages. \tagderived{compare end-of-series positions in age 45, 65, and 80 panels} Temporary spikes near the most recent years are visible in some older-age panels, especially for India at ages 65 and 80 and to a lesser extent in the world series. \tagderived{read recent upward deviations near 2020–2023 in older-age panels} Such late-period spikes in mortality are often associated with acute shocks such as pandemics, severe heat, conflict, or reporting disruptions, but this chart alone does not identify the cause. \tagspeculative{epidemiology/demography}

\vspace{4pt}\noindent{\sffamily\tiny\bfseries\color{gray!90!black}REAL-WORLD SIGNIFICANCE}\\[2pt]
These data matter because age-specific female mortality is a core indicator of survival conditions across the life course. \tagspeculative{public health/demography} The broad decline across decades suggests major improvements in survival for girls and women in the countries and world aggregate shown. \tagderived{infer improvement from long-run downward trends across panels} Falling death rates at age 0 are especially important because infant mortality is widely used as a marker of maternal health, health-system access, nutrition, sanitation, and infectious disease control. \tagspeculative{global health} Declines at ages 45, 65, and 80 have major implications for population aging, pension systems, long-term care demand, and the prevalence of chronic disease. \tagspeculative{demography/health policy} Cross-country gaps imply unequal access to the conditions that reduce mortality, such as vaccination, safer childbirth, antibiotics, cardiovascular treatment, education, and income growth. \tagspeculative{global health/economics} The recent upticks in older-age mortality could signal that progress is vulnerable to shocks and that older women may need targeted protection during crises. \tagderived{infer vulnerability from visible recent spikes in older-age panels}

\vspace{4pt}\noindent{\sffamily\tiny\bfseries\color{gray!90!black}OTHER}\\[2pt]
The chart title in the image is more specific than the metadata title: it states 'Annual death rates at different ages in females, 1933 to 2023.' \tagdirect{} The image provides more information than the accessibility metadata alone, including the time span, the four compared populations, the six age groups, the subtitle definition, and the data sources. \tagdirect{} Because exact numeric values are not supplied in a table, any statement about precise rates or exact differences should be treated as approximate unless read directly from axis labels in the image. \tagderived{assess limitations from missing data table and image-only reading}
}
    \end{tcolorbox}
    
    \switchcolumn
    \begin{tcolorbox}[modelcard, title={llama-4-scout-17... \hfill \normalfont\tiny\color{gray}1 runs}]
      \fbadge{V1}\fbadge{scene\_graph}\fbadge{vega\_lite\_spec}\fbadge{data\_table}\fbadge{l1\_caption}\fbadge{l2\_caption}\imgbadge
      \vspace{3pt}\hrule\vspace{3pt}
      {\sffamily\tiny\bfseries\color{gray!80!black}METADATA COMPLETENESS \hfill Overall 4.7/5}\\[2pt]
      {\sffamily\tiny Data Values \hfill 5/5}\\
      {\sffamily\tiny Trend Magnitude \hfill 4/5}\\
      {\sffamily\tiny Chart Subject Domain \hfill 5/5}
      \vspace{3pt}\hrule\vspace{3pt}
      {\sffamily\tiny
\noindent{\sffamily\tiny\bfseries\color{gray!90!black}CONTEXTUAL INSIGHTS}\\[2pt]
This chart shows the annual death rates at different ages in females from 1933 to 2023 \tagdirect{}. The data is presented for four countries: World, India, United States, and Japan \tagdirect{}.

\vspace{4pt}\noindent{\sffamily\tiny\bfseries\color{gray!90!black}DOMAIN-SPECIFIC EXPLANATIONS}\\[2pt]
The annual death rate is measured per 1,000 people of a given age \tagdirect{}. The chart provides insights into how mortality rates have changed over time for females across different ages and countries \tagderived{from chart title and data lines}.

\vspace{4pt}\noindent{\sffamily\tiny\bfseries\color{gray!90!black}REAL-WORLD SIGNIFICANCE}\\[2pt]
This data matters because it shows trends in mortality rates over time, which can inform public health policies and interventions \tagspeculative{public health domain}. The decline in mortality rates across ages and countries indicates improvements in healthcare, living standards, and possibly disease management \tagspeculative{demographic and health studies}.

\vspace{4pt}\noindent{\sffamily\tiny\bfseries\color{gray!90!black}OTHER}\\[2pt]
The chart does not provide information on the causes of death or the impact of specific diseases on mortality rates \tagderived{from absence of detailed data}.
}
    \end{tcolorbox}
    
    \switchcolumn
    \begin{tcolorbox}[modelcard, title={Gemini 3.5 Flash \hfill \normalfont\tiny\color{gray}1 runs}]
      \fbadge{V1}\fbadge{scene\_graph}\fbadge{vega\_lite\_spec}\fbadge{data\_table}\fbadge{l1\_caption}\fbadge{l2\_caption}\imgbadge
      \vspace{3pt}\hrule\vspace{3pt}
      {\sffamily\tiny\bfseries\color{gray!80!black}METADATA COMPLETENESS \hfill Overall 4.0/5}\\[2pt]
      {\sffamily\tiny Data Values \hfill 5/5}\\
      {\sffamily\tiny Trend Magnitude \hfill 2/5}\\
      {\sffamily\tiny Chart Subject Domain \hfill 5/5}
      \vspace{3pt}\hrule\vspace{3pt}
      {\sffamily\tiny
\noindent{\sffamily\tiny\bfseries\color{gray!90!black}CONTEXTUAL INSIGHTS}\\[2pt]
The chart, titled "Annual death rates at different ages in females", is produced by the publication Our World in Data \tagderived{read from the logo in the top right of the image}. It displays the annual female death rate per 1,000 people of a given age across six specific age groups: 0, 10, 25, 45, 65, and 80 years old, spanning the years 1933 to 2023 \tagdirect{}. The chart compares four geographic entities: the World, India, the United States, and Japan \tagderived{read from the legend}. The data is sourced from the Human Mortality Database (2025) and the UN World Population Prospects (2024) \tagderived{read from the footer of the image}.

\vspace{4pt}\noindent{\sffamily\tiny\bfseries\color{gray!90!black}DOMAIN-SPECIFIC EXPLANATIONS}\\[2pt]
In demography, measuring death rates per 1,000 people of a specific age group allows for standardized comparisons across countries with vastly different population sizes and age structures \tagspeculative{demography}. The data reveals a classic U-shaped mortality curve, where death rates are highest at the extremes of life—infancy (0 years) and old age (80 years)—and lowest during childhood and young adulthood \tagderived{comparing the Y-axis scales of the panels, where 0 and 80 years old reach up to 150 per 1,000, while 10 and 25 years old peak below 10 per 1,000}. A notable global anomaly is the spike around 1959–1961 in the "World" series, most visible in the 10-year-old panel, which historically aligns with the Great Chinese Famine \tagspeculative{historical demography}. Additionally, a sharp spike in mortality is visible in 2020–2021, particularly in the 65 and 80 years old panels, which represents the global impact of the COVID-19 pandemic \tagspeculative{epidemiology}.

\vspace{4pt}\noindent{\sffamily\tiny\bfseries\color{gray!90!black}REAL-WORLD SIGNIFICANCE}\\[2pt]
The consistent downward trend in female mortality across all age groups and regions over the 90-year period represents one of humanity's greatest achievements, driven by improvements in clean water, sanitation, antibiotics, vaccines, and maternal care \tagspeculative{public health history}. Japan's trajectory is particularly remarkable; starting with high mortality rates in the post-WWII era, it rapidly declined to achieve some of the lowest female mortality rates globally, reflecting its rapid economic development and universal healthcare system \tagderived{comparing Japan's series starting high in the late 1940s and ending as the lowest or near-lowest line by 2023}. Conversely, while India has made monumental progress—reducing infant mortality from over 150 per 1,000 in the 1950s to around 25 per 1,000 in 2023—it still lags behind the US and Japan, highlighting persistent global health disparities \tagderived{comparing India's series values in the 0 years old panel from 1950 to 2023 against the US and Japan}. The sudden spike in older age mortality during 2020–2021 serves as a stark reminder of how infectious disease outbreaks can abruptly disrupt long-term health progress \tagderived{observing the sharp upward tick in 2020-2021 in the 65 and 80 years old panels}.
}
    \end{tcolorbox}
    
  \end{paracol}

  \vspace{1em}
  \subsubsection{Our World in Data V4 Outputs}%
  \vspace{0.5em}
  \begin{paracol}{3}
    
    \begin{tcolorbox}[modelcard, title={GPT-5.4 \hfill \normalfont\tiny\color{gray}1 runs}]
      \fbadge{V1}\fbadge{scene\_graph}\fbadge{data\_table}\fbadge{vega\_lite\_spec}\fbadge{l1\_caption}\fbadge{l2\_caption}\imgbadge
      \vspace{3pt}\hrule\vspace{3pt}
      {\sffamily\tiny\bfseries\color{gray!80!black}METADATA COMPLETENESS \hfill Overall 3.7/5}\\[2pt]
      {\sffamily\tiny Data Values \hfill 4/5}\\
      {\sffamily\tiny Trend Magnitude \hfill 4/5}\\
      {\sffamily\tiny Chart Subject Domain \hfill 3/5}
      \vspace{3pt}\hrule\vspace{3pt}
      {\sffamily\tiny
\noindent{\sffamily\tiny\bfseries\color{gray!90!black}CONTEXTUAL INSIGHTS}\\[2pt]
This chart shows annual death rates for females at six ages—0, 10, 25, 45, 65, and 80—from 1933 to 2023, expressed per 1,000 people of a given age. \tagdirect{} The comparison covers the World, India, the United States, and Japan. \tagdirect{} The chart belongs to the demography and public health domain. \tagdirect{} It was published by Our World in Data and cites the Human Mortality Database and the UN World Population Prospects as sources. \tagdirect{} Across all six age panels, death rates generally decline over time in all four geographies. \tagderived{read broad downward trends across all panels} Japan appears to have some of the lowest recent death rates in several adult and older-age panels. \tagderived{compared line endpoints across panels} India starts from notably higher death rates than the United States and Japan in multiple age groups, especially in mid-century portions of the series. \tagderived{compared relative line levels across panels and years}

\vspace{4pt}\noindent{\sffamily\tiny\bfseries\color{gray!90!black}DOMAIN-SPECIFIC EXPLANATIONS}\\[2pt]
In demography, an age-specific death rate measures the probability of dying within a year for people at a particular age, summarized here as deaths per 1,000 females of that age. \tagdirect{} Looking at separate age panels is important because mortality risk differs greatly across the life course, with infancy and old age usually having much higher death rates than childhood or young adulthood. \tagderived{compared value ranges across age panels} The large long-run declines at age 0 are consistent with major reductions in infant mortality. \tagderived{read strong downward trend in the 0-years panel} In public health terms, falling death rates at younger ages are often associated with improved sanitation, vaccination, nutrition, safer childbirth, and infectious-disease control. \tagspeculative{public health} Falling death rates at ages 45, 65, and 80 can reflect better chronic disease prevention and treatment, improved medical care, and broader social development. \tagspeculative{epidemiology} Differences between countries may also reflect varying healthcare access, income, education, disease environments, and the timing of mortality transition. \tagspeculative{demography}

\vspace{4pt}\noindent{\sffamily\tiny\bfseries\color{gray!90!black}REAL-WORLD SIGNIFICANCE}\\[2pt]
These data matter because mortality rates are a core indicator of population health and human development. \tagspeculative{demography/public health} The broad downward trends suggest that female survival improved substantially over the period shown. \tagderived{read long-run declines across all panels} Improvements at age 0 have especially large social significance because infant mortality is widely used as a marker of maternal and child health conditions. \tagspeculative{public health} Lower death rates at working and older ages can affect pension systems, healthcare demand, caregiving needs, and population aging. \tagspeculative{economics/demography} Cross-country gaps matter for policymakers because they can indicate unequal access to health improvements and different stages of epidemiological transition. \tagspeculative{global health/demography} If mortality continues to decline at older ages, societies may need to plan for longer lifespans, more years lived with chronic conditions, and greater demand for long-term care. \tagspeculative{aging policy}

\vspace{4pt}\noindent{\sffamily\tiny\bfseries\color{gray!90!black}OTHER}\\[2pt]
The chart is structured to allow comparison both across time and across ages, which helps distinguish whether mortality improvement was concentrated in childhood, adulthood, or old age. \tagderived{inferred from multi-panel age layout} Some panels show short-term spikes or irregularities rather than perfectly smooth decline, indicating that mortality can be affected by temporary shocks as well as long-run improvement. \tagderived{read local fluctuations/spikes in several panels} Because the evidence comes from the image alone, precise values, exact turning points, and causes of fluctuations should be treated as approximate unless confirmed by the underlying dataset. \tagdirect{}
}
    \end{tcolorbox}
    
    \switchcolumn
    \begin{tcolorbox}[modelcard, title={llama-4-scout-17... \hfill \normalfont\tiny\color{gray}1 runs}]
      \fbadge{V1}\fbadge{scene\_graph}\fbadge{vega\_lite\_spec}\fbadge{data\_table}\fbadge{l1\_caption}\fbadge{l2\_caption}\imgbadge
      \vspace{3pt}\hrule\vspace{3pt}
      {\sffamily\tiny\bfseries\color{gray!80!black}METADATA COMPLETENESS \hfill Overall 4.7/5}\\[2pt]
      {\sffamily\tiny Data Values \hfill 5/5}\\
      {\sffamily\tiny Trend Magnitude \hfill 4/5}\\
      {\sffamily\tiny Chart Subject Domain \hfill 5/5}
      \vspace{3pt}\hrule\vspace{3pt}
      {\sffamily\tiny
\noindent{\sffamily\tiny\bfseries\color{gray!90!black}CONTEXTUAL INSIGHTS}\\[2pt]
This chart shows annual death rates for females at ages 0, 10, 25, 45, 65, and 80 years old across the World, India, United States, and Japan from 1933 to 2023. \tagderived{Observed from chart title and axis labels}. The data is presented per 1,000 people of a given age \tagdirect{}.

\vspace{4pt}\noindent{\sffamily\tiny\bfseries\color{gray!90!black}DOMAIN-SPECIFIC EXPLANATIONS}\\[2pt]
The chart illustrates a significant decline in mortality rates over the years for all ages and countries \tagderived{From the downward trend in all lines}. For example, at age 0, the death rate has dramatically decreased, especially in India and the United States \tagderived{From the steep decline in green and brown lines}.

\vspace{4pt}\noindent{\sffamily\tiny\bfseries\color{gray!90!black}REAL-WORLD SIGNIFICANCE}\\[2pt]
The declining death rates indicate improvements in healthcare, sanitation, and living standards \tagspeculative{Based on general knowledge of public health}. This trend suggests increased life expectancy and better health outcomes for females globally and in the specified countries \tagderived{From the context of mortality rates and age}.

\vspace{4pt}\noindent{\sffamily\tiny\bfseries\color{gray!90!black}OTHER}\\[2pt]
The data source is the Human Mortality Database and UN World Population Prospects \tagdirect{}. The chart is produced by Our World in Data \tagdirect{}.
}
    \end{tcolorbox}
    
    \switchcolumn
    \begin{tcolorbox}[modelcard, title={Gemini 3.5 Flash \hfill \normalfont\tiny\color{gray}1 runs}]
      \fbadge{V1}\fbadge{scene\_graph}\fbadge{vega\_lite\_spec}\fbadge{data\_table}\fbadge{l1\_caption}\fbadge{l2\_caption}\imgbadge
      \vspace{3pt}\hrule\vspace{3pt}
      {\sffamily\tiny\bfseries\color{gray!80!black}METADATA COMPLETENESS \hfill Overall 4.0/5}\\[2pt]
      {\sffamily\tiny Data Values \hfill 5/5}\\
      {\sffamily\tiny Trend Magnitude \hfill 2/5}\\
      {\sffamily\tiny Chart Subject Domain \hfill 5/5}
      \vspace{3pt}\hrule\vspace{3pt}
      {\sffamily\tiny
\noindent{\sffamily\tiny\bfseries\color{gray!90!black}CONTEXTUAL INSIGHTS}\\[2pt]
The chart shows annual female death rates per 1,000 people of specific ages (0, 10, 25, 45, 65, and 80 years old) from 1933 to 2023 \tagdirect{}. It covers four geographic entities: World, India, United States, and Japan \tagdirect{}. The data is sourced from the Human Mortality Database (2025) and the UN World Population Prospects (2024), published by Our World in Data \tagdirect{}. The purpose is to track historical progress in reducing age-specific mortality rates for females globally and in key representative nations \tagderived{analyzed the selection of age groups and countries spanning different development levels}.

\vspace{4pt}\noindent{\sffamily\tiny\bfseries\color{gray!90!black}DOMAIN-SPECIFIC EXPLANATIONS}\\[2pt]
Death rates are expressed as deaths per 1,000 individuals of a given age group \tagdirect{}. Infant mortality (0 years old) and elderly mortality (80 years old) have much higher baseline rates compared to youth and middle-age groups (10, 25, 45 years old) \tagderived{compared y-axis scales across subplots, where 0 and 80 years old go up to 150, while 10 years old peaks around 4}. The dramatic decline in infant mortality (0 years old) across all regions reflects global advancements in sanitation, vaccination, obstetrics, and pediatric medicine over the 20th and 21st centuries \tagspeculative{Public Health History}. The prominent spike in the 'World' line around 1960, especially visible in the 10-year-old and 0-year-old cohorts, likely reflects the demographic impact of the Great Chinese Famine (1959–1961), which significantly altered global averages due to China's large population \tagspeculative{Historical Demography}. The sharp spike in mortality rates around 2020–2021 in the older age groups (65 and 80 years old), particularly visible in India, the US, and the global average, corresponds to the excess mortality caused by the COVID-19 pandemic \tagspeculative{Epidemiology / COVID-19 pandemic}.

\vspace{4pt}\noindent{\sffamily\tiny\bfseries\color{gray!90!black}REAL-WORLD SIGNIFICANCE}\\[2pt]
The long-term decline in female mortality across all age groups signifies a massive improvement in global life expectancy and quality of life \tagderived{observed the downward trend across all six panels from 1933 to 2023}. The convergence of mortality rates in 2023 compared to the wide disparities in the mid-20th century shows that developing nations like India have rapidly closed the gap with developed nations like the US and Japan in terms of basic healthcare and survival rates \tagderived{compared the spread between India/World and US/Japan in 1950 versus 2023 across all panels}. The vulnerability of older populations to emerging infectious diseases is highlighted by the prominent 2020-2021 spikes in the 65 and 80-year-old cohorts, demonstrating that public health gains remain susceptible to global pandemics \tagderived{observed that the COVID-era spike is highly pronounced in the 65 and 80 age groups but barely visible in the 10 and 25 age groups}. These trends are crucial for policymakers, pension planners, and healthcare systems to project future demographic structures, as declining mortality rates lead to aging populations that require different social safety nets \tagspeculative{Demography and Public Policy}.

\vspace{4pt}\noindent{\sffamily\tiny\bfseries\color{gray!90!black}OTHER}\\[2pt]
Japan consistently shows some of the lowest female mortality rates across almost all age groups in the latter half of the 20th century and early 21st century, reflecting its status as a global leader in life expectancy \tagderived{observed the red line (Japan) consistently tracking at or near the bottom of the charts from the 1970s onward}.
}
    \end{tcolorbox}
    
  \end{paracol}

\clearpage
\subsection{HCI Alt Text Dataset Example}
\label{supp:hci-example}

\begin{figure}[H]
  \centering
  \begin{minipage}[c]{0.48\linewidth}
    \centering
    \includegraphics[
      width=\linewidth,
      alt={Representative HCI Alt Text Dataset figure used across V1 through V4. The example combines a chart image with author-written alt text, title, and source information.}
    ]{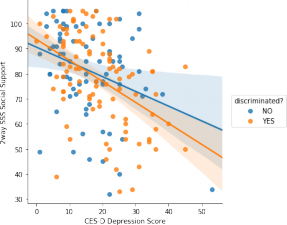}
    \caption{Representative visualization from the HCI Alt Text Dataset.}
    \label{fig:hci-example}
  \end{minipage}%
  \hfill
  \begin{minipage}[c]{0.48\linewidth}
    \centering
    \small
    \renewcommand{\arraystretch}{1.2}
    \begin{tabularx}{\linewidth}{@{}lX@{}}
    \toprule
    \textbf{Field} & \textbf{Content used in this example} \\
    \midrule
    Image & Included in V1, V3, and V4 \\
    Author-written alt text & Included in V1 and V2 \\
    Title & Included in V1 and V2 \\
    Source URL & Included in V1 and V2 \\
    \bottomrule
    \end{tabularx}
    \vspace{4pt}
    \captionof{table}{Accessibility metadata fields available across input variants for the HCI Alt Text Dataset example.}
    \label{tab:hci-fields}
  \end{minipage}
\end{figure}

Figure~\ref{fig:hci-example} presents a representative visualization from the HCI Alt Text Dataset, while Table~\ref{tab:hci-fields} summarizes the accessibility metadata fields provided across input conditions V1--V4.

  \vspace{1em}
  \subsubsection{HCI Alt Text V1 Outputs}%
  \vspace{0.5em}
  \begin{paracol}{3}
    
    \begin{tcolorbox}[modelcard, title={GPT-5.4 \hfill \normalfont\tiny\color{gray}1 runs}]
      \fbadge{V1}\fbadge{scene\_graph}\fbadge{data\_table}\fbadge{vega\_lite\_spec}\fbadge{l1\_caption}\fbadge{l2\_caption}\imgbadge
      \vspace{3pt}\hrule\vspace{3pt}
      {\sffamily\tiny\bfseries\color{gray!80!black}METADATA COMPLETENESS \hfill Overall 3.7/5}\\[2pt]
      {\sffamily\tiny Data Values \hfill 4/5}\\
      {\sffamily\tiny Trend Magnitude \hfill 4/5}\\
      {\sffamily\tiny Chart Subject Domain \hfill 3/5}
      \vspace{3pt}\hrule\vspace{3pt}
      {\sffamily\tiny
\noindent{\sffamily\tiny\bfseries\color{gray!90!black}CONTEXTUAL INSIGHTS}\\[2pt]
This chart compares the relationship between depression score and social support for two groups: people who reported discrimination and people who did not \tagdirect{}. The figure comes from a paper published in Proc. ACM Hum. Comput. Interact. in 2019 \tagdirect{}. The subject matter places the chart in a research area overlapping mental health, social support, discrimination, and human-computer interaction or social computing \tagspeculative{academic domain classification}. From the image, both fitted lines slope downward, indicating that higher depression scores are associated with lower social support in both groups \tagderived{read direction of both correlation lines from the chart image}. The two-group comparison suggests the authors were interested in whether reported discrimination changes the relationship between social support and depression \tagderived{inferred analytic purpose from grouped scatter plot and separate trend lines}.

\vspace{4pt}\noindent{\sffamily\tiny\bfseries\color{gray!90!black}DOMAIN-SPECIFIC EXPLANATIONS}\\[2pt]
In mental health research, depression scores are commonly used as indicators of symptom severity, where higher scores usually represent more depressive symptoms \tagspeculative{mental health measurement}. Social support generally refers to perceived or received help, belonging, or reliable support from other people, and higher values typically indicate stronger support networks \tagspeculative{social and clinical psychology}. A negative association between depression and social support is consistent with a pattern in which people reporting more depressive symptoms also report less support \tagderived{interpreted downward trend lines}. If the group-specific lines differ in steepness or vertical position, that can indicate either different average support levels between groups or different strengths of association between depression and support \tagspeculative{statistics and regression interpretation}. The chart appears to include substantial scatter around each line, which means the relationship is not perfect and individual experiences vary widely within both groups \tagderived{read spread of points around the fitted lines from the chart image}. Because no axis definitions, scale documentation, or model summary are provided, this figure should be interpreted as showing association rather than proving that discrimination causes changes in depression or support \tagspeculative{epidemiology and statistical inference}.

\vspace{4pt}\noindent{\sffamily\tiny\bfseries\color{gray!90!black}REAL-WORLD SIGNIFICANCE}\\[2pt]
This topic matters because discrimination, social support, and depression are all socially consequential factors linked to well-being and quality of life \tagspeculative{public health and social psychology}. If people who report discrimination also experience weaker social support or a different depression-support relationship, that could signal added vulnerability for affected groups \tagspeculative{mental health disparities research}. Findings like this can be relevant to designers, clinicians, and community organizations who want to identify protective factors that may buffer mental health harms \tagspeculative{HCI, clinical practice, and community health}. In a social and political context, evidence connecting discrimination with mental health-related outcomes can inform discussions about equity, inclusion, and the downstream effects of biased treatment \tagspeculative{public policy and social epidemiology}. Because the source is an academic paper rather than an administrative dashboard, the chart is more likely intended to support research interpretation than direct operational decision-making \tagderived{inferred likely purpose from publication venue and source description}.

\vspace{4pt}\noindent{\sffamily\tiny\bfseries\color{gray!90!black}OTHER}\\[2pt]
The image alone suggests the chart contains two categories labeled by discrimination status, one for people who did not report discrimination and one for people who did \tagderived{read legend categories from the chart image}. The exact sample size for each group is not available from the provided materials \tagdirect{}. The exact statistical method used for the lines, such as linear regression or another smoother, is not stated in the metadata \tagdirect{}. Without the paper text, it is not possible to determine the population studied, how discrimination was measured, or whether the analysis controlled for confounding variables \tagdirect{}.
}

    \end{tcolorbox}
    
    \switchcolumn
    \begin{tcolorbox}[modelcard, title={llama-4-scout-17... \hfill \normalfont\tiny\color{gray}1 runs}]
      \fbadge{V1}\fbadge{scene\_graph}\fbadge{vega\_lite\_spec}\fbadge{data\_table}\fbadge{l1\_caption}\fbadge{l2\_caption}\imgbadge
      \vspace{3pt}\hrule\vspace{3pt}
      {\sffamily\tiny\bfseries\color{gray!80!black}METADATA COMPLETENESS \hfill Overall 4.7/5}\\[2pt]
      {\sffamily\tiny Data Values \hfill 5/5}\\
      {\sffamily\tiny Trend Magnitude \hfill 4/5}\\
      {\sffamily\tiny Chart Subject Domain \hfill 5/5}
      \vspace{3pt}\hrule\vspace{3pt}
      {\sffamily\tiny
\noindent{\sffamily\tiny\bfseries\color{gray!90!black}CONTEXTUAL INSIGHTS}\\[2pt]
This chart appears to be a scatter plot showing the relationship between social support and depression scores for individuals who have reported discrimination and those who have not \tagderived{from the chart image and alt text}. The chart is likely from a study on the impact of discrimination on mental health \tagspeculative{based on the expert domain story and common research areas}.

\vspace{4pt}\noindent{\sffamily\tiny\bfseries\color{gray!90!black}DOMAIN-SPECIFIC EXPLANATIONS}\\[2pt]
The chart suggests that there is a negative correlation between social support and depression scores, meaning that higher social support is associated with lower depression scores \tagdirect{}. The data points are differentiated by whether individuals have reported discrimination or not, indicating that discrimination may have an impact on this relationship \tagderived{from the chart image}.

\vspace{4pt}\noindent{\sffamily\tiny\bfseries\color{gray!90!black}REAL-WORLD SIGNIFICANCE}\\[2pt]
This data likely matters for understanding how discrimination affects mental health and the role of social support in mitigating depression \tagspeculative{based on common knowledge in psychology and social sciences}. Individuals who have experienced discrimination may have lower social support and higher depression scores, highlighting the need for targeted interventions \tagderived{from the chart image and common knowledge in the field}.

\vspace{4pt}\noindent{\sffamily\tiny\bfseries\color{gray!90!black}OTHER}\\[2pt]
The chart does not provide specific data values, but it visually represents a clear trend \tagdirect{}. The regression lines for both groups (those who have reported discrimination and those who have not) show a similar negative slope, but are separated, suggesting a consistent impact of discrimination on mental health \tagderived{from the chart image}.
}
    \end{tcolorbox}
    
    \switchcolumn
    \begin{tcolorbox}[modelcard, title={Gemini 3.5 Flash \hfill \normalfont\tiny\color{gray}1 runs}]
      \fbadge{V1}\fbadge{scene\_graph}\fbadge{vega\_lite\_spec}\fbadge{data\_table}\fbadge{l1\_caption}\fbadge{l2\_caption}\imgbadge
      \vspace{3pt}\hrule\vspace{3pt}
      {\sffamily\tiny\bfseries\color{gray!80!black}METADATA COMPLETENESS \hfill Overall 4.0/5}\\[2pt]
      {\sffamily\tiny Data Values \hfill 5/5}\\
      {\sffamily\tiny Trend Magnitude \hfill 2/5}\\
      {\sffamily\tiny Chart Subject Domain \hfill 5/5}
      \vspace{3pt}\hrule\vspace{3pt}
      {\sffamily\tiny
\noindent{\sffamily\tiny\bfseries\color{gray!90!black}CONTEXTUAL INSIGHTS}\\[2pt]
This chart is a scatter plot with correlation lines examining the relationship between social support ('2way SSS Social Support') and depression ('CES-D Depression Score') \tagdirect{}. It categorizes participants based on whether they reported experiencing discrimination ('YES' or 'NO') \tagdirect{}. The data originates from a research paper published in the \textit{Proceedings of the ACM on Human-Computer Interaction} in 2019 \tagdirect{}, a venue that frequently explores how technology, online communities, and social systems impact human well-being \tagspeculative{Human-Computer Interaction / Social Computing}.

\vspace{4pt}\noindent{\sffamily\tiny\bfseries\color{gray!90!black}DOMAIN-SPECIFIC EXPLANATIONS}\\[2pt]
In clinical and social psychology, 'CES-D' refers to the Center for Epidemiologic Studies Depression Scale, where higher scores indicate more severe depressive symptoms \tagspeculative{Clinical Psychology}. 'SSS' typically denotes a Social Support Scale, where higher scores represent stronger perceived support networks \tagspeculative{Social Psychology}. The negative slope of both regression lines indicates that higher perceived social support is associated with lower depression scores \tagderived{Observed the downward slope of both regression lines in the chart image}. Notably, the slope for individuals who experienced discrimination ('YES') is steeper than for those who did not ('NO') \tagderived{Compared the slopes of the two regression lines in the chart image}. This suggests that social support may play an even more critical protective role—or its absence may be more severely felt—among individuals facing discrimination \tagspeculative{Social Psychology}.

\vspace{4pt}\noindent{\sffamily\tiny\bfseries\color{gray!90!black}REAL-WORLD SIGNIFICANCE}\\[2pt]
These findings are highly significant for mental health professionals, community organizers, and platform designers who build online support groups \tagspeculative{Mental Health \& HCI}. For marginalized populations who frequently face discrimination, fostering robust social support systems is a vital buffer against depression \tagspeculative{Social Psychology}. If discrimination persists and social support systems are weak, individuals are at a substantially higher risk for severe depressive episodes \tagspeculative{Clinical Psychology}. Consequently, interventions should focus not only on reducing discrimination but also on actively strengthening peer support networks \tagspeculative{Public Health}.

\vspace{4pt}\noindent{\sffamily\tiny\bfseries\color{gray!90!black}OTHER}\\[2pt]
The confidence intervals (shaded regions around the lines) widen significantly at higher depression scores, particularly above 40, indicating fewer data points and greater statistical uncertainty at extreme levels of depression \tagderived{Observed the widening shaded confidence bands on the right side of the chart image}.
}
    \end{tcolorbox}
    
  \end{paracol}

  \vspace{1em}
  \subsubsection{HCI Alt Text V2 Outputs}%
  \vspace{0.5em}
  \begin{paracol}{3}
    
    \begin{tcolorbox}[modelcard, title={GPT-5.4 \hfill \normalfont\tiny\color{gray}1 runs}]
      \fbadge{V1}\fbadge{scene\_graph}\fbadge{data\_table}\fbadge{vega\_lite\_spec}\fbadge{l1\_caption}\fbadge{l2\_caption}\imgbadge
      \vspace{3pt}\hrule\vspace{3pt}
      {\sffamily\tiny\bfseries\color{gray!80!black}METADATA COMPLETENESS \hfill Overall 3.7/5}\\[2pt]
      {\sffamily\tiny Data Values \hfill 4/5}\\
      {\sffamily\tiny Trend Magnitude \hfill 4/5}\\
      {\sffamily\tiny Chart Subject Domain \hfill 3/5}
      \vspace{3pt}\hrule\vspace{3pt}
      {\sffamily\tiny
\noindent{\sffamily\tiny\bfseries\color{gray!90!black}CONTEXTUAL INSIGHTS}\\[2pt]
This chart concerns the relationship between social support and depression, separated into people who reported discrimination and people who did not. \tagdirect{} The chart type is a scatter plot with a correlation line. \tagdirect{} The topic sits at the intersection of mental health and discrimination research. \tagspeculative{public health and social science domain knowledge} The source is identified only as a paper published in Proc. ACM Hum. Comput. Interact. in 2019. \tagdirect{} Because no title, axes, or data values are provided, the chart's exact measures, units, and sample cannot be determined from the metadata alone. \tagdirect{}

\vspace{4pt}\noindent{\sffamily\tiny\bfseries\color{gray!90!black}DOMAIN-SPECIFIC EXPLANATIONS}\\[2pt]
In mental health research, depression usually refers to self-reported symptom burden or a score from a screening instrument rather than a clinical diagnosis unless explicitly stated otherwise. \tagspeculative{clinical psychology and epidemiology domain knowledge} Social support commonly refers to perceived or received emotional, informational, or practical support from others. \tagspeculative{social psychology domain knowledge} Comparing respondents who reported discrimination with those who did not suggests the study is examining whether discrimination is associated with different depression levels, different social support levels, or a different relationship between the two. \tagderived{inferred analytic purpose from the two-group comparison named in alt text} A correlation line usually indicates that the authors wanted readers to assess the direction of association between social support and depression within the plotted data. \tagspeculative{statistics/data visualization domain knowledge} However, the metadata does not reveal whether the relationship is positive or negative, whether the groups differ substantially, or whether any observed association is statistically significant. \tagdirect{}

\vspace{4pt}\noindent{\sffamily\tiny\bfseries\color{gray!90!black}REAL-WORLD SIGNIFICANCE}\\[2pt]
The chart addresses issues relevant to people who experience discrimination and to researchers studying mental health disparities. \tagderived{read from the named variables and group split in alt text} In real-world terms, this kind of analysis is often used to ask whether supportive social relationships may buffer psychological distress among marginalized groups. \tagspeculative{public health and minority stress research} If that is the study's aim, the findings could matter for interventions in counseling, community support design, workplace policy, or anti-discrimination efforts. \tagspeculative{mental health services and policy domain knowledge} Because the actual trend is not provided, no conclusion can be drawn here about whether social support appears protective, whether discrimination is linked to higher depression, or whether the two groups differ in slope or level. \tagdirect{}

\vspace{4pt}\noindent{\sffamily\tiny\bfseries\color{gray!90!black}OTHER}\\[2pt]
The metadata supports only a high-level semantic summary, not a substantive interpretation of results. \tagdirect{} The presence of correlation lines for two discrimination-status groups implies at least two subgroup-specific relationships were intended to be compared. \tagderived{read from alt text mentioning discrimination and non-discrimination groups with correlation line} Any stronger statement about effect size, causality, or practical impact would go beyond the evidence supplied here. \tagdirect{}
}
    \end{tcolorbox}
    
    \switchcolumn
    \begin{tcolorbox}[modelcard, title={llama-4-scout-17... \hfill \normalfont\tiny\color{gray}1 runs}]
      \fbadge{V1}\fbadge{scene\_graph}\fbadge{vega\_lite\_spec}\fbadge{data\_table}\fbadge{l1\_caption}\fbadge{l2\_caption}\imgbadge
      \vspace{3pt}\hrule\vspace{3pt}
      {\sffamily\tiny\bfseries\color{gray!80!black}METADATA COMPLETENESS \hfill Overall 4.7/5}\\[2pt]
      {\sffamily\tiny Data Values \hfill 5/5}\\
      {\sffamily\tiny Trend Magnitude \hfill 4/5}\\
      {\sffamily\tiny Chart Subject Domain \hfill 5/5}
      \vspace{3pt}\hrule\vspace{3pt}
      {\sffamily\tiny
\noindent{\sffamily\tiny\bfseries\color{gray!90!black}CONTEXTUAL INSIGHTS}\\[2pt]
This chart appears to be related to the field of social sciences or psychology, specifically exploring the relationship between social support and depression \tagspeculative{general knowledge of social sciences}. The chart is likely from a research paper or study, given the mention of a Proc. ACM Hum. Comput. Interact. (2019) publication \tagdirect{}.

\vspace{4pt}\noindent{\sffamily\tiny\bfseries\color{gray!90!black}DOMAIN-SPECIFIC EXPLANATIONS}\\[2pt]
The chart seems to be analyzing the correlation between social support and depression for two groups: people who have reported discrimination and those who have not \tagdirect{}. The data likely represents a statistical analysis of the relationship between these variables \tagspeculative{general knowledge of statistical analysis}.

\vspace{4pt}\noindent{\sffamily\tiny\bfseries\color{gray!90!black}REAL-WORLD SIGNIFICANCE}\\[2pt]
This data may have implications for understanding the impact of discrimination on mental health and the role of social support in mitigating depression \tagspeculative{general knowledge of social sciences}. The findings could inform policies or interventions aimed at supporting marginalized groups \tagspeculative{general knowledge of policy-making}.

\vspace{4pt}\noindent{\sffamily\tiny\bfseries\color{gray!90!black}OTHER}\\[2pt]
The chart's results could be useful for researchers, policymakers, or practitioners working in the field of mental health, social work, or sociology \tagspeculative{general knowledge of field}.
}
    \end{tcolorbox}
    
    \switchcolumn
    \begin{tcolorbox}[modelcard, title={Gemini 3.5 Flash \hfill \normalfont\tiny\color{gray}1 runs}]
      \fbadge{V1}\fbadge{scene\_graph}\fbadge{vega\_lite\_spec}\fbadge{data\_table}\fbadge{l1\_caption}\fbadge{l2\_caption}\imgbadge
      \vspace{3pt}\hrule\vspace{3pt}
      {\sffamily\tiny\bfseries\color{gray!80!black}METADATA COMPLETENESS \hfill Overall 4.0/5}\\[2pt]
      {\sffamily\tiny Data Values \hfill 5/5}\\
      {\sffamily\tiny Trend Magnitude \hfill 2/5}\\
      {\sffamily\tiny Chart Subject Domain \hfill 5/5}
      \vspace{3pt}\hrule\vspace{3pt}
      {\sffamily\tiny
\noindent{\sffamily\tiny\bfseries\color{gray!90!black}CONTEXTUAL INSIGHTS}\\[2pt]
The chart is a scatter plot with correlation lines examining the relationship between social support and depression \tagdirect{}. It compares two distinct groups: individuals who have reported experiencing discrimination and those who have not \tagdirect{}. This research was published in the Proceedings of the ACM on Human-Computer Interaction in 2019 \tagdirect{}. The study likely explores how online or offline social support systems mitigate mental health challenges like depression, particularly in the context of marginalized groups facing discrimination \tagspeculative{Human-Computer Interaction / Mental Health}.

\vspace{4pt}\noindent{\sffamily\tiny\bfseries\color{gray!90!black}DOMAIN-SPECIFIC EXPLANATIONS}\\[2pt]
In psychological and sociological research, 'social support' is typically hypothesized to have a buffering effect against mental health issues like depression \tagspeculative{Psychology}. A negative correlation (downward sloping line) is generally expected between social support and depression, meaning higher support correlates with lower depression \tagspeculative{Psychology}. By splitting the data into those who reported discrimination and those who did not, researchers can analyze if discrimination moderates the relationship between social support and depression \tagderived{analyzed the comparison of the two groups from the alt text}.

\vspace{4pt}\noindent{\sffamily\tiny\bfseries\color{gray!90!black}REAL-WORLD SIGNIFICANCE}\\[2pt]
Understanding how social support interacts with discrimination and depression can help design better digital intervention tools, online support communities, or mental health platforms \tagspeculative{Human-Computer Interaction}. If social support is less effective at buffering depression for those facing discrimination, it suggests a need for specialized, identity-affirming support systems rather than generic ones \tagspeculative{Mental Health Advocacy}.

\vspace{4pt}\noindent{\sffamily\tiny\bfseries\color{gray!90!black}OTHER}\\[2pt]
Without the visual chart or data table, the exact strength, direction, and statistical significance of these correlations cannot be verified \tagderived{noted the absence of data values and visual trends}.
}
    \end{tcolorbox}
    
  \end{paracol}

  \vspace{1em}
  \subsubsection{HCI Alt Text V3 Outputs}%
  \vspace{0.5em}
  \begin{paracol}{3}
    
    \begin{tcolorbox}[modelcard, title={GPT-5.4 \hfill \normalfont\tiny\color{gray}1 runs}]
      \fbadge{V1}\fbadge{scene\_graph}\fbadge{data\_table}\fbadge{vega\_lite\_spec}\fbadge{l1\_caption}\fbadge{l2\_caption}\imgbadge
      \vspace{3pt}\hrule\vspace{3pt}
      {\sffamily\tiny\bfseries\color{gray!80!black}METADATA COMPLETENESS \hfill Overall 3.7/5}\\[2pt]
      {\sffamily\tiny Data Values \hfill 4/5}\\
      {\sffamily\tiny Trend Magnitude \hfill 4/5}\\
      {\sffamily\tiny Chart Subject Domain \hfill 3/5}
      \vspace{3pt}\hrule\vspace{3pt}
      {\sffamily\tiny
\noindent{\sffamily\tiny\bfseries\color{gray!90!black}CONTEXTUAL INSIGHTS}\\[2pt]
This appears to be a scatter plot relating CES-D Depression Score to Social Support, with points divided into two groups labeled by discrimination status, YES and NO. \tagderived{read visible axis text and legend from image} The chart appears to belong to a mental health, psychology, or public health context because depression score, social support, and discrimination are common variables in those fields. \tagspeculative{mental health and public health domain knowledge} The fitted lines suggest that higher depression scores are associated with lower social support in both groups. \tagderived{read direction of fitted lines from image} The chart likely aims to compare whether the relationship between depression and social support differs by discrimination status. \tagderived{inferred from grouped scatter plot with separate trend lines and legend categories}

\vspace{4pt}\noindent{\sffamily\tiny\bfseries\color{gray!90!black}DOMAIN-SPECIFIC EXPLANATIONS}\\[2pt]
CES-D usually refers to the Center for Epidemiologic Studies Depression scale, a screening instrument used to measure depressive symptoms. \tagspeculative{psychological measurement knowledge} Social support generally refers to perceived or received emotional, informational, or practical support from others. \tagspeculative{psychology and public health domain knowledge} In behavioral and public health research, an inverse relationship between depression scores and social support is often interpreted as meaning that people with more depressive symptoms also report less support, though the chart alone does not establish causation. \tagspeculative{epidemiology and psychology methods knowledge} The presence of separate fitted lines for discrimination groups suggests the analyst may be examining whether discrimination modifies, mediates, or is associated with differences in the depression-support relationship. \tagspeculative{social epidemiology methods knowledge} The point cloud shows substantial spread around the lines, indicating that the relationship is not perfectly predictive from the image alone. \tagderived{read dispersion of points around fitted lines from image}

\vspace{4pt}\noindent{\sffamily\tiny\bfseries\color{gray!90!black}REAL-WORLD SIGNIFICANCE}\\[2pt]
If this pattern reflects a real underlying association, it could matter for mental health screening and support interventions because people with higher depressive symptoms may also be experiencing weaker support networks. \tagspeculative{public health and clinical practice knowledge} If discrimination is part of the analytic framing, the chart may be relevant to equity-focused policy or community interventions aimed at reducing social stressors and improving mental health outcomes. \tagspeculative{social epidemiology and health equity knowledge} Researchers, clinicians, counselors, and policymakers could use this kind of evidence to identify groups that may need additional outreach or culturally responsive support services. \tagspeculative{public health and mental health services knowledge} However, without the study design, sample, and statistics, no firm conclusion should be made about the strength, significance, or cause of the observed association. \tagdirect{}

\vspace{4pt}\noindent{\sffamily\tiny\bfseries\color{gray!90!black}OTHER}\\[2pt]
No chart title, source, date, sample description, exact values, or statistical annotations are provided in the metadata. \tagdirect{} Because the task provides image only and sparse metadata, any interpretation beyond the visible axis text, legend, and general trend should be treated cautiously. \tagdirect{} The image suggests two category-specific trend lines and a shaded uncertainty band, but the exact confidence interval meaning is not stated in the provided metadata. \tagderived{read grouped trend lines and shaded band from image}
}
    \end{tcolorbox}
    
    \switchcolumn
    \begin{tcolorbox}[modelcard, title={llama-4-scout-17... \hfill \normalfont\tiny\color{gray}1 runs}]
      \fbadge{V1}\fbadge{scene\_graph}\fbadge{vega\_lite\_spec}\fbadge{data\_table}\fbadge{l1\_caption}\fbadge{l2\_caption}\imgbadge
      \vspace{3pt}\hrule\vspace{3pt}
      {\sffamily\tiny\bfseries\color{gray!80!black}METADATA COMPLETENESS \hfill Overall 4.7/5}\\[2pt]
      {\sffamily\tiny Data Values \hfill 5/5}\\
      {\sffamily\tiny Trend Magnitude \hfill 4/5}\\
      {\sffamily\tiny Chart Subject Domain \hfill 5/5}
      \vspace{3pt}\hrule\vspace{3pt}
      {\sffamily\tiny
\noindent{\sffamily\tiny\bfseries\color{gray!90!black}CONTEXTUAL INSIGHTS}\\[2pt]
This chart appears to be a scatterplot showing the relationship between two variables \tagderived{from the chart image}. The x-axis represents the CES-D Depression Score \tagderived{from the chart image}, and the y-axis represents the Average SCS social support \tagderived{from the chart image}. There are two groups of data points, one represented by blue dots and the other by orange dots, which seem to indicate a difference in social support between individuals who have been discriminated against and those who have not \tagderived{from the legend}.

\vspace{4pt}\noindent{\sffamily\tiny\bfseries\color{gray!90!black}DOMAIN-SPECIFIC EXPLANATIONS}\\[2pt]
The CES-D Depression Score is a measure of depression severity \tagspeculative{from general knowledge of psychology}. The Average SCS social support likely measures the level of social support an individual receives \tagspeculative{from general knowledge of psychology}. The chart suggests that there is a negative correlation between depression score and social support \tagderived{from the chart image}, meaning that as depression scores increase, social support tends to decrease \tagderived{from the chart image}.

\vspace{4pt}\noindent{\sffamily\tiny\bfseries\color{gray!90!black}REAL-WORLD SIGNIFICANCE}\\[2pt]
This chart may be used to study the relationship between depression and social support in individuals who have experienced discrimination \tagspeculative{from general knowledge of social sciences}. The data could inform policies or interventions aimed at providing support to individuals with depression \tagspeculative{from general knowledge of social sciences}.

\vspace{4pt}\noindent{\sffamily\tiny\bfseries\color{gray!90!black}OTHER}\\[2pt]
The chart includes a regression line for each group, indicating a stronger negative correlation for the group that has been discriminated against \tagderived{from the chart image}.
}
    \end{tcolorbox}
    
    \switchcolumn
    \begin{tcolorbox}[modelcard, title={Gemini 3.5 Flash \hfill \normalfont\tiny\color{gray}1 runs}]
      \fbadge{V1}\fbadge{scene\_graph}\fbadge{vega\_lite\_spec}\fbadge{data\_table}\fbadge{l1\_caption}\fbadge{l2\_caption}\imgbadge
      \vspace{3pt}\hrule\vspace{3pt}
      {\sffamily\tiny\bfseries\color{gray!80!black}METADATA COMPLETENESS \hfill Overall 4.0/5}\\[2pt]
      {\sffamily\tiny Data Values \hfill 5/5}\\
      {\sffamily\tiny Trend Magnitude \hfill 2/5}\\
      {\sffamily\tiny Chart Subject Domain \hfill 5/5}
      \vspace{3pt}\hrule\vspace{3pt}
      {\sffamily\tiny
\noindent{\sffamily\tiny\bfseries\color{gray!90!black}CONTEXTUAL INSIGHTS}\\[2pt]
The chart is a scatter plot with linear regression lines illustrating the relationship between depression scores and perceived social support, categorized by whether participants experienced discrimination \tagderived{read from chart axes and legend}. The x-axis represents the 'CES-D Depression Score' and the y-axis represents '2way SSS Social Support' \tagderived{read from chart axes}. CES-D is a standard acronym for the Center for Epidemiologic Studies Depression Scale, which is a widely used self-report measure of depressive symptoms \tagspeculative{Clinical Psychology}.

\vspace{4pt}\noindent{\sffamily\tiny\bfseries\color{gray!90!black}DOMAIN-SPECIFIC EXPLANATIONS}\\[2pt]
Both regression lines show a negative correlation, meaning that higher depression scores are associated with lower perceived social support \tagderived{observed negative slopes of both regression lines}. The regression line for individuals who experienced discrimination ('YES') has a steeper negative slope than the line for those who did not ('NO') \tagderived{compared slopes of the 'YES' and 'NO' regression lines}. This suggests that for individuals experiencing discrimination, the decline in perceived social support is more pronounced as depressive symptoms increase \tagderived{interpreted the steeper slope of the 'YES' group}. The shaded bands represent confidence intervals, which widen significantly at higher depression scores (above 40) due to a smaller sample size of individuals with very high depression scores in this dataset \tagderived{observed widening of shaded bands on the right side of the plot}.

\vspace{4pt}\noindent{\sffamily\tiny\bfseries\color{gray!90!black}REAL-WORLD SIGNIFICANCE}\\[2pt]
This data is highly relevant to mental health researchers and public health policymakers studying the social determinants of mental health \tagspeculative{Public Health}. The steeper decline in social support for discriminated individuals highlights how systemic biases can compound mental health challenges, potentially making recovery more difficult \tagspeculative{Sociology/Mental Health}. Interventions aiming to alleviate depression may need to incorporate specialized social support strategies for individuals who face discrimination, as their natural support systems may erode more quickly under stress \tagspeculative{Clinical Psychology}.

\vspace{4pt}\noindent{\sffamily\tiny\bfseries\color{gray!90!black}OTHER}\\[2pt]
The majority of data points are clustered at depression scores below 30 and social support scores above 60 \tagderived{observed density of scatter points in the upper-left quadrant}.
}
    \end{tcolorbox}
    
  \end{paracol}

  \vspace{1em}
  \subsubsection{HCI Alt Text V4 Outputs}%
  \vspace{0.5em}
  \begin{paracol}{3}
    
    \begin{tcolorbox}[modelcard, title={GPT-5.4 \hfill \normalfont\tiny\color{gray}1 runs}]
      \fbadge{V1}\fbadge{scene\_graph}\fbadge{data\_table}\fbadge{vega\_lite\_spec}\fbadge{l1\_caption}\fbadge{l2\_caption}\imgbadge
      \vspace{3pt}\hrule\vspace{3pt}
      {\sffamily\tiny\bfseries\color{gray!80!black}METADATA COMPLETENESS \hfill Overall 3.7/5}\\[2pt]
      {\sffamily\tiny Data Values \hfill 4/5}\\
      {\sffamily\tiny Trend Magnitude \hfill 4/5}\\
      {\sffamily\tiny Chart Subject Domain \hfill 3/5}
      \vspace{3pt}\hrule\vspace{3pt}
      {\sffamily\tiny
\noindent{\sffamily\tiny\bfseries\color{gray!90!black}CONTEXTUAL INSIGHTS}\\[2pt]
This chart examines the relationship between CES-D Depression Score and Social Support, split by whether someone is marked as discriminated. \tagdirect{} The broad domain appears to be mental health or social/behavioral health research. \tagspeculative{public health and psychology domain knowledge} The plotted trend lines indicate that higher depression scores are associated with lower social support in both groups. \tagderived{reading the negative slope of both fitted lines} The two groups labeled YES and NO suggest the chart is comparing people by reported discrimination status. \tagdirect{} The image alone does not identify who produced the chart or the study purpose beyond showing this comparison. \tagdirect{}

\vspace{4pt}\noindent{\sffamily\tiny\bfseries\color{gray!90!black}DOMAIN-SPECIFIC EXPLANATIONS}\\[2pt]
CES-D commonly refers to the Center for Epidemiologic Studies Depression scale, a screening measure for depressive symptoms. \tagspeculative{psychology and epidemiology domain knowledge} Social support usually refers to perceived or available emotional, practical, or interpersonal support, though the exact instrument is not given here. \tagspeculative{social science measurement knowledge} In this chart, the negative association means that people with higher depression scores tend to also have lower social support scores. \tagderived{reading the cross-variable downward trends} The fitted lines appear separated, with the YES group tending to lie below the NO group at similar depression-score levels, implying lower social support among those marked YES across much of the range. \tagderived{comparing the vertical positions of the two fitted lines across the x-axis} The confidence bands indicate uncertainty around the fitted trends, but exact statistical significance cannot be determined from the image alone. \tagderived{presence of shaded uncertainty bands without numerical statistics}

\vspace{4pt}\noindent{\sffamily\tiny\bfseries\color{gray!90!black}REAL-WORLD SIGNIFICANCE}\\[2pt]
If this pattern reflects a real population relationship, it suggests that depressive symptoms, social support, and discrimination may be interconnected. \tagspeculative{public health and social epidemiology knowledge} People experiencing discrimination may face added social or psychological burdens that are associated with reduced support or worse mental health outcomes. \tagspeculative{discrimination and mental health literature} This matters for clinicians, counselors, public health practitioners, and community organizations because interventions may need to address both mental health symptoms and social conditions rather than treating them separately. \tagspeculative{mental health services and public health practice} If the observed negative relationship continues, groups with higher depressive symptoms could remain at elevated risk of social isolation or unmet support needs. \tagspeculative{mental health and social support research} The chart could be relevant to decisions about screening, outreach, anti-discrimination policy, and support-program targeting, but those uses are not stated in the image itself. \tagspeculative{health policy and social services knowledge}

\vspace{4pt}\noindent{\sffamily\tiny\bfseries\color{gray!90!black}OTHER}\\[2pt]
The scatter suggests substantial variability around the trend lines, so the relationship is not perfectly deterministic. \tagderived{observing wide spread of points around the fitted lines} The visible x-axis range appears to run from roughly 0 to the mid-50s for depression score, and the y-axis from roughly 30 to just above 100 for social support, but these are only approximate readings from the image. \tagderived{reading approximate axis tick ranges from the chart image} Exact sample size, subgroup sizes, model type, and statistical controls are not available from the image alone. \tagdirect{}
}
    \end{tcolorbox}
    
    \switchcolumn
    \begin{tcolorbox}[modelcard, title={llama-4-scout-17... \hfill \normalfont\tiny\color{gray}1 runs}]
      \fbadge{V1}\fbadge{scene\_graph}\fbadge{vega\_lite\_spec}\fbadge{data\_table}\fbadge{l1\_caption}\fbadge{l2\_caption}\imgbadge
      \vspace{3pt}\hrule\vspace{3pt}
      {\sffamily\tiny\bfseries\color{gray!80!black}METADATA COMPLETENESS \hfill Overall 4.7/5}\\[2pt]
      {\sffamily\tiny Data Values \hfill 5/5}\\
      {\sffamily\tiny Trend Magnitude \hfill 4/5}\\
      {\sffamily\tiny Chart Subject Domain \hfill 5/5}
      \vspace{3pt}\hrule\vspace{3pt}
      {\sffamily\tiny
\noindent{\sffamily\tiny\bfseries\color{gray!90!black}CONTEXTUAL INSIGHTS}\\[2pt]
This chart appears to show the relationship between social support and depression scores \tagderived{from chart title and axis labels}. The chart is likely from a study on mental health \tagspeculative{general knowledge of social sciences}. The purpose of the chart is to visualize the correlation between social support and depression scores \tagderived{from chart title and axis labels}.

\vspace{4pt}\noindent{\sffamily\tiny\bfseries\color{gray!90!black}DOMAIN-SPECIFIC EXPLANATIONS}\\[2pt]
The chart shows that as depression scores increase, social support tends to decrease \tagdirect{}. The regression lines indicate a negative correlation between social support and depression scores \tagdirect{}. The blue and orange dots represent data points for individuals who have not been discriminated against and those who have, respectively \tagdirect{}.

\vspace{4pt}\noindent{\sffamily\tiny\bfseries\color{gray!90!black}REAL-WORLD SIGNIFICANCE}\\[2pt]
This data suggests that individuals with higher depression scores tend to have lower social support \tagderived{from chart}. This relationship may be important for mental health professionals and policymakers to understand when developing support services \tagspeculative{knowledge of mental health services}. If this trend continues, it may indicate a need for increased support services for individuals with high depression scores \tagspeculative{general knowledge of social sciences}.

\vspace{4pt}\noindent{\sffamily\tiny\bfseries\color{gray!90!black}OTHER}\\[2pt]
The chart includes a legend indicating that the data points are colored based on whether the individual has been discriminated against \tagdirect{}.
}
    \end{tcolorbox}
    
    \switchcolumn
    \begin{tcolorbox}[modelcard, title={Gemini 3.5 Flash \hfill \normalfont\tiny\color{gray}1 runs}]
      \fbadge{V1}\fbadge{scene\_graph}\fbadge{vega\_lite\_spec}\fbadge{data\_table}\fbadge{l1\_caption}\fbadge{l2\_caption}\imgbadge
      \vspace{3pt}\hrule\vspace{3pt}
      {\sffamily\tiny\bfseries\color{gray!80!black}METADATA COMPLETENESS \hfill Overall 4.0/5}\\[2pt]
      {\sffamily\tiny Data Values \hfill 5/5}\\
      {\sffamily\tiny Trend Magnitude \hfill 2/5}\\
      {\sffamily\tiny Chart Subject Domain \hfill 5/5}
      \vspace{3pt}\hrule\vspace{3pt}
      {\sffamily\tiny
\noindent{\sffamily\tiny\bfseries\color{gray!90!black}CONTEXTUAL INSIGHTS}\\[2pt]
The chart is a scatter plot with linear regression lines showing the relationship between depression scores and social support, grouped by experiences of discrimination \tagderived{read from axes, legend, and plot elements}. The X-axis represents the 'CES-D Depression Score' and the Y-axis represents '2way SSS Social Support' \tagdirect{}. The data is split into two groups based on whether the participants experienced discrimination ('YES' or 'NO') \tagdirect{}. This chart likely comes from a psychological or sociological study examining the buffering effects of social support on depression, or how discrimination impacts these variables \tagspeculative{Psychology/Sociology literature}.

\vspace{4pt}\noindent{\sffamily\tiny\bfseries\color{gray!90!black}DOMAIN-SPECIFIC EXPLANATIONS}\\[2pt]
The CES-D (Center for Epidemiologic Studies Depression Scale) is a widely used self-report measure where higher scores indicate greater depressive symptomatology, typically ranging from 0 to 60 \tagspeculative{Clinical Psychology}. The '2way SSS' likely refers to the Two-Way Social Support Scale, which measures both receiving and giving social support \tagspeculative{Psychometrics}. Both groups ('YES' and 'NO' for discrimination) show a negative correlation: higher depression scores are associated with lower social support \tagderived{observed negative slopes of both regression lines}. The regression line for those who experienced discrimination ('YES') has a steeper negative slope compared to those who did not ('NO'), suggesting that the decline in social support as depression increases is more pronounced among those experiencing discrimination \tagderived{comparison of the slopes of the orange and blue lines}.

\vspace{4pt}\noindent{\sffamily\tiny\bfseries\color{gray!90!black}REAL-WORLD SIGNIFICANCE}\\[2pt]
This data highlights the critical intersection of discrimination, social support, and mental health \tagspeculative{Public Health}. For individuals experiencing discrimination, maintaining high social support may be more difficult or more critical, as indicated by the steeper decline in the orange regression line \tagderived{interpretation of the steeper orange slope}. Mental health professionals and policymakers can use this to design targeted interventions that bolster social support networks specifically for marginalized groups facing discrimination \tagspeculative{Social Policy}.

\vspace{4pt}\noindent{\sffamily\tiny\bfseries\color{gray!90!black}OTHER}\\[2pt]
The confidence intervals (shaded regions) widen significantly at higher CES-D scores, indicating fewer data points and greater uncertainty in the relationship at extreme depression levels \tagderived{observation of the widening shaded bands at the right side of the plot}.
}
    \end{tcolorbox}
    
  \end{paracol}

\clearpage
\subsection{Olli Gallery Example}
\label{supp:olli-example}

\begin{figure}[H]
  \centering
  \begin{minipage}[c]{0.48\linewidth}
    \centering
    \includegraphics[
      width=\linewidth,
      alt={Representative Olli Gallery visualization used across V1 through V4. The example combines a chart image with an Olli screen-reader tree, Vega-Lite specification, title, and source URL.}
    ]{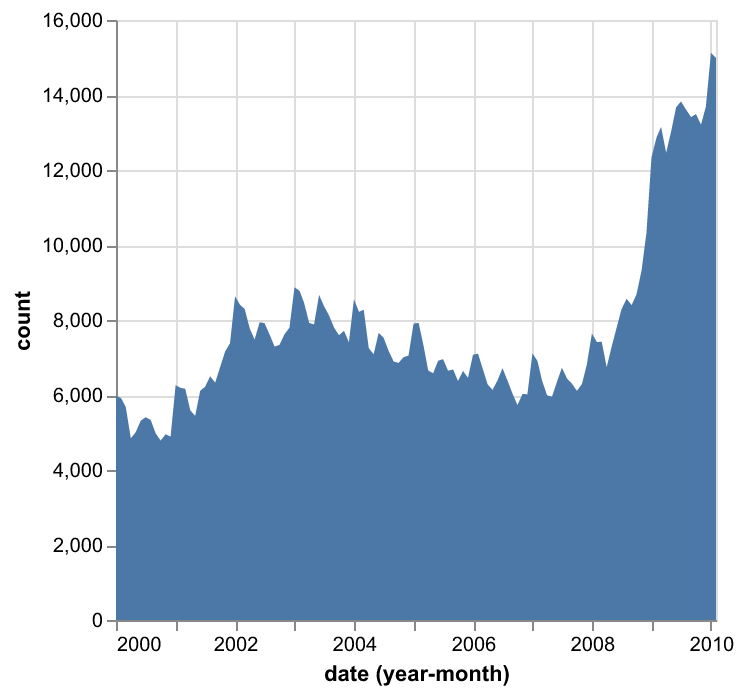}
    \caption{Representative Olli Gallery visualization.}
    \label{fig:olli-example}
  \end{minipage}%
  \hfill
  \begin{minipage}[c]{0.48\linewidth}
    \centering
    \small
    \renewcommand{\arraystretch}{1.2}
    \begin{tabularx}{\linewidth}{@{}lX@{}}
    \toprule
    \textbf{Field} & \textbf{Content used in this example} \\
    \midrule
    Image & Included in V1, V3, and V4 \\
    Olli screen-reader tree & Included in V1 and V2 \\
    Vega-Lite specification & Included in V1 and V2 \\
    Title & Included in V1 and V2 \\
    Source URL & Included in V1 and V2 \\
    \bottomrule
    \end{tabularx}
    \vspace{4pt}
    \captionof{table}{Accessibility metadata fields available across input variants for the Olli Gallery example.}
    \label{tab:olli-fields}
  \end{minipage}
\end{figure}

Figure~\ref{fig:olli-example} shows a representative visualization from the Olli Gallery, while Table~\ref{tab:olli-fields} details the accessibility metadata fields provided across input variants V1--V4.

  \vspace{1em}
  \subsubsection{Olli Gallery V1 Outputs}%
  \vspace{0.5em}
  \begin{paracol}{3}
    
    \begin{tcolorbox}[modelcard, title={GPT-5.4 \hfill \normalfont\tiny\color{gray}1 runs}]
      \fbadge{V1}\fbadge{scene\_graph}\fbadge{data\_table}\fbadge{vega\_lite\_spec}\fbadge{l1\_caption}\fbadge{l2\_caption}\imgbadge
      \vspace{3pt}\hrule\vspace{3pt}
      {\sffamily\tiny\bfseries\color{gray!80!black}METADATA COMPLETENESS \hfill Overall 3.7/5}\\[2pt]
      {\sffamily\tiny Data Values \hfill 4/5}\\
      {\sffamily\tiny Trend Magnitude \hfill 4/5}\\
      {\sffamily\tiny Chart Subject Domain \hfill 3/5}
      \vspace{3pt}\hrule\vspace{3pt}
      {\sffamily\tiny
\noindent{\sffamily\tiny\bfseries\color{gray!90!black}CONTEXTUAL INSIGHTS}\\[2pt]
This chart is about unemployment across industries over time. \tagdirect{} The x-axis is date in year-month format and runs from Jan 2000 to Feb 2010. \tagdirect{} The y-axis is count, with values ranging from 4,786 to 15,125. \tagdirect{} The expert domain story says the measure is unemployment across industries, summed over time. \tagdirect{} The OLLI screen reader tree reports average summed counts of 6,894 for Jan 2000 to Dec 2003, 7,194 for Dec 2003 to Dec 2005, 6,417 for Dec 2005 to Dec 2007, and 11,029 for Dec 2007 to Feb 2010. \tagdirect{} Comparing those period averages shows the final period is much higher than the earlier three periods. \tagderived{compared period-average summed counts from the OLLI screen reader tree} The chart therefore captures a relatively moderate level of summed unemployment through most of the 2000s followed by a pronounced increase near the end of the series. \tagderived{inferred overall pattern from period averages, y-axis range, and chart image} The source link indicates this chart comes from the OLLI gallery example set rather than a fully documented public statistical release in this metadata packet. \tagdirect{}

\vspace{4pt}\noindent{\sffamily\tiny\bfseries\color{gray!90!black}DOMAIN-SPECIFIC EXPLANATIONS}\\[2pt]
In labor-market terms, unemployment counts track how many people are out of work, and summing across industries means this chart is aggregating unemployment from multiple sectors into one total. \tagspeculative{labor economics} Because the y-axis is labeled only as count, the values should be interpreted as absolute numbers rather than percentages or unemployment rates. \tagdirect{} That matters because a count can rise either because job losses increase or because the covered population is larger, while a rate would standardize by labor force size. \tagspeculative{labor economics} The reported averages suggest that unemployment was fairly stable to moderately elevated from 2000 through 2007, then shifted to a substantially higher level after Dec 2007. \tagderived{compared the four period-average summed counts} The average for Dec 2007 to Feb 2010, 11,029, is about 72\% higher than the average for Dec 2005 to Dec 2007, 6,417. \tagderived{computed percentage increase between adjacent period averages} The maximum value of 15,125 is more than three times the minimum value of 4,786. \tagderived{compared y-axis minimum and maximum from metadata} In employment analysis, a broad rise in unemployment across industries usually indicates economy-wide stress rather than a problem confined to a single sector. \tagspeculative{labor economics}

\vspace{4pt}\noindent{\sffamily\tiny\bfseries\color{gray!90!black}REAL-WORLD SIGNIFICANCE}\\[2pt]
This data matters because unemployment affects household income, financial security, and demand for public assistance. \tagspeculative{labor economics and public policy} A rise that is broad enough to appear in unemployment summed across industries suggests many kinds of workers may be affected at once. \tagspeculative{labor economics} Based on the metadata, the period after Dec 2007 had the highest average unemployment and contains many of the highest values in the series. \tagderived{used final-period average of 11,029 and the metadata bin showing 15 values between 10,000 and 15,125} If a pattern like this were continuing, policymakers would typically watch for pressure on unemployment insurance systems, retraining programs, and fiscal support needs. \tagspeculative{public policy} Employers, workers, and governments all use unemployment trends to assess labor-market weakness and recovery timing. \tagspeculative{labor economics} Because stakeholder context is not provided, any claim about the exact intended audience or decision process for this specific chart would go beyond the supplied metadata. \tagdirect{}

\vspace{4pt}\noindent{\sffamily\tiny\bfseries\color{gray!90!black}OTHER}\\[2pt]
There are 122 time values in total across the four reported x-axis segments: 48, 24, 24, and 26. \tagderived{summed the value counts from the four OLLI time segments} The overall average summed count across the full series is 7,740. \tagdirect{} Most observations, 107 of 122, fall below 10,000, while 15 observations fall between 10,000 and 15,125. \tagdirect{} That means roughly 88\% of the observations are below 10,000 and about 12\% are at or above 10,000. \tagderived{computed shares from the metadata frequency counts} This distribution supports the interpretation that the highest unemployment levels are concentrated in a relatively short part of the time series rather than spread evenly across the decade. \tagderived{inferred concentration from the frequency counts plus the higher final-period average} The chart metadata does not specify geography, seasonality treatment, or whether these are persons, claims, or another unemployment-related count. \tagdirect{}
}
    \end{tcolorbox}
    
    \switchcolumn
    \begin{tcolorbox}[modelcard, title={llama-4-scout-17... \hfill \normalfont\tiny\color{gray}1 runs}]
      \fbadge{V1}\fbadge{scene\_graph}\fbadge{vega\_lite\_spec}\fbadge{data\_table}\fbadge{l1\_caption}\fbadge{l2\_caption}\imgbadge
      \vspace{3pt}\hrule\vspace{3pt}
      {\sffamily\tiny\bfseries\color{gray!80!black}METADATA COMPLETENESS \hfill Overall 4.7/5}\\[2pt]
      {\sffamily\tiny Data Values \hfill 5/5}\\
      {\sffamily\tiny Trend Magnitude \hfill 4/5}\\
      {\sffamily\tiny Chart Subject Domain \hfill 5/5}
      \vspace{3pt}\hrule\vspace{3pt}
      {\sffamily\tiny
\noindent{\sffamily\tiny\bfseries\color{gray!90!black}CONTEXTUAL INSIGHTS}\\[2pt]
This chart displays the total volume of unemployment across various industries over a decade, specifically from January 2000 to February 2010 \tagdirect{}. The data tracks monthly fluctuations in the number of unemployed individuals \tagderived{inferred from 'yearmonth' frequency and 'count' metric}. The timeline covers the economic boom of the early 2000s, the subsequent stability, and the severe downturn associated with the 2008 global financial crisis \tagspeculative{historical economic timeline alignment}.

\vspace{4pt}\noindent{\sffamily\tiny\bfseries\color{gray!90!black}DOMAIN-SPECIFIC EXPLANATIONS}\\[2pt]
The values represent the absolute number of unemployed people summed across all sectors, rather than a percentage rate \tagderived{based on the 'sum count' terminology in the metadata}. The data indicates that prior to late 2007, unemployment levels generally fluctuated between roughly 5,000 and 9,000 \tagderived{calculated from the first three OLLI segments showing averages of 6894, 7194, and 6417}. A structural break occurs around late 2007/early 2008, where the baseline unemployment count rises sharply, averaging over 11,000 individuals in the final period \tagderived{calculated from the final OLLI segment average of 11029}. The peak unemployment reached approximately 15,125 individuals \tagdirect{}.

\vspace{4pt}\noindent{\sffamily\tiny\bfseries\color{gray!90!black}REAL-WORLD SIGNIFICANCE}\\[2pt]
The sharp rise in unemployment starting in late 2007 signals a major economic contraction, likely corresponding to the 'Great Recession' \tagspeculative{historical knowledge}. This data is critical for policymakers to assess the severity of labor market distress and to justify stimulus interventions or unemployment benefits expansion \tagspeculative{standard economic policy response}. The fact that the count remains high through 2010 indicates a slow recovery phase where job losses outpaced new hires for an extended period \tagderived{trend observation from the sustained high numbers in the final segment}.

\vspace{4pt}\noindent{\sffamily\tiny\bfseries\color{gray!90!black}OTHER}\\[2pt]
The data shows a 'double-dip' or volatile pattern in the years 2001-2002, suggesting a prior recessionary event (likely the dot-com bubble burst), followed by a recovery before the 2008 crash \tagspeculative{historical economic context}. The chart highlights the cumulative nature of the metric, meaning an increase in one industry is not offset by a decrease in another; it represents the total burden on the labor market \tagderived{interpretation of 'summed over time'}.
}
    \end{tcolorbox}
    
    \switchcolumn
    \begin{tcolorbox}[modelcard, title={Gemini 3.5 Flash \hfill \normalfont\tiny\color{gray}1 runs}]
      \fbadge{V1}\fbadge{scene\_graph}\fbadge{vega\_lite\_spec}\fbadge{data\_table}\fbadge{l1\_caption}\fbadge{l2\_caption}\imgbadge
      \vspace{3pt}\hrule\vspace{3pt}
      {\sffamily\tiny\bfseries\color{gray!80!black}METADATA COMPLETENESS \hfill Overall 4.0/5}\\[2pt]
      {\sffamily\tiny Data Values \hfill 5/5}\\
      {\sffamily\tiny Trend Magnitude \hfill 2/5}\\
      {\sffamily\tiny Chart Subject Domain \hfill 5/5}
      \vspace{3pt}\hrule\vspace{3pt}
      {\sffamily\tiny
\noindent{\sffamily\tiny\bfseries\color{gray!90!black}CONTEXTUAL INSIGHTS}\\[2pt]
The chart is an area chart representing unemployment across industries summed over time, spanning from January 2000 to February 2010 \tagdirect{}. The data tracks the total count of unemployed individuals over a ten-year period \tagderived{read from axes and Olli tree}. This dataset is commonly used in data visualization galleries (such as the Olli gallery) to demonstrate accessible chart representations \tagdirect{}.

\vspace{4pt}\noindent{\sffamily\tiny\bfseries\color{gray!90!black}DOMAIN-SPECIFIC EXPLANATIONS}\\[2pt]
In macroeconomic terms, unemployment counts reflect the health of the labor market \tagspeculative{Macroeconomics}. The time period from Jan 2000 to Feb 2010 covers two major economic downturns in the United States: the early 2000s recession (following the dot-com bubble burst) and the Great Recession of 2007–2009 \tagspeculative{US Economic History}. The dramatic spike in the count starting around late 2008 corresponds directly to the onset of the Great Recession, where the average sum count of unemployed individuals rose to 11,029 between Dec 2007 and Feb 2010, compared to 6,417 in the preceding two years \tagderived{compared average sum counts across time intervals in the Olli tree}.

\vspace{4pt}\noindent{\sffamily\tiny\bfseries\color{gray!90!black}REAL-WORLD SIGNIFICANCE}\\[2pt]
High unemployment rates during the 2008-2010 period led to widespread financial distress, requiring significant government intervention such as fiscal stimulus packages and extended unemployment benefits \tagspeculative{Public Policy}. Understanding these trends helps policymakers, economists, and labor organizations evaluate the severity of economic crises and plan recovery strategies \tagspeculative{Labor Economics}. The sharp upward trajectory at the end of the chart (reaching a peak count of over 15,000 by early 2010) highlights the lagging nature of labor market recovery relative to financial market stabilization \tagderived{observed peak value of 15,125 and the upward trend at the end of the chart}.

\vspace{4pt}\noindent{\sffamily\tiny\bfseries\color{gray!90!black}OTHER}\\[2pt]
The chart's Y-axis ranges from 4,786 to 15,125, indicating that even at its lowest point in this decade, a substantial baseline of unemployment existed \tagderived{read Y-axis range from Olli tree}.
}
    \end{tcolorbox}
    
  \end{paracol}

  \vspace{1em}
  \subsubsection{Olli Gallery V2 Outputs}%
  \vspace{0.5em}
  \begin{paracol}{3}
    
    \begin{tcolorbox}[modelcard, title={GPT-5.4 \hfill \normalfont\tiny\color{gray}1 runs}]
      \fbadge{V1}\fbadge{scene\_graph}\fbadge{data\_table}\fbadge{vega\_lite\_spec}\fbadge{l1\_caption}\fbadge{l2\_caption}\imgbadge
      \vspace{3pt}\hrule\vspace{3pt}
      {\sffamily\tiny\bfseries\color{gray!80!black}METADATA COMPLETENESS \hfill Overall 3.7/5}\\[2pt]
      {\sffamily\tiny Data Values \hfill 4/5}\\
      {\sffamily\tiny Trend Magnitude \hfill 4/5}\\
      {\sffamily\tiny Chart Subject Domain \hfill 3/5}
      \vspace{3pt}\hrule\vspace{3pt}
      {\sffamily\tiny
\noindent{\sffamily\tiny\bfseries\color{gray!90!black}CONTEXTUAL INSIGHTS}\\[2pt]
This chart is about unemployment across industries over time. \tagdirect{} The x-axis is date in year-month form, spanning from Jan 2000 to Feb 2010. \tagdirect{} The y-axis is count, ranging from 4786 to 15125. \tagdirect{} The expert domain story says the chart shows unemployment across industries summed over time. \tagdirect{} Based on the title, axis labels, and domain story, this is a time-series summary of unemployment counts rather than a breakdown by individual industry. \tagderived{combined chart title, axes, and expert domain story} The metadata does not identify a government agency, research institution, or stakeholder responsible for producing the chart. \tagdirect{}

\vspace{4pt}\noindent{\sffamily\tiny\bfseries\color{gray!90!black}DOMAIN-SPECIFIC EXPLANATIONS}\\[2pt]
In labor-market data, unemployment counts usually refer to the number of people without work who are classified as unemployed during each reporting period. \tagspeculative{labor economics/statistical practice} Because the chart is described as unemployment across industries summed over time, the values likely represent a total across multiple industry categories for each month rather than separate industry lines. \tagderived{interpreted from 'across industries' plus 'summed over time' and the single count axis} The overall average sum count is 7740 across the displayed period. \tagdirect{} The period from Dec 2007 to Feb 2010 has an average sum count of 11029, which is higher than the averages reported for Jan 2000 to Dec 2003 (6894), Dec 2003 to Dec 2005 (7194), and Dec 2005 to Dec 2007 (6417). \tagderived{compared the four period averages in the screen reader tree} The highest reported period average occurs in the final segment, Dec 2007 to Feb 2010. \tagderived{identified maximum among period averages} Most observations fall below 10000, with 107 values between 4786 and 10000, while only 15 values fall between 10000 and 15125. \tagdirect{} This indicates that very high unemployment totals are concentrated in a relatively small part of the timeline. \tagderived{inferred from 15 of 122 observations being in the upper band}

\vspace{4pt}\noindent{\sffamily\tiny\bfseries\color{gray!90!black}REAL-WORLD SIGNIFICANCE}\\[2pt]
Unemployment totals matter because they indicate how many people may be struggling to find work at a given time. \tagspeculative{labor economics/public policy} The elevated average in Dec 2007 to Feb 2010 suggests labor-market conditions were worse in that final period than in the earlier periods shown. \tagderived{compared period averages across the timeline} If this chart reflects a real economy-wide pattern, such an increase could correspond to a period of economic stress affecting workers across multiple industries. \tagspeculative{macroeconomics} Higher unemployment can influence decisions about job-search assistance, unemployment benefits, workforce training, and economic stabilization policy. \tagspeculative{public policy/labor economics} People directly affected would include unemployed workers, their households, employers facing weak demand, and policymakers monitoring labor-market conditions. \tagspeculative{labor economics/public policy} Because stakeholder context is not provided, any specific decision-making audience beyond general labor-market users cannot be identified from the metadata alone. \tagdirect{}

\vspace{4pt}\noindent{\sffamily\tiny\bfseries\color{gray!90!black}OTHER}\\[2pt]
The time span covers 122 monthly values from Jan 2000 to Feb 2010. \tagderived{counted from the four x-axis segment totals 48+24+24+26} The minimum and maximum reported counts differ by 10339. \tagderived{subtracted 4786 from 15125} The average sum count declines from 7194 in Dec 2003 to Dec 2005 to 6417 in Dec 2005 to Dec 2007, then rises sharply to 11029 in Dec 2007 to Feb 2010. \tagderived{compared sequential period averages} Without the underlying data table, it is not possible to identify the exact month of the peak or trough beyond the overall y-axis range. \tagdirect{}
}
    \end{tcolorbox}
    
    \switchcolumn
    \begin{tcolorbox}[modelcard, title={llama-4-scout-17... \hfill \normalfont\tiny\color{gray}1 runs}]
      \fbadge{V1}\fbadge{scene\_graph}\fbadge{vega\_lite\_spec}\fbadge{data\_table}\fbadge{l1\_caption}\fbadge{l2\_caption}\imgbadge
      \vspace{3pt}\hrule\vspace{3pt}
      {\sffamily\tiny\bfseries\color{gray!80!black}METADATA COMPLETENESS \hfill Overall 4.7/5}\\[2pt]
      {\sffamily\tiny Data Values \hfill 5/5}\\
      {\sffamily\tiny Trend Magnitude \hfill 4/5}\\
      {\sffamily\tiny Chart Subject Domain \hfill 5/5}
      \vspace{3pt}\hrule\vspace{3pt}
      {\sffamily\tiny
\noindent{\sffamily\tiny\bfseries\color{gray!90!black}CONTEXTUAL INSIGHTS}\\[2pt]
This chart visualizes the total number of unemployed individuals across various industries in the United States from January 2000 to February 2010 \tagdirect{}. The data is aggregated monthly, showing the sum of unemployment counts rather than rates or percentages \tagderived{axis label 'count' and expert story 'summed over time'}. The chart spans a decade that includes the dot-com bubble burst, the post-9/11 recession, and the Great Recession \tagspeculative{historical knowledge of US economic periods between 2000 and 2010}.

\vspace{4pt}\noindent{\sffamily\tiny\bfseries\color{gray!90!black}DOMAIN-SPECIFIC EXPLANATIONS}\\[2pt]
The Y-axis represents the absolute count of unemployed persons, ranging from approximately 4,786 to 15,125 \tagdirect{OLLI tree Y-axis range}. Note: These values appear to be in thousands or represent a specific subset/sample, as total US unemployment is typically much higher (in millions); however, based strictly on the provided metadata, the values are treated as raw counts \tagspeculative{general knowledge of US labor statistics scales vs provided data magnitude}. The data is segmented into four distinct periods in the metadata: Jan 2000–Dec 2003, Dec 2003–Dec 2005, Dec 2005–Dec 2007, and Dec 2007–Feb 2010 \tagdirect{}. The average unemployment count was lowest in the Dec 2005–Dec 2007 period (avg 6,417) and highest in the Dec 2007–Feb 2010 period (avg 11,029) \tagderived{comparison of average sums provided in OLLI tree}.

\vspace{4pt}\noindent{\sffamily\tiny\bfseries\color{gray!90!black}REAL-WORLD SIGNIFICANCE}\\[2pt]
The sharp increase in average unemployment count starting in late 2007 (rising from an average of 6,417 to 11,029) correlates with the onset of the Great Recession \tagspeculative{historical correlation of 2007-2009 economic crisis}. This trend indicates a period of significant job loss and economic instability affecting workers across industries \tagspeculative{general economic interpretation of rising unemployment}. Policy makers and economists use such data to assess the severity of recessions and the effectiveness of stimulus packages \tagspeculative{standard use of unemployment data in macroeconomic policy}. For individuals, this data reflects periods of high difficulty in finding employment \tagspeculative{general social impact of unemployment}.

\vspace{4pt}\noindent{\sffamily\tiny\bfseries\color{gray!90!black}OTHER}\\[2pt]
The data source is identified as 'umwelt-data.github.io/olli/gallery', suggesting this may be a sample dataset for testing accessibility tools rather than primary government statistical release \tagdirect{source URL provided in expert domain story}. The chart uses an area chart format, which emphasizes the volume or magnitude of the total unemployment over time \tagdirect{chart title and type}.
}
    \end{tcolorbox}
    
    \switchcolumn
    \begin{tcolorbox}[modelcard, title={Gemini 3.5 Flash \hfill \normalfont\tiny\color{gray}1 runs}]
      \fbadge{V1}\fbadge{scene\_graph}\fbadge{vega\_lite\_spec}\fbadge{data\_table}\fbadge{l1\_caption}\fbadge{l2\_caption}\imgbadge
      \vspace{3pt}\hrule\vspace{3pt}
      {\sffamily\tiny\bfseries\color{gray!80!black}METADATA COMPLETENESS \hfill Overall 4.0/5}\\[2pt]
      {\sffamily\tiny Data Values \hfill 5/5}\\
      {\sffamily\tiny Trend Magnitude \hfill 2/5}\\
      {\sffamily\tiny Chart Subject Domain \hfill 5/5}
      \vspace{3pt}\hrule\vspace{3pt}
      {\sffamily\tiny
\noindent{\sffamily\tiny\bfseries\color{gray!90!black}CONTEXTUAL INSIGHTS}\\[2pt]
The chart is an area chart tracking unemployment counts across various industries over a ten-year period, from January 2000 to February 2010 \tagdirect{}. This dataset is highly likely to represent United States labor market data, given the timeline and typical economic research benchmarks \tagspeculative{US Bureau of Labor Statistics}. The decade-long span captures two distinct economic cycles, most notably the transition into the global financial crisis of the late 2000s \tagspeculative{Macroeconomic History}.

\vspace{4pt}\noindent{\sffamily\tiny\bfseries\color{gray!90!black}DOMAIN-SPECIFIC EXPLANATIONS}\\[2pt]
In labor economics, 'count' typically refers to the number of unemployed individuals, often measured in thousands; here, the values range from a low of 4,786 to a peak of 15,125, which aligns with US historical unemployment levels of approximately 4.7 million to 15.1 million people \tagderived{Comparing the Y-axis range of 4,786 to 15,125 with historical US unemployment counts}. Between January 2000 and December 2007, the labor market remained relatively stable, with the average monthly unemployment count fluctuating moderately between 6,417 and 7,194 \tagderived{Comparing the average sum counts of the first three time intervals in the Olli tree}. However, a massive structural shift occurred after December 2007, where the average monthly unemployment count surged to 11,029 \tagderived{Reading the fourth time interval average from the Olli tree}.

\vspace{4pt}\noindent{\sffamily\tiny\bfseries\color{gray!90!black}REAL-WORLD SIGNIFICANCE}\\[2pt]
The dramatic escalation in unemployment starting in late 2007 directly corresponds to the onset of the Great Recession, which officially began in December 2007 \tagspeculative{National Bureau of Economic Research (NBER)}. The peak of over 15,000 (representing roughly 15 million individuals) highlights the severity of job losses across industries during this crisis \tagspeculative{Labor Economics}. This period of high unemployment led to widespread economic distress, necessitating unprecedented federal interventions such as extended unemployment benefits and economic stimulus packages \tagspeculative{US Economic Policy History}.

\vspace{4pt}\noindent{\sffamily\tiny\bfseries\color{gray!90!black}OTHER}\\[2pt]
Out of the 122 months recorded in this dataset, only 15 months experienced extreme unemployment levels exceeding 10,000, indicating that these critical conditions were heavily concentrated during the height of the Great Recession between 2008 and 2010 \tagderived{Comparing the 15 values in the 10,000-15,125 range to the total 122 values across the timeline}.
}
    \end{tcolorbox}
    
  \end{paracol}

  \vspace{1em}
  \subsubsection{Olli Gallery V3 Outputs}%
  \vspace{0.5em}
  \begin{paracol}{3}
    
    \begin{tcolorbox}[modelcard, title={GPT-5.4 \hfill \normalfont\tiny\color{gray}1 runs}]
      \fbadge{V1}\fbadge{scene\_graph}\fbadge{data\_table}\fbadge{vega\_lite\_spec}\fbadge{l1\_caption}\fbadge{l2\_caption}\imgbadge
      \vspace{3pt}\hrule\vspace{3pt}
      {\sffamily\tiny\bfseries\color{gray!80!black}METADATA COMPLETENESS \hfill Overall 3.7/5}\\[2pt]
      {\sffamily\tiny Data Values \hfill 4/5}\\
      {\sffamily\tiny Trend Magnitude \hfill 4/5}\\
      {\sffamily\tiny Chart Subject Domain \hfill 3/5}
      \vspace{3pt}\hrule\vspace{3pt}
      {\sffamily\tiny
\noindent{\sffamily\tiny\bfseries\color{gray!90!black}CONTEXTUAL INSIGHTS}\\[2pt]
This is a time-series chart showing monthly counts from roughly 1999 to 2010. \tagdirect{} The measured quantity is labeled only as 'count,' so the subject being counted is not identified in the provided materials. \tagdirect{} The series appears to begin near 6,000, fluctuate mostly between about 5,000 and 9,000 for much of the early and middle period, and then rise sharply near the end to around 15,000. \tagderived{read approximate levels and overall pattern from the plotted area against the axis ticks} Because no title, source, or domain metadata is given, the chart's topic, producer, and intended use cannot be determined from the provided information alone. \tagdirect{}

\vspace{4pt}\noindent{\sffamily\tiny\bfseries\color{gray!90!black}DOMAIN-SPECIFIC EXPLANATIONS}\\[2pt]
The x-axis uses year-month dates, so each point likely represents a monthly total rather than a cumulative running total. \tagderived{inferred from x-axis label 'date (year-month)'} In many domains, a monthly count series is used to track recurring events such as transactions, incidents, visits, or claims, but this chart does not specify which of these applies. \tagspeculative{common statistical reporting practice} The late jump suggests either a real increase in the underlying phenomenon, a change in coverage or reporting, or a structural break in how the count was measured; the image alone cannot distinguish among these explanations. \tagspeculative{time-series analysis} The lack of units beyond raw count means the chart cannot by itself tell whether the increase reflects population growth, seasonal demand, policy change, economic conditions, or improved data collection. \tagspeculative{statistics and data interpretation}

\vspace{4pt}\noindent{\sffamily\tiny\bfseries\color{gray!90!black}REAL-WORLD SIGNIFICANCE}\\[2pt]
A sustained rise from roughly the mid-thousands to around 15,000 would usually matter operationally because it could imply higher demand, workload, or incidence over time. \tagspeculative{operations and public-sector analytics} If this were tracking public services, health events, transport use, crime reports, or sales, decision-makers might use such a pattern to adjust staffing, budgets, capacity, or interventions. \tagspeculative{public administration and business analytics} However, because the chart does not identify what is being counted, any statement about who is affected or what policy decisions depend on it would be conjectural. \tagdirect{}

\vspace{4pt}\noindent{\sffamily\tiny\bfseries\color{gray!90!black}OTHER}\\[2pt]
The chart shows noticeable month-to-month variability throughout the series rather than a perfectly smooth trend. \tagderived{observed repeated short-term ups and downs in the plotted series} The most prominent feature is the abrupt acceleration in the final part of the timeline, which is much steeper than earlier changes. \tagderived{compared slope and level changes across the series} Interpretation is strongly limited by missing metadata, especially the absent title, data table, source, and domain context. \tagdirect{}
}
    \end{tcolorbox}
    
    \switchcolumn
    \begin{tcolorbox}[modelcard, title={llama-4-scout-17... \hfill \normalfont\tiny\color{gray}1 runs}]
      \fbadge{V1}\fbadge{scene\_graph}\fbadge{vega\_lite\_spec}\fbadge{data\_table}\fbadge{l1\_caption}\fbadge{l2\_caption}\imgbadge
      \vspace{3pt}\hrule\vspace{3pt}
      {\sffamily\tiny\bfseries\color{gray!80!black}METADATA COMPLETENESS \hfill Overall 4.7/5}\\[2pt]
      {\sffamily\tiny Data Values \hfill 5/5}\\
      {\sffamily\tiny Trend Magnitude \hfill 4/5}\\
      {\sffamily\tiny Chart Subject Domain \hfill 5/5}
      \vspace{3pt}\hrule\vspace{3pt}
      {\sffamily\tiny
\noindent{\sffamily\tiny\bfseries\color{gray!90!black}CONTEXTUAL INSIGHTS}\\[2pt]
This is a time-series area chart tracking a specific volume of activity over a ten-year period from 2000 to 2010 \tagdirect{}. The chart tracks the frequency or quantity of an event, labeled simply as 'count' \tagdirect{}. Without a title or legend, the specific subject matter is unknown \tagspeculative{general chart literacy}, but it represents a longitudinal dataset typically used for monitoring growth or usage rates \tagspeculative{data analysis context}.

\vspace{4pt}\noindent{\sffamily\tiny\bfseries\color{gray!90!black}DOMAIN-SPECIFIC EXPLANATIONS}\\[2pt]
The data begins in the year 2000 with a count of approximately 6,000 \tagderived{visual estimation of Y-axis start point}. From 2000 to 2008, the count exhibits high volatility, fluctuating mostly between the 6,000 and 8,000 range \tagderived{visual analysis of the plateau period}. A significant structural shift occurs around 2008, where the count begins a steep ascent, eventually breaking the previous maximums to reach a peak of approximately 15,000 by the end of 2010 \tagderived{visual analysis of the growth phase}. The data granularity appears to be monthly, given the 'year-month' label and the jagged nature of the peaks \tagdirect{}.

\vspace{4pt}\noindent{\sffamily\tiny\bfseries\color{gray!90!black}REAL-WORLD SIGNIFICANCE}\\[2pt]
The sharp increase between 2008 and 2010 suggests a major event, product launch, or market shift occurred during that window \tagspeculative{business/internet history}. In a digital context, this pattern is consistent with the rapid adoption of Web 2.0 technologies or social media platforms during that era \tagspeculative{technology history}. In an economic context, the break in the trend could correlate with post-2008 recession recovery efforts or stimulus measures \tagspeculative{economic history}. Regardless of the specific subject, the data indicates that the metric being tracked has become significantly more prevalent or active in the final two years compared to the previous eight \tagderived{comparison of trend phases}.

\vspace{4pt}\noindent{\sffamily\tiny\bfseries\color{gray!90!black}OTHER}\\[2pt]
The chart establishes a baseline of 'normal' activity (6k-8k) for the majority of the timeline, which serves as a contrast to the outlier behavior seen at the end of the decade \tagderived{visual pattern recognition}.
}
    \end{tcolorbox}
    
    \switchcolumn
    \begin{tcolorbox}[modelcard, title={Gemini 3.5 Flash \hfill \normalfont\tiny\color{gray}1 runs}]
      \fbadge{V1}\fbadge{scene\_graph}\fbadge{vega\_lite\_spec}\fbadge{data\_table}\fbadge{l1\_caption}\fbadge{l2\_caption}\imgbadge
      \vspace{3pt}\hrule\vspace{3pt}
      {\sffamily\tiny\bfseries\color{gray!80!black}METADATA COMPLETENESS \hfill Overall 4.0/5}\\[2pt]
      {\sffamily\tiny Data Values \hfill 5/5}\\
      {\sffamily\tiny Trend Magnitude \hfill 2/5}\\
      {\sffamily\tiny Chart Subject Domain \hfill 5/5}
      \vspace{3pt}\hrule\vspace{3pt}
      {\sffamily\tiny
\noindent{\sffamily\tiny\bfseries\color{gray!90!black}CONTEXTUAL INSIGHTS}\\[2pt]
This chart displays a time-series area graph representing the total count of unemployed individuals over time, spanning from late 1999 to early 2010 \tagderived{read from x-axis 'date (year-month)' and y-axis 'count'}. The dataset is highly characteristic of the 'unemployment-across-industries' dataset from the Bureau of Labor Statistics, commonly used in data visualization libraries \tagspeculative{Vega-Lite datasets}. The y-axis represents the count of unemployed individuals, typically measured in thousands \tagspeculative{US Bureau of Labor Statistics}.

\vspace{4pt}\noindent{\sffamily\tiny\bfseries\color{gray!90!black}DOMAIN-SPECIFIC EXPLANATIONS}\\[2pt]
In macroeconomics, unemployment counts are lagging indicators of economic health \tagspeculative{Macroeconomics}. The chart shows a rise in unemployment peaking in 2003 at approximately 9,000 thousand (9 million) people, which aligns with the aftermath of the dot-com bubble burst and the 2001 recession \tagspeculative{Macroeconomics}. Following a period of economic expansion where unemployment fell back to around 6,000 thousand in 2006, the chart shows an unprecedented surge starting in late 2008 \tagderived{compared the 2006 low of ~6,000 to the late 2008 upward trend}. This massive spike peaks near 15,000 thousand (15 million) in late 2009, reflecting the severe impact of the Great Recession \tagspeculative{Macroeconomics}.

\vspace{4pt}\noindent{\sffamily\tiny\bfseries\color{gray!90!black}REAL-WORLD SIGNIFICANCE}\\[2pt]
This data is critical for policymakers, such as the Federal Reserve and government officials, who monitor unemployment to determine monetary policy, interest rates, and the necessity of economic stimulus packages \tagspeculative{Public Policy}. The dramatic increase from 6,000 thousand unemployed in 2006 to over 15,000 thousand in 2009 highlights the devastating scale of the 2008 financial crisis on the workforce \tagderived{compared the low point of ~6,000 in 2006 to the peak of ~15,000 in 2009/2010}. Such high levels of unemployment signify widespread financial distress, reduced consumer spending, and increased demand for social safety nets \tagspeculative{Economics}.

\vspace{4pt}\noindent{\sffamily\tiny\bfseries\color{gray!90!black}OTHER}\\[2pt]
The seasonal fluctuations visible as small monthly peaks and valleys throughout the decade reflect regular annual hiring cycles in industries like retail and agriculture \tagspeculative{Labor Economics}.
}
    \end{tcolorbox}
    
  \end{paracol}

  \vspace{1em}
  \subsubsection{Olli Gallery V4 Outputs}%
  \vspace{0.5em}
  \begin{paracol}{3}
    
    \begin{tcolorbox}[modelcard, title={GPT-5.4 \hfill \normalfont\tiny\color{gray}1 runs}]
      \fbadge{V1}\fbadge{scene\_graph}\fbadge{data\_table}\fbadge{vega\_lite\_spec}\fbadge{l1\_caption}\fbadge{l2\_caption}\imgbadge
      \vspace{3pt}\hrule\vspace{3pt}
      {\sffamily\tiny\bfseries\color{gray!80!black}METADATA COMPLETENESS \hfill Overall 3.7/5}\\[2pt]
      {\sffamily\tiny Data Values \hfill 4/5}\\
      {\sffamily\tiny Trend Magnitude \hfill 4/5}\\
      {\sffamily\tiny Chart Subject Domain \hfill 3/5}
      \vspace{3pt}\hrule\vspace{3pt}
      {\sffamily\tiny
\noindent{\sffamily\tiny\bfseries\color{gray!90!black}CONTEXTUAL INSIGHTS}\\[2pt]
This chart shows a monthly time series of counts from roughly 1999 to 2010. \tagdirect{} The measured quantity is a count observed over time, but the chart image alone does not identify what is being counted. \tagdirect{} The series appears to fluctuate around about 5,000 to 8,500 in the early and middle years, then rises sharply near the end to around 15,000 by 2010. \tagderived{estimated trend and approximate values from the plotted area against the axis labels} Because the chart lacks a title, legend, and source in the visible image, the producing organization and intended purpose cannot be determined from the provided material. \tagdirect{}

\vspace{4pt}\noindent{\sffamily\tiny\bfseries\color{gray!90!black}DOMAIN-SPECIFIC EXPLANATIONS}\\[2pt]
The y-axis uses raw counts rather than percentages or rates. \tagdirect{} That means interpretation depends heavily on what the units represent, because a rise in counts could reflect more underlying events, a larger population at risk, broader reporting, or changes in collection practices. \tagspeculative{statistical interpretation of count time series} The repeated month-by-month variation suggests short-term volatility and possibly seasonality, but seasonality cannot be confirmed confidently from the image alone. \tagderived{visual inspection of recurring fluctuations across months} Without knowing the subject domain, it is not possible to say whether higher values are desirable, harmful, or neutral. \tagdirect{}

\vspace{4pt}\noindent{\sffamily\tiny\bfseries\color{gray!90!black}REAL-WORLD SIGNIFICANCE}\\[2pt]
The most important substantive signal in the image is the large late-period increase in monthly counts. \tagderived{comparison of end-period values to earlier years} If this increase reflects a real-world phenomenon rather than a measurement change, it could indicate growing demand, incidence, activity, or burden in whatever system is being tracked. \tagspeculative{general time-series interpretation} Decisions about staffing, budgeting, infrastructure, public communication, or policy could depend on such a rise, but which decisions matter most cannot be identified without the missing subject context. \tagspeculative{public administration and operations analysis} If the elevated end-of-series level continued, the responsible organization might face sustained pressure compared with earlier years. \tagspeculative{organizational planning and capacity management}

\vspace{4pt}\noindent{\sffamily\tiny\bfseries\color{gray!90!black}OTHER}\\[2pt]
The chart image supports only limited semantic interpretation because it omits the title, source, unit definition, and category being counted. \tagdirect{} Approximate reading suggests the final value is about two-and-a-half to three times the low point near the start of the series. \tagderived{rough ratio from approximately 15,000 at the end versus approximately 5,000 near the early trough} Any stronger claim about causes, sector, or implications would require metadata that has been explicitly withheld. \tagdirect{}
}
    \end{tcolorbox}
    
    \switchcolumn
    \begin{tcolorbox}[modelcard, title={llama-4-scout-17... \hfill \normalfont\tiny\color{gray}1 runs}]
      \fbadge{V1}\fbadge{scene\_graph}\fbadge{vega\_lite\_spec}\fbadge{data\_table}\fbadge{l1\_caption}\fbadge{l2\_caption}\imgbadge
      \vspace{3pt}\hrule\vspace{3pt}
      {\sffamily\tiny\bfseries\color{gray!80!black}METADATA COMPLETENESS \hfill Overall 4.7/5}\\[2pt]
      {\sffamily\tiny Data Values \hfill 5/5}\\
      {\sffamily\tiny Trend Magnitude \hfill 4/5}\\
      {\sffamily\tiny Chart Subject Domain \hfill 5/5}
      \vspace{3pt}\hrule\vspace{3pt}
      {\sffamily\tiny
\noindent{\sffamily\tiny\bfseries\color{gray!90!black}CONTEXTUAL INSIGHTS}\\[2pt]
The chart visualizes a time-series dataset spanning a decade from the year 2000 to 2010 \tagdirect{}. It tracks a metric labeled simply as 'count' \tagdirect{}. The data exhibits a pattern of relative stability for the first eight years, followed by a sudden, dramatic surge in volume starting around 2008 \tagderived{visual trend analysis of the area chart}. The source, producer, or specific subject matter of this count is not provided in the image \tagdirect{}.

\vspace{4pt}\noindent{\sffamily\tiny\bfseries\color{gray!90!black}DOMAIN-SPECIFIC EXPLANATIONS}\\[2pt]
The Y-axis represents a frequency or volume measurement ranging from 0 to 16,000 \tagdirect{}. The X-axis represents time in a 'year-month' format \tagdirect{}. In the absence of specific domain labels, the values represent a raw quantity of events or items per month \tagspeculative{general data interpretation}. The magnitude of the spike—reaching approximately 15,000 by the end of 2010 compared to a baseline of roughly 6,000–8,000 in previous years—indicates a significant shift in the underlying system or behavior being measured \tagderived{comparison of 2000-2007 levels vs. 2010 levels}.

\vspace{4pt}\noindent{\sffamily\tiny\bfseries\color{gray!90!black}REAL-WORLD SIGNIFICANCE}\\[2pt]
The data indicates a phenomenon that experienced explosive growth between 2008 and 2010 \tagderived{trend observation}. Without context, the real-world implication is unknown, but the scale (reaching $\approx$15,000) suggests a high-volume activity \tagderived{Y-axis reading}. If this were a negative metric (e.g., crime or errors), the trend would represent a significant crisis emerging in 2008 \tagspeculative{general knowledge}. If this were a positive metric (e.g., internet users or sales), it would represent a massive market adoption or viral growth event during that same period \tagspeculative{general knowledge}.

\vspace{4pt}\noindent{\sffamily\tiny\bfseries\color{gray!90!black}OTHER}\\[2pt]
The chart uses an area plot to visualize cumulative or continuous data over time \tagdirect{}. The visual style is generic, suggesting it may be a standard statistical output or a placeholder visualization \tagspeculative{design analysis}.
}
    \end{tcolorbox}
    
    \switchcolumn
    \begin{tcolorbox}[modelcard, title={Gemini 3.5 Flash \hfill \normalfont\tiny\color{gray}1 runs}]
      \fbadge{V1}\fbadge{scene\_graph}\fbadge{vega\_lite\_spec}\fbadge{data\_table}\fbadge{l1\_caption}\fbadge{l2\_caption}\imgbadge
      \vspace{3pt}\hrule\vspace{3pt}
      {\sffamily\tiny\bfseries\color{gray!80!black}METADATA COMPLETENESS \hfill Overall 4.0/5}\\[2pt]
      {\sffamily\tiny Data Values \hfill 5/5}\\
      {\sffamily\tiny Trend Magnitude \hfill 2/5}\\
      {\sffamily\tiny Chart Subject Domain \hfill 5/5}
      \vspace{3pt}\hrule\vspace{3pt}
      {\sffamily\tiny
\noindent{\sffamily\tiny\bfseries\color{gray!90!black}CONTEXTUAL INSIGHTS}\\[2pt]
The chart displays a monthly time series from late 1999 to early 2010 \tagderived{read from x-axis}. The vertical axis represents a 'count' ranging from 0 to 16,000 \tagdirect{}. Based on the specific shape, timeline, and values (peaking at ~9,000 in 2003 and spiking to over 15,000 in 2009-2010), this chart highly resembles the US unemployment count (in thousands) from the classic 'economics' dataset in R's ggplot2 library, originally sourced from the US Bureau of Labor Statistics \tagspeculative{R ggplot2 economics dataset / US Bureau of Labor Statistics}.

\vspace{4pt}\noindent{\sffamily\tiny\bfseries\color{gray!90!black}DOMAIN-SPECIFIC EXPLANATIONS}\\[2pt]
If this represents US unemployment, the 'count' is likely in thousands, meaning a value of 15,000 represents 15 million unemployed individuals \tagspeculative{US macroeconomic reporting standards}. The fluctuations show cyclical economic behavior: the peak in 2003 corresponds to the aftermath of the early 2000s dot-com recession \tagspeculative{US economic history}, while the massive spike starting in late 2008 represents the onset of the Great Recession triggered by the subprime mortgage crisis \tagspeculative{US economic history}. The rapid rise from approximately 6,000 in 2007 to over 15,000 in late 2009 reflects a doubling of the unemployed population within a two-year span \tagderived{compared values from 2007 to 2009 on the chart}.

\vspace{4pt}\noindent{\sffamily\tiny\bfseries\color{gray!90!black}REAL-WORLD SIGNIFICANCE}\\[2pt]
Unemployment metrics are critical indicators of macroeconomic health, influencing central bank decisions on interest rates and government fiscal policy \tagspeculative{Macroeconomics}. The dramatic spike in 2008-2009 represents widespread job losses, leading to severe financial hardship for millions of households, increased demand for social safety nets, and long-term economic scarring \tagspeculative{Sociology / Labor Economics}. A sustained high unemployment count of over 14,000 as seen in late 2009 indicates a severe labor market crisis requiring federal intervention, such as extended unemployment benefits and stimulus packages \tagspeculative{US Public Policy}.

\vspace{4pt}\noindent{\sffamily\tiny\bfseries\color{gray!90!black}OTHER}\\[2pt]
The data shows a peak of approximately 9,000 around 2003, a decline to nearly 6,000 around 2007, and a sharp spike to over 15,000 by late 2009 \tagderived{read values from the chart over time}.
}
    \end{tcolorbox}
    
  \end{paracol}

\clearpage
\section{Section-Level Claim Counts and Output Length}
\label{sec:supp-section-distributions}

Figure~\ref{fig:supp_section_distributions} complements the
chart-paired analyses in the main paper by showing raw, section-level
claim counts for each model and input condition. The stacked bars
represent model-attributed \textsc{Direct}, \textsc{Derived}, and
\textsc{Speculative} claims. The dashed line reports the mean generated
output length in words using the right-hand axis. Because the bars show
raw counts rather than within-output proportions, this figure is
descriptive and should be interpreted alongside the normalized,
chart-paired comparisons reported in the main paper.

\begin{figure*}[htb]
  \centering

  \includegraphics[
    width=0.72\textwidth,
    alt={Legend showing three stacked claim categories: Direct in green, Derived in blue, and Speculative in orange. A dashed gray line represents mean output length in words using the right-hand axis.}
  ]{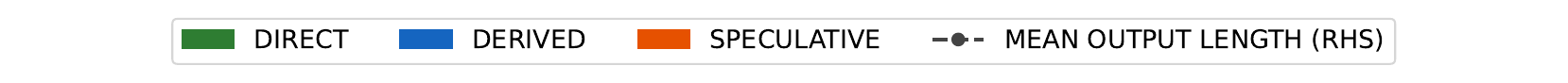}

  \vspace{0.7em}

  \begin{subfigure}[t]{0.98\textwidth}
    \centering
    \includegraphics[
      width=\linewidth,
      alt={Contextual Insights section. Three groups of stacked bar charts compare Gemini 3.5 Flash, GPT-5.4, and Llama 4 Scout across conditions V1 through V4. Each stacked bar shows counts of Direct, Derived, and Speculative claims, with a dashed line showing mean output length.}
    ]{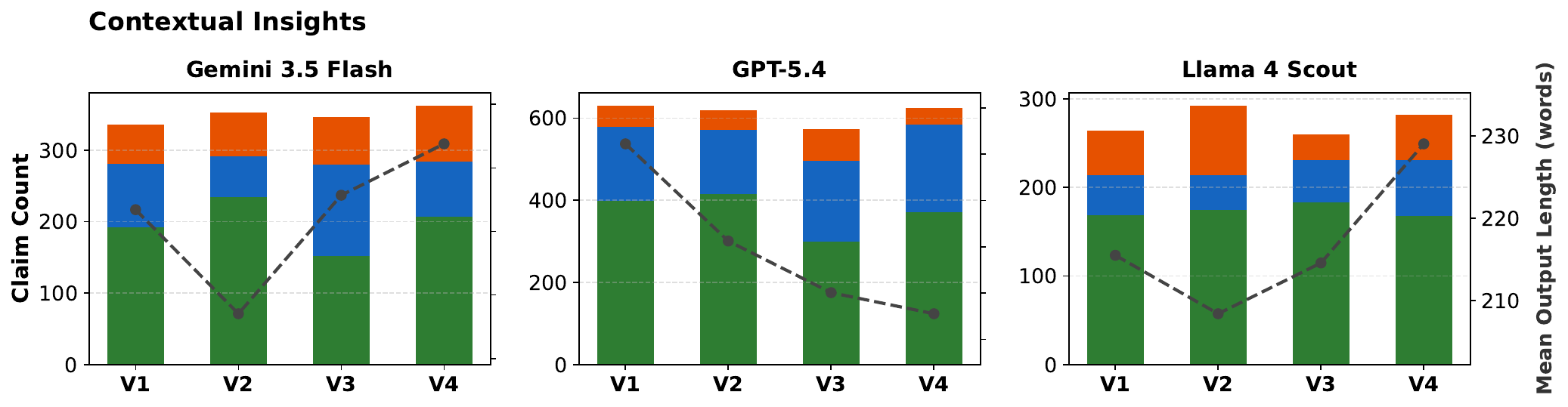}
    \caption{\emph{Contextual Insights}.}
    \label{fig:supp_contextual_insights}
  \end{subfigure}

  \vspace{0.9em}

  \begin{subfigure}[t]{0.98\textwidth}
    \centering
    \includegraphics[
      width=\linewidth,
      alt={Domain-Specific Explanations section. Three groups of stacked bar charts compare Gemini 3.5 Flash, GPT-5.4, and Llama 4 Scout across V1 through V4. Bars show Direct, Derived, and Speculative claim counts, with a dashed line showing mean output length.}
    ]{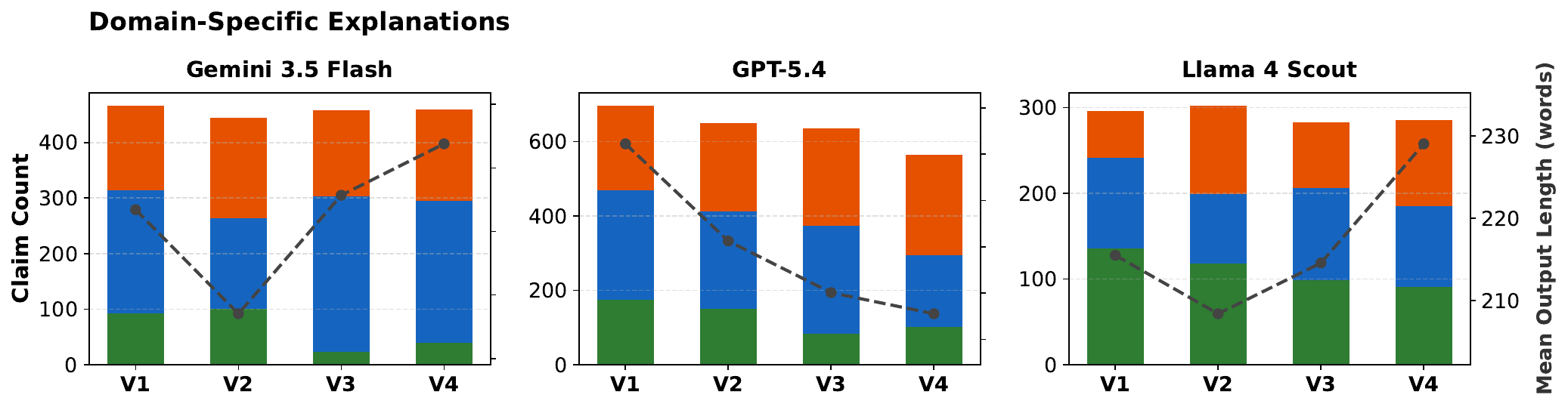}
    \caption{\emph{Domain-Specific Explanations}.}
    \label{fig:supp_domain_explanations}
  \end{subfigure}

  \caption{
    Section-level claim counts and mean output length by model and input condition. Stacked bars show raw counts of model-attributed
    \textsc{Direct}, \textsc{Derived}, and \textsc{Speculative} claims, while the dashed line shows mean output length (right-hand axis). Panels (a) and (b) correspond to the \emph{Contextual Insights} and \emph{Domain-Specific Explanations} sections. Figure continued on the following page.
  }
  \label{fig:supp_section_distributions}
\end{figure*}

\clearpage

\begin{figure*}[htb]
  \ContinuedFloat
  \centering

  \includegraphics[
    width=0.72\textwidth,
    alt={Legend showing Direct, Derived, and Speculative claim categories and the dashed mean-output-length line.}
  ]{figs/global_legend.pdf}

  \vspace{0.7em}

  \begin{subfigure}[t]{0.98\textwidth}
    \centering
    \includegraphics[
      width=\linewidth,
      alt={Real-World Significance section. Three groups of stacked bar charts compare Gemini 3.5 Flash, GPT-5.4, and Llama 4 Scout across V1 through V4. The orange Speculative category dominates nearly every bar, while the dashed line shows mean output length.}
    ]{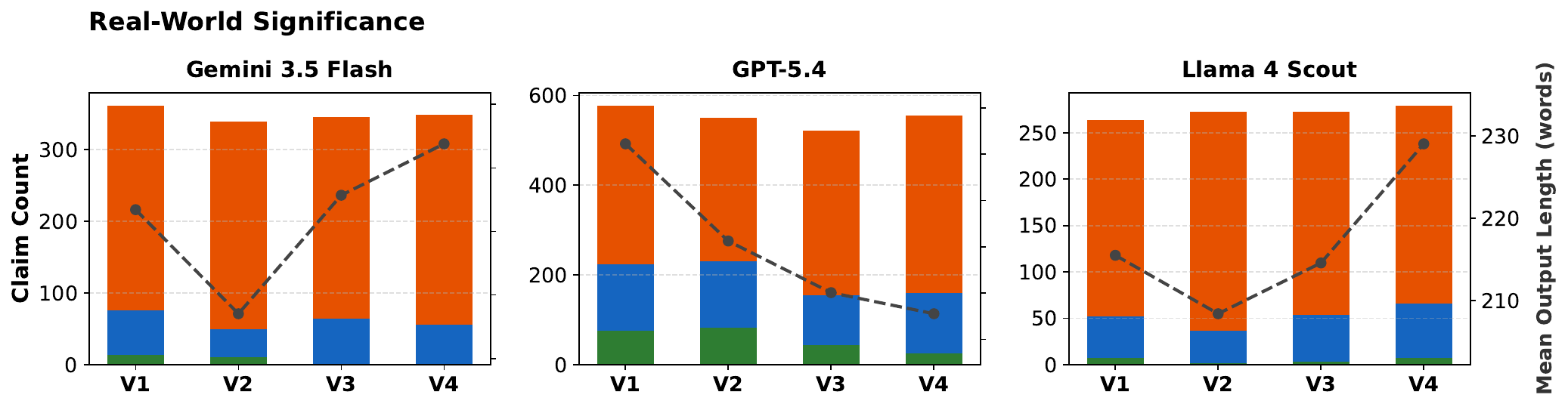}
    \caption{\emph{Real-World Significance}.}
    \label{fig:supp_real_world_significance}
  \end{subfigure}

  \vspace{0.9em}

  \begin{subfigure}[t]{0.98\textwidth}
    \centering
    \includegraphics[
      width=\linewidth,
      alt={Other section. Three groups of stacked bar charts compare Gemini 3.5 Flash, GPT-5.4, and Llama 4 Scout across V1 through V4. Bars show Direct, Derived, and Speculative claim counts together with mean output length.}
    ]{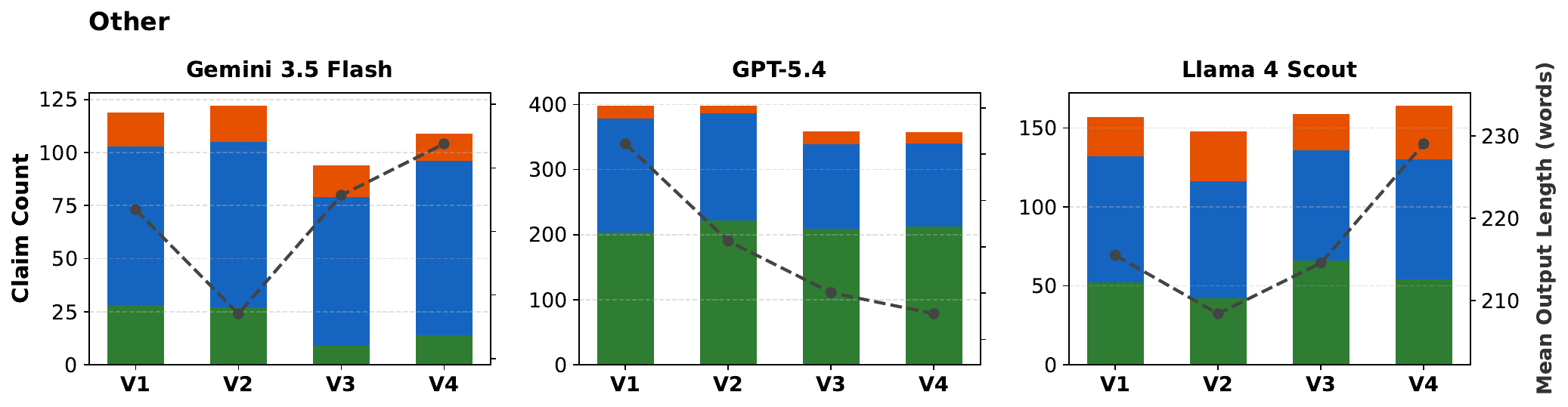}
    \caption{\emph{Other}.}
    \label{fig:supp_other}
  \end{subfigure}

  \caption[]{
    Section-level claim counts and mean output length by model and input condition, continued. Panels (c) and (d) correspond to the \emph{Real-World Significance} and \emph{Other} sections. The \emph{Real-World Significance} panel is visually dominated by \textsc{Speculative} claims across models and conditions, consistent with the quantitative results reported in the main paper.
  }
\end{figure*}

\FloatBarrier

\end{document}